\documentclass[10pt,journal,compsoc]{IEEEtran}

\usepackage{graphicx,times,amsmath,amssymb,cite,cases,mathrsfs,booktabs,multirow,makecell,psfrag,subfigure,stfloats}
\usepackage{algorithm,algorithmic,amsthm,ulem,latexsym,bm}
\usepackage{dsfont,diagbox}
\usepackage[colorlinks,linkcolor=black,anchorcolor=blue,citecolor=red]{hyperref}

\newtheorem{theorem}{Theorem}
\newtheorem{lemma}[theorem]{Lemma}
\graphicspath{{images/}}

\newcommand{\tbfmu}{\tilde{\bm{\mu}}}

\newcommand{\bfw}{\mathbf{w}}
\newcommand{\hbfw}{\hat{\mathbf{w}}}

\begin{document}

\title{Unsupervised Domain Adaptation via Discriminative Manifold Propagation}
\author{You-Wei Luo, Chuan-Xian Ren, Dao-Qing Dai, Hong Yan~\IEEEmembership{Fellow,~IEEE}
\thanks{Y.W. Luo, C.X. Ren, and D.Q. Dai are with the Intelligent Data Center, School of Mathematics, Sun Yat-Sen University, Guangzhou, 510275, China. H. Yan is with the Department of Electrical Engineering, City University of Hong Kong, Hong Kong.  Luo and Ren contributed equally to this work.}
\thanks{This work is supported in part by the National Natural Science Foundation of China under Grants 61976229, 61906046, 61572536, 11631015, U1611265, in part by the Science and Technology Program of Guangzhou under Grant 201804010248 and City University of Hong Kong (Project 9610460).} }

\IEEEtitleabstractindextext{%
\begin{abstract}
Unsupervised domain adaptation is effective in leveraging rich information from a labeled source domain to an unlabeled target domain. Though deep learning and adversarial strategy made a significant breakthrough in the adaptability of features, there are two issues to be further studied. First, hard-assigned pseudo labels on the target domain are arbitrary and error-prone, and direct application of them may destroy the intrinsic data structure. Second, batch-wise training of deep learning limits the characterization of the global structure. In this paper, a Riemannian manifold learning framework is proposed to achieve transferability and discriminability simultaneously. For the first issue, this framework establishes a probabilistic discriminant criterion on the target domain via soft labels. Based on pre-built prototypes, this criterion is extended to a global approximation scheme for the second issue. Manifold metric alignment is adopted to be compatible with the embedding space. The theoretical error bounds of different alignment metrics are derived for constructive guidance. The proposed method can be used to tackle a series of variants of domain adaptation problems, including both vanilla and partial settings. Extensive experiments have been conducted to investigate the method and a comparative study shows the superiority of the discriminative manifold learning framework.
\end{abstract}

}

\maketitle

\IEEEdisplaynontitleabstractindextext

%
\IEEEpeerreviewmaketitle

\IEEEraisesectionheading{\section{Introduction}\label{sec:introduction}}
\IEEEPARstart{I}{n} machine learning, the amount of labeled data plays a crucial role during the learning process. Convolutional Neural Networks (CNNs) can achieve a significant advance in various tasks via a large number of well-labeled samples. Unfortunately, such data are often prohibitively expensive to obtain in many real-world scenarios. Applying a learned model in a new environment, i.e., the cross-domain scheme, may cause a significant degradation of recognition performance \cite{hu2015deep,long2019learning}.

\begin{figure}[t]
\centering
\includegraphics[width=1\linewidth]{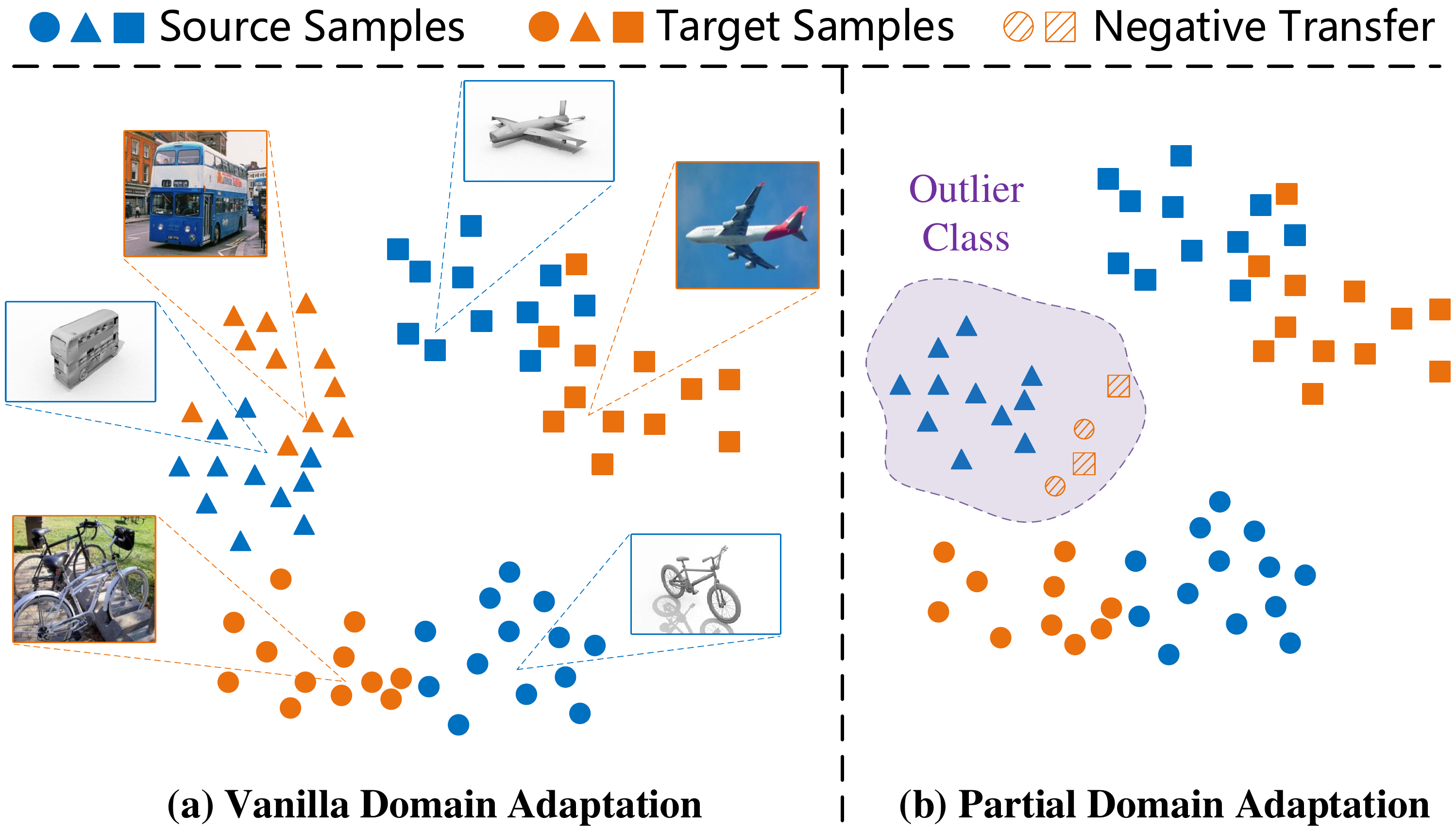}
   \caption{Illustration of the vanilla and partial domain adaptation problems. (a) Vanilla domain adaptation assumes that the label spaces of the source and target domains are equivalent, and the \textit{domain shift} problem is the major difficulty. (b) Partial domain adaptation assumes that the label space of the target domain is a subset of the source domain. Performance can be further degraded by the \textit{negative transfer} problem, which means that target samples are aligned to outlier classes (e.g., the ``triangle''). Best viewed in color.}
\label{fig:problem_illustration}
\vspace{-10pt}
\end{figure}

Unsupervised Domain Adaptation (UDA) is designed to deal with the shortage of labels by leveraging rich labels and strong supervision from the source domain to the target domain where there is no access to the annotations \cite{shao2014transfer}. Datasets composed specifically of exploratory factors and variants, such as background, style, illumination, camera views or resolution, often lead to shifting distributions (the domain shift) \cite{pan2010domain}. Classical UDA assumes the label spaces of two domains are equivalent, i.e., the vanilla UDA setting in Figure \ref{fig:problem_illustration}(a). According to the transfer theory established by Ben-David et al. \cite{ben2010theory}, the primary task for cross-domain adaptation is to learn the discriminative representations while narrowing the discrepancy between domains. As this strong condition in the vanilla setting is an obstacle to some special real-world applications, partial UDA is explored as a subproblem of UDA, i.e., Figure \ref{fig:problem_illustration}(b). Partial UDA relaxes the equivalence assumption on the label space by taking the label space of the target domain as a subset of the source domain \cite{zhang2018importance,cao2018PADA}. Under the partial setting, the source domain is still a large-scale annotated dataset (e.g., ImageNet \cite{ImageNet}) while the target domain can be any smaller dataset with fewer categories (e.g., Caltech-256 \cite{griffin2007caltech} and PASCAL VOC \cite{everingham2010pascal}). Unfortunately, as generalizations of vanilla methods, partial UDA methods usually also suffer from the negative transfer problem.

Early domain adaptation approaches with hand-crafted features focussed on learning domain-invariant information. Statistical methods attempt to align the domains in embedding space based on moment statistics, such as the mean and covariance \cite{pan2010domain,long2013transfer}, and then minimize the distribution discrepancy in that space. Manifold and subspace learning methods also achieve significant successes in transfer learning tasks \cite{ren2019heterogeneous,moon2020multi}. In \cite{gopalan2011domain}, the authors build a geodesic flow on the manifold to model the domain shift and form an optimal distance measurement between samples from different domains. Moon et al. \cite{moon2020multi} explore the global information in an online UDA by proposing a mean target space which considers the coherency among all target-data batches. Dictionary learning approaches merge the common features of different domains into a shared dictionary and find sparse or low-rank codings for input samples \cite{shekhar2015coupled}.

CNNs can learn abstract representations with nonlinear transformation \cite{yosinski2014transferable}, which suppress the negative effects caused by variable explanatory factors in domain shift \cite{long2019learning}. Early work \cite{long2019learning,sun2016deep} attempted to transfer the source classifier, with sufficient supervision, to the target domain by minimizing domain discrepancy, supposing that a well distributed alignment leads to an effective application of a trained classifier on the target domain. Adversarial confusion methods \cite{ganin2016domain,sankaranarayanan2018generate}, which are inspired by Generative Adversarial Nets (GANs) \cite{ge2019dual}, produce generated features that are domain-indistinguishable and form a well-aligned marginal distribution. Adversarial-based weighting methods use the discriminator of GANs to evaluate the probability of negative transfer in a partial UDA from the instance-level \cite{zhang2018importance,cao2019learning}, class-level \cite{cao2018PADA}, or both \cite{cao2018SAN}. However, the conditional distributions of those methods are not guaranteed \cite{long2018conditional,ren2018generalized}. Several methods achieve improvements in accuracy by employing uncertainty information, e.g., pseudo labels and soft labels, to enhance the discrimination in the target domain \cite{long2018conditional,pinheiro2018unsupervised,das2018graph}. Other methods revisit the trade-off between transferability and discrimination, and then build a more discriminative \cite{chen2019transferability} or transferable \cite{xu2019larger} model.

In this paper, we propose a novel framework called Discriminative Manifold Propagation (\textbf{DMP}) to deal with both vanilla and partial UDA problems. DMP primarily considers two issues in existing UDA methods. First, direct utilization of uncertain information is error-prone and should be treated cautiously \cite{long2018conditional}, as hard-assigned labels may change the intrinsic data structure \cite{ding2019deep}. Second, batch-wise training in deep learning limits the capture of global information, thus models may be misled by extreme local distributions.
As the ability to transfer and discriminate are both valuable \cite{chen2019transferability}, the method proposed here develops a unified rule for these two properties. The main idea is to describe the domains by a sequence of latent manifolds. In contrast to earlier works, which build discriminative models directly on the source domain \cite{chen2019joint} or both domains \cite{pan2019transferrable}, we establish a more relaxed criterion and enhance target discriminability transductively. We extend this criterion to a global approximation scheme, which overcomes problems caused by batch-wise training. Inspired by prior work on manifold learning \cite{gopalan2011domain,huang2017cross}, we employ the manifold metrics to measure the domain discrepancy.

This paper extends previous work \cite{luo2020unsupervised} by: 1) extending the manifold learning framework by including the affine Grassmann distance and the Log-Euclidean metric; 2) deriving a new error bound for the affine Grassmann distance; 3) studying another variant of UDA (the partial UDA problem) and extending a theoretical error bound to this case; 4) extending a unified algorithm to simultaneously deal with the domain shift and negative transfer problems in different UDA settings; 5) conducting experiments, including parameter selection and ablation analysis, to validate the effectiveness of DMP in both the vanilla and partial settings. Our contributions are summarized as follows.
\begin{itemize}
\item To explore the discriminative structure of the target domain and reduce uncertainty information, a probabilistic manifold embedding criterion is proposed. This criterion constructs an intra-class separable structure on the source domain. Target discriminability is achieved by a probabilistic and truncated intra-class compactness constraint and the inter-class separability is transduced from the source domain. A global structure learning scheme is extended based on the pre-built prototypes.
\item A manifold alignment framework that is consistent with the manifold assumption on the embedding space is proposed. It establishes a series of abstract descriptors (i.e. the basis) for the original data based on different manifolds, and aligns the domains by minimizing the discrepancy between the abstract descriptors. The theoretical error bounds are derived to facilitate the selection of components.
\item Both the vanilla and partial UDA problems can be tackled effectively by the proposed method. It extends the discriminant criterion and manifold alignment to a weighting scheme, which alleviates negative transfer from the outlier classes. A theoretical error bound is derived under the partial setting, which gives a theoretical interpretation of the proposed weighting strategy.
\end{itemize}

\section{Related Work}\label{sec:RelatedWorks}
In this section, we review the two subproblems of UDA. The primary goal of UDA is to learn a classifier that generalizes well on the target domain. Since there are no annotations on the target domain, the key is to learn the abstract representations, which transduce discriminative information about the source to the target domains.

\subsection{Vanilla UDA Methods}\label{sec:VanillaDA}
The label space of the source domain $\mathcal{X}^s$ and target domain $\mathcal{X}^t$ is denoted as $\mathcal{C}^s$ and $\mathcal{C}^t$, respectively. Under the vanilla UDA setting, the label spaces completely overlap, i.e., $\mathcal{C}^s = \mathcal{C}^t$. A simple yet effective solution is to mitigate the domain shift or bias.

Approaches with hand-crafted features usually focus on the learning of domain-invariant or discriminative features \cite{pan2010domain,long2013transfer}. Based on the manifold assumption, Gopalan et al. \cite{gopalan2011domain} take the source and target domains as points on the Grassmann manifold, and propose to generate multiple subspaces from those points. Then the distance is measured by the geodesic flow between those points. Shekhar et al. \cite{shekhar2015coupled} construct a shared dictionary for both source and target domains, while minimizing the reconstructed error based on the learned dictionary. Ren et al. \cite{ren2019heterogeneous} explore an optimal experimental design based on the covariance of structured feature translators to tackle the nonlinear and heterogeneity problems. Das et al. \cite{das2018graph} propose to match the vertexes and edges of the domains, and refine the pseudo labels by keeping the unlabelled samples away from the decision boundaries.

Deep learning methods enhance transferability by exploring the abstract representations that disentangle the exploratory factors of variants hidden in the data \cite{yosinski2014transferable}. Distribution alignment methods directly minimize the discrepancy between domains based on moment statistics directly. For the first-order statistic, maximum mean discrepancy (MMD) is a popular and effective metric. Deep Adaptation Network (DAN) \cite{long2019learning} exploits the multiple kernel variant of MMD to maximize test power and reduce the probability of Type \uppercase\expandafter{\romannumeral2} error jointly. Ren et al. \cite{ren2019learning} improved conditional MMD (CMMD) under the auto-encoder framework. For second-order moment statistics, Deep CORAL is a simple, yet effective, method that measures the distance between domains by their corresponding covariances \cite{sun2016deep}. Chen et al. \cite{chen2019joint} add instances- and centers-based loss functions to the Deep CORAL model to enhance the discriminability of the source domain. Inspired by GANs, lots of adversarial approaches with different purposes were developed \cite{ganin2016domain,ren2020domain}. Joint Adaptation Network (JAN) \cite{Long2017Deep} proposes a joint MMD distance and adopts adversarial training to make the domains more distinguishable. Based on the domain shared generator, Unsupervised Domain Adaptation with Similarity Learning (SimNet) \cite{pinheiro2018unsupervised} develops a classifier composed of the prototypes from different classes, then the target samples are classified by the most similar prototype.

Domain-specific and Task-specific methods aim to tackle the problem of compact representations in high-level layers. Chang et al. \cite{chang2019domain} assign different domains using distinct batch-normalization layers and shared feature extraction layers in the first stage, and then train the target classifier by gradually modifying pseudo labels. Ding et al. \cite{ding2019deep} propose an end-to-end low-rank coding method via domain-specific dictionaries. Maximum Classifier Discrepancy (MCD) \cite{saito2018maximum} minimizes the maximum discrepancy of classifiers adversarially to form a tight classification boundary. Recent research suggests that discriminability plays a crucial role in the formation of class distributions~\cite{long2018conditional,chen2019transferability,ren2018generalized}. Conditional Domain Adversarial Network (CDAN) \cite{long2018conditional} encodes target predictions into deep features, then models the joint distributions of features and labels. Batch Spectral Penalization (BSP) \cite{chen2019transferability} builds a spectral penalty to enhance transferability while keeping the main discriminability.

\subsection{Partial UDA Methods}\label{sec:PartialDA}

The most typical characterization of the partial UDA setting is $\mathcal{C}^t \subset \mathcal{C}^s$, thus those two domains partially overlap. The outlier classes refer to the categories that are not shared by the two domains, i.e., $\mathcal{C}^s / \mathcal{C}^t$. If target samples are aligned to the outlier categories, they are likely to be misclassified.

Weight-based methods assign lower weights to the less transferable samples or classes (i.e., the outlier samples or classes), which leads to a declining influence of negative transfer. Partial Adversarial Domain Adaptation (PADA) \cite{cao2018PADA} is a class-level weighting method that takes the target predictions as outlier class measurements. As the probability values of the outlier classes are supposed to be significantly smaller than the shared classes, the source classes are weighted by the mean value of target predictions to reduce the misaligned samples. Importance Weighted Adversarial Nets (IWAN) \cite{zhang2018importance} introduces an auxiliary domain classifier to estimate the probability that the source sample belongs to the outlier classes. The sample with larger probability will be assigned a smaller weight during domain confusion. Example Transfer Network (ETN) \cite{cao2019learning} extends the idea of IWAN by introducing an auxiliary label predictor with leaky-softmax activation. Selective Adversarial Networks (SAN) \cite{cao2018SAN} proposes class-specific domain discriminators to circumvent negative transfer. Adversarial learning is guided by both instance-level and class-level weights from the target predictions.

Some recent methods revisit the UDA problem from the perspective of feature transferability \cite{chen2019transferability,xu2019larger}. They investigate the discriminability or transferability based on the singular values or norms of learned features. Adaptive Feature Norm (AFN) \cite{xu2019larger} builds a unified computation for the vanilla and partial UDA problems by progressively matching the feature norms of two domains to a large value. It suggests that the larger norm regions are more suitable for safe transfers.

\begin{figure*}[t]
\centering
\includegraphics[width=0.95\linewidth]{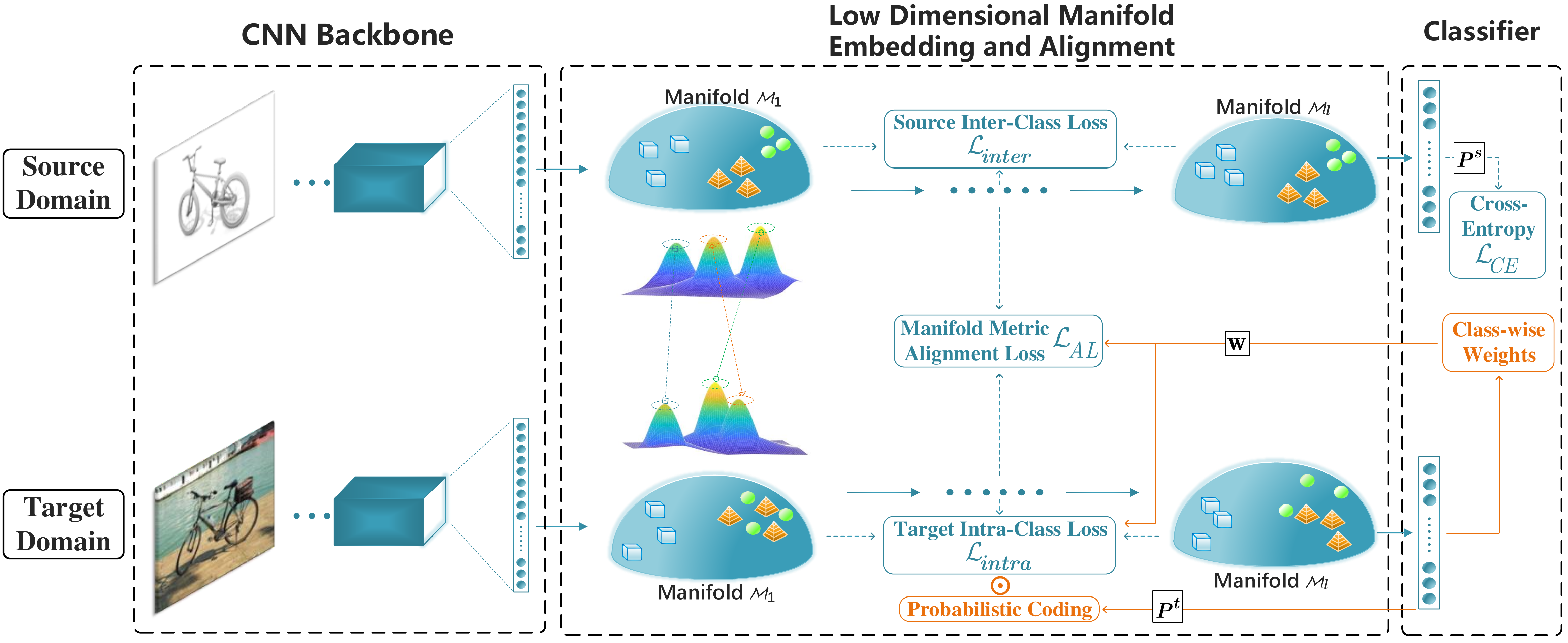}
   \caption{Overview of the proposed multilayer Riemannian manifolds embedding and alignment network. A CNN-based feature extractor is adopted to learn the common representations of both domains. The discriminative information are transferred via the Riemannian manifold layers, where fully connected layers are equipped with the proposed proposed soft discriminant criterion and manifold metric domain alignment. Best viewed in color.}
\label{fig:network}
\vspace{-10pt}
\end{figure*}

\section{Multilayer Riemannian Manifold Embedding and Alignment}\label{sec:ProposedRimennian}
In this section, we propose the DMP framework. The motivation is presented in Section \ref{subsec:background&motivation}. An overview of DMP and the network architecture are shown in Section \ref{subsec:ProposalOverview&Net}. In Section \ref{subsec:Discriminant}, we propose the global discriminant criterion for manifold embedding. The manifold metric alignment is presented in Section \ref{subsec:ManifoldAlignment}.

\subsection{Motivations}\label{subsec:background&motivation}
The Riemannian manifold $\mathcal{M}$ usually consists of objects such as a linear subspace, an affine/convex hull, and a symmetric positive definite (SPD) matrix \cite{huang2017cross}. From the perspective of discriminative embedding, graph-based criteria \cite{yan2007graph} are widely adopted in manifold learning and domain adaptation. These methods establish an instances-based connection graph or similarity graph to construct a separable space \cite{shekhar2015coupled,pinheiro2018unsupervised}. One of the most common assumptions in UDA based on a statistical distribution, the alignment based on covariance matrices, that lie on the Riemannian manifold, equips the domain with the manifold and statistical properties. Motivated by previous attempts \cite{huang2017cross,ren2019discriminative}, our work aims to embed a graph-based discriminant criterion to the target domain, and align the source and target domain based on the manifold assumption.

Given a feature matrix ${\bf X} \in \mathbb{R}^{d\times n}$ and its mean vector $\bar{{\bf x}}\in\mathbb{R}^{d}$, where $d$ denotes the dimension of features and $n$ the sample size. Denote by $S$ the input space (e.g., Euclidean space and Hilbert space), manifold learning aims to learn a nonlinear mapping
\begin{equation*}
f:~~ S ~ \rightarrow ~ \mathcal{M},
\end{equation*}
where $\mathcal{M}$ is the low-dimensional embedding manifold. Based on the SPD representation setting, the image of a given covariance matrix ${\bf C(X)}=\frac{1}{n-1} ({\bf X}-\bar{{\bf x}}{\bf 1}^T_n)({\bf X}-\bar{{\bf x}}{\bf 1}^T_n)^T \in\mathbb{R}^{d\times d}$ is a lower-order SPD matrix ${\bf C}' = f({\bf C}) \in\mathbb{R}^{d'\times d'}$, where ${\bf 1}_n$ is an $n$-dimensional vector with all elements equal to 1 and $(\cdot)^T$ is the transpose operation. ${\bf C}'$ can be decomposed as the inner product of a lower-order matrix ${\bf X}'$, i.e., ${\bf C}'={\bf X}'{\bf X}'^T$. Learning of the projector $f$ can be deduced to find a nonlinear transformation
\begin{equation*}
g:~ {\bf X} ~ \mapsto g({\bf X}),
\end{equation*}
where $g({\bf X})$ is the approximation of ${\bf X}'$. That is, the image of mapping $f$ can be approximated by the inner product of $g({\bf X})$, i.e., $f({\bf C}) \approx g({\bf X})g({\bf X})^{T}$.

For domain adaptation, the original source and target domains can be taken as two Euclidean spaces, where the discriminant information is relatively inadequate. Thus the latent manifolds, i.e., $\mathcal{M}^s$ of the source and $\mathcal{M}^t$ target, should be compact, representative and discriminative. Compared with the images, the manifold embedding features are supposed to contain more task-specific information (i.e., class information) and be domain indistinguishable. In DMP, we extract the task-specific information from the images by equipping the discriminant criterion with the predictive information, while removing most of the domain information (e.g., styles and views) by aligning the domains based on the manifold metrics.

\subsection{Low-Dimensional Manifold Layers}\label{subsec:ProposalOverview&Net}

We now focus on the nonlinear transformation $g$ for the input features ${\bf X}$. In general, CNNs are used to obtain the projection $g$. Figure \ref{fig:network} shows the detailed network architecture of the proposed method. Let $\bm{\Theta}$ be the parameters of the networks. To explore the latent Riemannian representations of the raw features, the output features of the CNN backbone are sent into progressive low-dimensional manifold layers, as shown in Figure \ref{fig:network}. Since there are natural geometric differences between the raw and embedding spaces, a multilayer scheme is adopted to reduce the dimension of features progressively.

The Riemannian manifold layers $\{\mathcal{M}_l|l=1,2,\ldots,L\}$ are achieved via fully connected layers. The CNNs and Riemannian manifold layers are shared by both domains so that the common projections can be exploited to map the two domains to a shared low-dimensional space. Therefore, any manifold layer $\mathcal{M}_i$ should have the following properties.
\begin{itemize}
\item Discriminative Structure: The intra-class samples of the target domain are compact, while the inter-class samples are separable.
\item Consistent Structure: The source and target domains are aligned with the manifold metrics. Then the domain discrepancy can be represented as the distance between two submanifolds on $\mathcal{M}_i$, and minimized based on the defined manifold metrics (e.g., Grassmann distance, affine Grassmann distance, Log-Euclidean metric or manifold principal angle similarity).
\end{itemize}

We model the properties by loss terms $\mathcal{L}_{DS}$ and $\mathcal{L}_{AL}$, which will be mathematically formulated later. To satisfy the first property, two similarity-based criteria are explored, i.e., the source discriminant inter-class loss and the target discriminant intra-class loss in Figure \ref{fig:network}. Specifically, the target intra-class loss is used to enlarge the similarities between the target samples and their corresponding source class-wise centers, while the source inter-class loss is used to find a balanced geometric structure of the source class-wise mean vectors. The discriminative structure loss is denoted by
\begin{equation*}
\mathcal{L}_{DS} = \sum_{l} (\mathcal{L}^l_{inter} + \mathcal{L}^l_{intra}),
\end{equation*}
where $\mathcal{L}^l_{inter}$ and $\mathcal{L}^l_{inter}$ are the similarity-based loss functions of the $l$-th Riemannian manifold layer $\mathcal{M}_l$.

The second property is satisfied by the manifold metric alignment loss in Figure \ref{fig:network}. The overall alignment loss can be written as
\begin{equation*}
\mathcal{L}_{AL} = \sum_{l} \mathcal{L}_{align}^{l},
\end{equation*}
where $\mathcal{L}_{align}^{l}$ is the alignment loss of $\mathcal{M}_l$.

Finally, the proposed objective function is
\begin{equation}\label{eq:Objective}
\mathop {\min}\limits_{\bm{\Theta}} \mathcal{L} = \mathcal{L}_{CE} + \lambda_1 \mathcal{L}_{DS}  + \lambda_2 \mathcal{L}_{AL},
\end{equation}
where $\mathcal{L}_{CE}$ is the cross-entropy loss of the classifier on the source domain and \{$\lambda_1,\lambda_2$\} are penalty parameters. The introduction and derivation of these loss terms are presented in the following sections.

\begin{figure*}[t]
\centering
\includegraphics[width=0.94\linewidth]{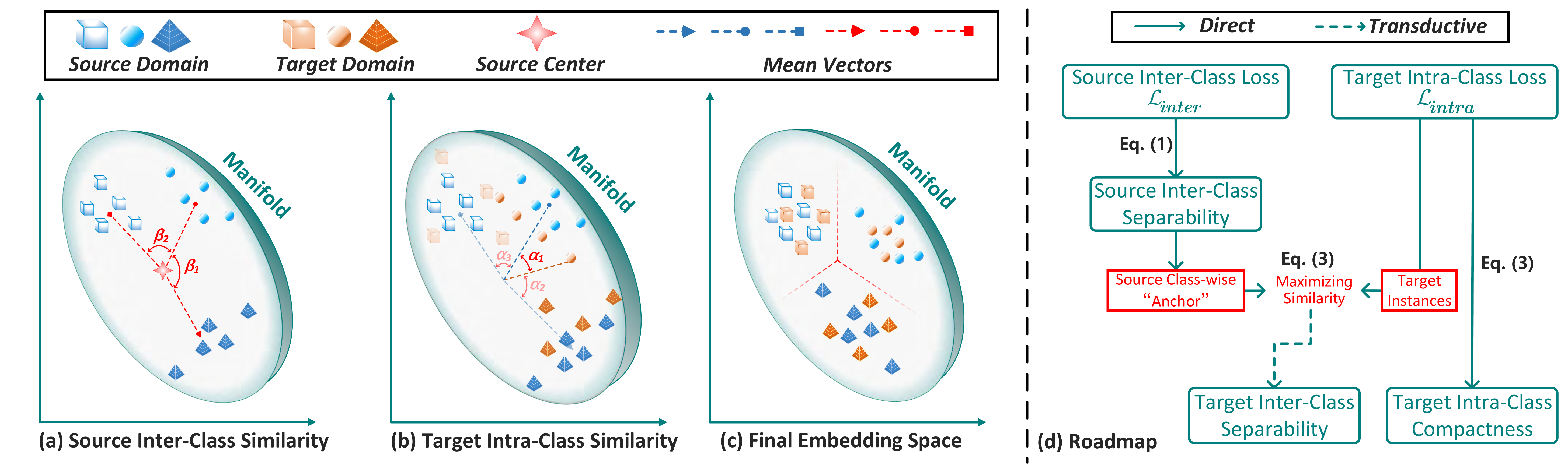}
   \caption{Illustration of the discriminative structure learning framework. (a) The source inter-class similarity forms a separable space for the source samples. (b) The target intra-class similarity constructs a compact space for the samples from the same category. (c) The final embedding space, where the target domain is discriminative. (d) The roadmap for achieving the discriminative property on the target domain. The target intra-class compactness is implemented by $\mathcal{L}_{intra}$ directly and the target inter-class separability is reached by $\mathcal{L}_{inter}$ transductively. Best viewed in color.}
\label{fig:discriminant}
\vspace{-10pt}
\end{figure*}

\subsection{Discriminative Structure Learning}\label{subsec:Discriminant}
In this section, we describe how to embed the discriminative structure into the manifold layers. The main idea is shown in Figure \ref{fig:discriminant}. Since there exists a distribution discrepancy between different domains (e.g., Figure \ref{fig:discriminant}(b)), models that build a discriminant criterion only on the source domain are less discriminative on the target domain. The discriminant learning process will be error-prone if only the target uncertain information is used. If discriminability is required for both domains, the ability of model to generalize may be decreased. To relax the constraint, we propose to focus on the inter-class separability of the source domain and the soft intra-class compactness of the target domain. This uses prototypes (e.g., center vectors), which have been shown to be effective and robust to the domain shift problem \cite{pinheiro2018unsupervised,pan2019transferrable}, as an intermediary during discriminant learning. As illustrated in Figure \ref{fig:discriminant}(d), the proposed method achieves the discriminative property on the target domain transductively.

Without loss of generality, we only formulate the loss terms in the $l$-th Riemannian manifold layer $\mathcal{M}_l$. Let ${\bf H}^s_l=[{\bf h}^s_{l,1},{\bf h}^s_{l,2},\ldots,{\bf h}^s_{l,n^s}]\in \mathbb{R}^{d_l\times n^s}$ and ${\bf H}^t_{l}\in \mathbb{R}^{d_l\times n^t}$ be the feature matrices of $\mathcal{M}_l$. Since class centers of the source domain are used in both loss terms, the source mean vector ${\bf \bar{h}}^s_l\in\mathbb{R}^{d_l}$ and source class-wise mean matrix ${\bf \bar{H}}^s_l\in\mathbb{R}^{d_l\times c}$ are computed, where $c=|\mathcal{C}^s|$ is the number of source categories. Let ${\bf P}^t = [{\bf p}^t_1,{\bf p}^t_2,\ldots,{\bf p}^t_{n^t}] \in \mathbb{R}^{c\times n^t}$ and ${\bf P}^s \in \mathbb{R}^{c\times n^s}$ be the softmax predictions of the target and source samples from the classifier layer, respectively, and $y_i\in\{1,2,\ldots,c\}$ be the ground-truth label of the $i$-th source sample. The lower case letter with subscript $ij$ (e.g., $a_{ij}$) represents the $(i,j)$ entry of its corresponding matrix (e.g., ${\bf A}$).

\subsubsection{Source Inter-Class Similarity}\label{subsubsec:SourceInter}
To learn a separable geometric structure of the class distribution, the similarity measurement is adopted here, which is also shown in Figure \ref{fig:discriminant}(a). Rather than computing the similarities between the class-wise centers and the source center directly, we consider the class-wise centers as follows
\begin{equation}\label{eq:CentralizedClassMeans}
{\bf \hat{H}}^s_l \triangleq  {\bf \bar{H}}^s_l - {\bf \bar{h}}^s_l{\bf 1}_c^T.
\end{equation}
We call ${\bf \hat{H}}^s_l\in \mathbb{R}^{d_l \times c}$ the centralized class means hereafter. If the columns of ${\bf \hat{H}}^s_l$ are normalized with the $\ell_2$ norm, the between-class cosine matrix is derived as
\begin{equation*}
{\bf S}^b_{l} = {\bf \hat{H}}^{s^T}_l{\bf \hat{H}}^s_l,
\end{equation*}
where $s^b_{l,ij}={\bf \hat{h}}^{s^T}_{l,i} {\bf \hat{h}}^{s}_{l,j}$ indicates the similarity between $i$-th class and $j$-th class. Then the separable structure is reached by maximizing the dissimilarities between the centralized class mean vectors. Equivalently, it can be achieved by minimizing the following inter-class loss:
\begin{equation}\label{eq:InterLoss}
\mathcal{L}^l_{inter}({\bf H}^{s}_l) = \frac{2}{c(c-1)}\sum_{i<j} s^b_{l,ij}.
\end{equation}

Let us take Figure \ref{fig:discriminant}(a) as an example. There is a 2-dimensional space with 3 classes. Let \{1,2,3\} be the labels of ``Ball'', ``Pyramid'' and ``Cube'', respectively. Under this situation, $ s^b_{l,12}$ and $s^b_{l,13}$ are depicted as $\cos(\beta_1)$ and $\cos(\beta_2)$, respectively. According to the goal of Eq. \eqref{eq:InterLoss} and ignoring the constraints, the optimal solution occurs at $\beta_1=\beta_2=\frac{2}{3}\pi$, and the minimal $\mathcal{L}^l_{inter}$ equals to $-\frac{1}{2}$ (which can also be seen as the lower bound of constrained scenarios).

\subsubsection{Target Intra-Class Similarity}\label{subsubsec:TargetIntra}
Since there are no labels on the target domain, discriminant learning is facilitated by the soft labels ${\bf P}^t$ (i.e., the output of the softmax layer). Since ${\bf P}^t$ can be regarded as the confidence or probability of classification, the predictions are used to weight the importance, or confidence, of the supervised information provided by the soft labels. Similarly, assuming the columns of ${\bf \bar{H}}^s_l$ and ${\bf H}^t_{l}$ have unit length, the similarities under all classification cases can be written as
\begin{equation}\label{eq:TargetSimilarity}
{\bf S}^w_{l} = {\bf \bar{H}}^{s^T}_l {\bf H}^t_l.
\end{equation}
Note that the centers of the source classes are used instead of those of the target. The main reasons are that the inter-class structure learned from the source domain can be transduced to the target domain and that the source class centers computed from ground-truth labels are more reliable. Because there is so much uncertainty when pseudo labels are used straightforwardly on the target domain, we establish a probabilistic discriminative criterion to make use of most of the information provided by the soft labels. Intuitively, ${\bf P}^t$ is a naturally choice for the probabilistical weighting model. Then the probabilistic intra-class loss is formulated as
\begin{equation}\label{eq:IntraLoss}
\mathcal{L}^l_{intra}({\bf H}^t_{l}, {\bf P}^t) = -\frac{1}{n^t c}\sum_{i=1}^{c} \sum_{j=1}^{n^t} p^t_{ij} s^w_{l,ij}.
\end{equation}

However, there is much noise in ${\bf P}^t$, and its values are very small. Empirically, ${\bf p}_i^t$ tends to be the a one-hot vectors if the softmax classifier is convergent. As truncation is an efficient way for denoising, we develop a Top-$k$ preserving scheme for the truncated intra-class loss. Let $V_j = \{(i,j)|i=v_{1j},v_{2j},\ldots,v_{kj}\}$ be the index set of the $k$-largest elements in ${\bf p}^t_j$, $j=1,2,\ldots,n^t$. Then an indicator matrix is defined as
\begin{equation*}
\scalebox{1.3}{$\chi$}_{ij}  = \left\{
\begin{array}{cc}
1,&~~ (i,j)\in V_j \\
0,&~~ (i,j)\notin V_j
\end{array}
\right..
\end{equation*}
Finally, the intra-class loss is modified by the truncated matrix \scalebox{1.3}{$\chi$} and written as
\begin{equation}\label{eq:IntraLoss&Truncated}
\mathcal{L}^l_{intra}({\bf H}^t_{l}, {\bf P}^t) = -\frac{1}{n^t k}\sum_{i=1}^{c} \sum_{j=1}^{n^t} \scalebox{1.3}{$\chi$}_{ij} p^t_{ij} s^w_{l,ij}.
\end{equation}

In conclusion, the two proposed loss terms build a probabilistic discriminant criterion on the target domain. The ground-truth labels on the source domain provide a reliable separable structure directly, where the intra-class structure is not required. The target samples are attached to the corresponding source class-wise centers via soft labels. As shown in Figure \ref{fig:discriminant}(c) and \ref{fig:discriminant}(d), the intra-class relationship on the source domain does not change much while the discriminative property of the target domain is satisfied. The motivation behind using the cosine function is that the Euclidean distance may fail to learn a separate inter-class structure when some class centers nearly overlap, e.g., the case that $\beta_2 \approx 0$ in Figure \ref{fig:discriminant}(a). However, the cosine-based metric learning measures the similarity on the unit hypersphere. It can achieve the discriminative feature structure from geometric aspect, i.e., large inter-class angles and small intra-class angles.

\subsubsection{Global Structure Learning}\label{subsubsec:GlaobalStruct}
For the batch-wise training manner in deep learning models, training batch sizes of the source and target domains are set as $b_s$, i.e., $n^s=n^t=b_s$. The complete relation graph between the instances is time and memory consuming to obtain in deep networks. Moreover, the mean statistics ${\bf \bar{h}}^s_l$ and ${\bf \bar{H}}^s_l$ computed from the batch data are unable to reflect the complete categorical information, because the class number in the batch is often smaller than $c$. The direct application of classical graph embedding may be misled by some extreme local distributions, which will lead to a suboptimal solution.

Supposing that the geometry of the manifold does not change drastically after several updates, we build two \textit{anchors} in the whole data space to acquire the global information. We propose to fix the \textit{anchors} in each batch iteration and update them after every epoch or several iterations. Specifically, the \textit{anchors}, i.e., ${\bf \bar{h}}^s_l$ in Eq. \eqref{eq:CentralizedClassMeans} and ${\bf \bar{H}}^{s}_l$ in Eq. \eqref{eq:TargetSimilarity}, are dynamically updated. Note that the \textit{anchors} are treated as constants in optimization. The variables, i.e., ${\bf \bar{H}}^{s}_l$ in Eq. \eqref{eq:CentralizedClassMeans} and ${\bf H}^t_l$ in Eq. \eqref{eq:TargetSimilarity}, are obtained from batch data. If there are no samples of the $i$-th class in the current batch, then the corresponding class center ${\bf \bar{h}}^{s}_{l,i}$ is an all-zero vector. The inter-class loss is strongly supervised by source labels at the beginning, while the intra-class loss, facilitated by soft labels, is used after a certain number of iterations/epochs.

\subsection{Manifold Metric Alignment}\label{subsec:ManifoldAlignment}
A manifold metric alignment method is proposed to satisfy domain consistency. The second-order moment statistic is an important tool to represent a manifold $\mathcal{M}$. Therefore, the alignment based on covariance not only meets the requirement of the manifold assumption, but also possesses useful statistical properties, such as the distribution assumption.

Let ${\bf C}^s_l$ and ${\bf C}^t_l$ be the covariance matrices of the source and target domains computed from batch-wise features, respectively. Assume $\mathcal{M}^s_l$ and $\mathcal{M}^t_l$ are two submanifolds of $\mathcal{M}_l$, which are represented by their corresponding covariance matrices. Before the alignment process, these two submanifolds may partially overlap, and our goal is to minimize the discrepancy under the metric, $\mathcal{M}_l$. In general, the manifold metric alignment loss is expressed as
\begin{equation*}\label{eq:AlignLoss}
\mathcal{L}_{align}^{l}({\bf H}^s_{l},{\bf H}^t_{l}) \triangleq dist(\mathcal{M}^s_l,\mathcal{M}^t_l) = d_{\mathcal{M}}({\bf C}^s_l,{\bf C}^t_l),
\end{equation*}
where $d_{\mathcal{M}}(\cdot,\cdot)$ is the manifold metric to be defined.

\subsubsection{Grassmann Manifold}\label{subsubsec:GrassDist}
The Grassmann manifold \cite{edelman1998geometry} is a well-known type of Riemannian manifold. It is a projected subspace $\mathbb{R}^{d'_l}$ deduced from the originally high-dimensional space $\mathbb{R}^{d_l}$, $d'_l<d_l$. Thus, two submanifolds $\mathcal{M}^s_l$ and $\mathcal{M}^t_l$ lying on the Grassmann manifold $\mathcal{M}_l$ are represented as two individual points. The distance between these two points is measured by the discrepancy between the their projection orthogonal bases ${\bf U}^s_l$ and ${\bf U}^t_l$ which can be obtained from the Singular Value Decomposition (SVD) of the covariance matrices ${\bf C}^s_l$ and ${\bf C}^t_l$, respectively, i.e.,
\begin{equation}\label{eq:GrassMetric}
d^{G}_{\mathcal{M}}({\bf C}^s_l,{\bf C}^t_l) = \frac{1}{d^2_l}\| {\bf U}^s_l {\bf U}^{s^T}_l - {\bf U}^t_l {\bf U}^{t^T}_l \|^2_F,
\end{equation}
where $\| \cdot \|_F$ is the Frobenius norm.

As the dimension $d'_l$ is required in the Grassmann distance, we establish a theoretical error bound for the selection of $d'_l$. Denoting the covariance of a given distribution $D$ by ${\bf C}$, and the covariance drawn i.i.d. from $D$ with sample size $n$ by $\tilde{\bf C}$. Then, Zwald et al. \cite{zwald2006convergence} give the following theorem.
\begin{theorem}\label{thm:U_Bound}
{\rm \textbf{\cite[Theorem 4]{zwald2006convergence}}}
Supposing that $\sup_{{\bf x}\in\mathcal{X}} \| {\bf x} \| \leq M$, where $\mathcal{X}$ is the measurable space where variable ${\bf x}$ take its value. Let ${\bf U}^{d'}_{\bf C}$ and ${\bf U}^{d'}_{\tilde{{\bf C}}}$ be the orthogonal projectors of the subspaces spanned by the first $d'$ eigenvectors of ${\bf C}$ and $\tilde{\bf C}$, respectively. Let $\lambda_1>\lambda_2>\cdots>\lambda_{d'}>\lambda_{d'+1}\geq0$ be the first $d'+1$ eigenvalues of ${\bf C}$, then for any
$n \geq\left(\frac{4 M}{\lambda_{d'}-\lambda_{d'+1}}\left(1+\sqrt{\frac{\ln (1 / \delta)}{2}}\right)\right)^{2}$
with probability at least $1-\delta$ we have:
\begin{equation}\label{eq:U_Bound}
\|{\bf U}^{d'}_{\bf C}-{\bf U}^{d'}_{\tilde{{\bf C}}}\| \leq \frac{4 M}{\sqrt{n}\left(\lambda_{d'}-\lambda_{d'+1}\right)}\left(1+\sqrt{\frac{\ln (1 / \delta)}{2}}\right).
\end{equation}
\end{theorem}
This theorem shows the relation between the error and $d'$. Denote the right side of Eq. \eqref{eq:U_Bound} as $\frac{E(\delta)}{\lambda_{d'}-\lambda_{d'+1}}$. To extend the inequality to the Grassmann distance, we derive following lemma.
\begin{lemma}\label{thm:UU_Bound}
Based on the condition in Theorem \ref{thm:U_Bound}, we have
\begin{equation*}
\|{\bf U}^{d'}_{\bf C}{\bf U}^{d'^T}_{\bf C}-{\bf U}^{d'}_{\tilde{{\bf C}}}{\bf U}^{d'^T}_{\tilde{{\bf C}}}\|_F \leq 2\sqrt{2}E(\delta) \frac{\sqrt{d'}}{\lambda_{d'}-\lambda_{d'+1}}
\end{equation*}
with probability at least $1-\delta$.
\end{lemma}

Based on Lemma \ref{thm:UU_Bound}, the following theorem gives a error bound of $d_{\mathcal{M}}({\bf C}^s,{\bf C}^t)$ w.r.t. its $n$ sample approximation $d_{\tilde{\mathcal{M}}}(\tilde{\bf C}^s,\tilde{\bf C}^t)$.
\begin{theorem}\label{thm:Grass_Bound}
Assuming the condition in Theorem \ref{thm:U_Bound} is specified by domains. Specifically, $\lambda_i^s$ and $\lambda_i^t$ denote the $i$-th largest eigenvalue of the domain-specific covariance matrices ${\bf C}_s$ and ${\bf C}_t$, respectively. Denote the error index by:
\begin{equation*}
e^{G}(d')=\frac{\sqrt{d'}}{\lambda_{d'}^s-\lambda_{d'+1}^s} + \frac{\sqrt{d'}}{\lambda_{d'}^t-\lambda_{d'+1}^t}.
\end{equation*}
Then the following error bound holds with probability at least $1-\delta$:
\begin{equation*}
|d^{G}_{\mathcal{M}}({\bf C}^s,{\bf C}^t)-d^{G}_{\tilde{\mathcal{M}}}(\tilde{\bf C}^s,\tilde{\bf C}^t)| \leq 2\sqrt{2}E(\delta)e^{G}(d').
\end{equation*}
\end{theorem}

Theorem \ref{thm:Grass_Bound} suggests that the upper bound of the error is proportional to $e(d')$. It means that we should search for the maximal gap between the continuous eigenvalues with consideration of the inflation factor $\sqrt{d'}$. As, in a batch learning setting, the batch size $b_s$ is usually smaller than $d$, thus $d'$ only needs to be searched in $\{1,2,\ldots,b_s-1\}$. The proofs are provided in Section S.1 and Section S.2 of the supplementary material, respectively.

\subsubsection{Affine Grassmann Manifold}\label{subsubsec:AffineGrassDist}
The affine Grassmann manifold is a smooth manifold that consists of all $d'$-dimensional affine subspaces in $\mathbb{R}^d$, thus it is also called the Grassmannian of the affine subspaces \cite{lim2019numerical}. For any matrix, its representation on affine Grassmann manifold is the affine combination of orthonormal $d'$-frames ${\bf U}$ with displacement $\bm{\mu}$, where $\bm{\mu}$ is the mean of the matrix. In \cite{huang2017cross}, Huang et al. extended the similarity function on the affine Grassmann manifold to a distance metric. The distance between the corresponding representations of ${\bf C}^s_l$ and ${\bf C}^t_l$ on the affine Grassmann manifold is defined as
\begin{equation}\label{eq:AffineGrassMetric}
\begin{split}
d^{AG}_{\mathcal{M}}({\bf C}^s_l, & {\bf C}^t_l) =  \frac{1}{d^2_l} \bigg( \| {\bf U}^s_l {\bf U}^{s^T}_l - {\bf U}^t_l {\bf U}^{t^T}_l \|_F   \\
&  + \| ( {\bf I} - {\bf U}^s_l {\bf U}^{s^T}_l ) \bm{\mu}^s_l - ( {\bf I} - {\bf U}^t_l {\bf U}^{t^T}_l ) \bm{\mu}^t_l \|_2 \bigg) ,
\end{split}
\end{equation}
where ${\bf I}$ is the identity matrix, $\bm{\mu}^s_l$ and $\bm{\mu}^t_l$ are the mean vectors of covariances ${\bf C}^s_l$ and ${\bf C}^t_l$, respectively. Analogously, we also derive an error bound for the empirical estimation of the affine Grassmann metric.
\begin{theorem}\label{thm:AffineGrass_Bound}
Assuming the condition in Theorem \ref{thm:U_Bound} is specified by domains. Denote $\lambda_i^s$ and $\lambda_i^t$ as the $i$-th largest eigenvalue of the domain-specific covariance matrices ${\bf C}_s$ and ${\bf C}_t$, respectively. Let
\begin{equation*}
e^{AG}(d')=\frac{\sqrt{d'} \| \tbfmu^s \|_2}{\lambda^s_{d'}-\lambda^s_{d'+1}} + \frac{\sqrt{d'} \| \tbfmu^t \|_2}{\lambda^t_{d'}-\lambda^t_{d'+1}}
\end{equation*}
be the error index. Then the following error bound holds with probability at least $1-\delta$:
\begin{equation*}
|d^{AG}_{\mathcal{M}}({\bf C}^s,{\bf C}^t)-d^{AG}_{\tilde{\mathcal{M}}}(\tilde{\bf C}^s,\tilde{\bf C}^t)| \leq 2\sqrt{2}E(\delta) \left( \frac{\sqrt{2d}}{4} +  e^{AG}(d') \right).
\end{equation*}
\end{theorem}

Compared with the error index of the Grassmann distance $e^G$, the error terms of different domains in $e^{AG}$ are weighted by the $\ell_2$-norms of their mean vectors. This conclusion is consistent with the definition of an affine Grassmann manifold as the mean vectors are the coefficients of the affine transformation in Eq.~\eqref{eq:AffineGrassMetric}. The proof is provided in Section S.3 of the supplementary material.

\subsubsection{SPD manifold}\label{subsubsec:LogEucDist}
Since the space of SPD matrices can be taken as a special type of manifold called the SPD manifold, theoretical research has been conducted to explore the non-Euclidean geometry of SPD manifolds \cite{pennec2006riemannian,arsigny2007geometric}. The affine invariant framework \cite{pennec2006riemannian}, called the Affine Invariant Riemannian Metric, analyzed the SPD manifold based on the inner product. Arsigny et al. \cite{arsigny2007geometric} proposed a novel framework called Log-Euclidean which overcame the computational drawback of the affine invariant framework. Let ${\bf C}^{s/t}_l={\bf U}^{s/t}_l {\bf D}^{s/t}_l {\bf U}^{s/t^T}_l$ be the eigendecomposition. Note that the Log-Euclidean metric requires all orthonormal basis and assumes the matrices are positive definite. Thus, we regularize the covariance matrices as ${\bf C}+\varepsilon \mathbf{I}$. The Log-Euclidean metric between ${\bf C}^s_l$ and ${\bf C}^t_l$ is defined as the distance between their matrix logarithms, i.e.,
\begin{equation}\label{eq:LogEucMetric}
d^{LE}_{\mathcal{M}}({\bf C}^s_l,{\bf C}^t_l) = \frac{1}{d^2_l}\| \log{( {\bf C}^s_l )} - \log{( {\bf C}^t_l )} \|_F,
\end{equation}
where $\log{( {\bf C}^{s/t}_l )}={\bf U}^{s/t}_l \log{( {\bf D}^{s/t}_l )} {\bf U}^{s/t^T}_l$.

\section{Extension for Partial UDA}\label{sec:UniformExtension}
To build a unified algorithm for the vanilla and partial UDA problems, we introduce the weight-based extension for the DMP method in this section.

\subsection{The Weighting Strategy}\label{subsec:WeightedDMP}
The essential problem for this extension is how can the negative information transduced from the outlier classes be mitigated. Under the partial UDA setting, the number of shared classes (i.e., target classes) is not larger than that of the source classes, i.e., $|\mathcal{C}^t| \leq |\mathcal{C}^s|$. Since some shared classes are similar to the outlier classes in appearance (e.g., the bus in shared classes and car in outlier classes), the vanilla methods that align the entire feature spaces of two domains will probably lead to negative transfers. Also, the discriminant learning supervised by the soft labels should ignore the outlier classes, since intra-class compactness of the outlier classes is unnecessary and error-prone. To tackle these problems, we propose to learn the discriminative embedding and alignment with the class-wise weights ${\bf w}=(w_1,w_2,\ldots,w_c)^T$. It aims to assign the potential outlier classes with smaller weights during the learning process, then the model will be less sensitive to the outlier samples.

First, the discriminant criterion is modified to loosen the intra-class compactness constraints in the outlier classes. This modification preserves the intrinsic structure of the target domain and enhances discriminant learning in shared classes. The weighted intra-class loss is derived as
\begin{equation}\label{eq:WeightedIntraLoss}
\mathcal{L}^l_{intra}({\bf H}^t_{l}, {\bf P}^t) = -\frac{1}{n^t k}\sum_{i=1}^{c} w_i \sum_{j=1}^{n^t} \scalebox{1.3}{$\chi$}_{ij} p^t_{ij} s^w_{l,ij}.
\end{equation}
The inter-class loss remains the same regardless of the UDA settings, since separabilities between any two classes are equally important to the final classification. The weighted discriminant criterion tends to align target instances to the class with high contribution $w_i$ for the partial setting and treats the classes equally for the vanilla setting.

For the domain alignment, the target domain is supposed to be partially aligned to the source domain under the partial setting. Denoting the source manifold based on the shared classes by $\mathcal{M}^{s'}$, the optimal domain discrepancy is measured by $dist(\mathcal{M}^{s'},\mathcal{M}^{t})$. Unfortunately, the manifold $\mathcal{M}^{s'}$ cannot be obtained, because the shared classes are unknown. To reduce the impact of outlier samples, the source feature matrix ${\bf H}^s$ is weighted by ${\bf w}$ as
\begin{equation*}
{\bf h}^{s'}_i = w_{y_i}{\bf h}^s_i, ~~~~i=1,2,\ldots,n^s.
\end{equation*}
Then, the weighted covariance ${\bf C}^{s'}$ can be computed from the weighted feature matrix ${\bf H}^{s'}$. The weighted alignment loss of the $l$-th manifold layer is described as
\begin{equation}\label{eq:WeightedAlignLoss}
\mathcal{L}_{align}^{l}({\bf H}^s_{l}, {\bf H}^t_{l}) = dist(\mathcal{M}^{s'}_l,\mathcal{M}^t_l) \approx d_{\mathcal{M}}({\bf C}^{s'}_l,{\bf C}^t_l).
\end{equation}
The weighted covariance ${\bf C}^{s'}$ will well characterize the manifold $\mathcal{M}^{s'}$ if the weight ${\bf w}$ is reliable, and the domain alignment will match the source and target domains partially to reduce the negative transfer.

\begin{algorithm}[!t]
\caption {DMP Method}\label{alg:DMP}
\begin{algorithmic}[1]
\REQUIRE {Source data $\mathcal{X}^s$, Source labels ${\bf y}$, Target data $\mathcal{X}^t$, Batch-size $b_s$, Maximum iteration $T_{max}$, Update iteration $T_{up}$, Learning rate $\lambda$, Parameter $\lambda_1$ and $\lambda_2$;}
\ENSURE {Learned network $\bm{\Theta}$, Target prediction $\hat{\bf y}^t$;}\\
\STATE Initialize the network parameter $\bm{\Theta}$; \\
\% \textit{Training Stage}\\
\FOR {$t=1,2,\ldots,T_{max}$}
\IF {$(t \bmod T_{up}) = 0$}
\STATE Forward propagate entire $\mathcal{X}^s$ without gradients; compute the source centers ${\bf \bar{H}}^s_{l}$, ${\bf \bar{h}}^s_{l}$ ($l=l_m,\ldots, L$) and class weight vector $\hat{{\bf w}}$ from Eq.~\eqref{eq:ClassWeight};
\ENDIF
\STATE Random select and forward propagate $b_s$ samples from $\mathcal{X}^t$ and $\mathcal{X}^s$ (with label ${\bf y}$), respectively;
\STATE Compute the objective function from Eq.~\eqref{eq:Objective}.
\FOR {$l=1,2\ldots,L$}
\STATE Compute the gradient $\nabla \bm{\Theta}_l$ with parameters $\lambda_1$ and $\lambda_2$ from Eq.~\eqref{eq:TotalGradient};
\STATE Update the network parameter: $\bm{\Theta}_l \leftarrow \bm{\Theta}_l - \lambda \nabla \bm{\Theta}_l$;
\ENDFOR
\ENDFOR
\\
\% \textit{Testing Stage}\\
\FOR {$\mathcal{X}^t_j$ \textbf{in} $\mathcal{X}^t$}
\STATE Forward propagate $\mathcal{X}^t_j$ to obtain probability prediction ${\bf p}^t_j$;
\STATE Compute the predicted label as $\hat{y}^t_j = \mathop{\arg\max}_{i} p^t_{ij} $
\ENDFOR
\STATE Return the network parameter $\bm{\Theta}= \{ \bm{\Theta}_1, \bm{\Theta}_2, \ldots, \bm{\Theta}_L \}$ and the target prediction $\hat{\bf y}^t = [\hat{y}^t_1,\hat{y}^t_2,\ldots]^T$;
\end{algorithmic}
\end{algorithm}

As the target prediction matrix ${\bf P}^t$ reveals the contributions from different categories, it can be used to evaluate the probability that a certain class belongs to the shared classes. For the partial setting, the class-wise contribution vector $\bm{\pi} \in \mathbb{R}^c$ is computed as the mean value of the target prediction ${\bf P}^t$, i.e., $\bm{\pi} = (\sum_{i=1}^{n^t}{\bf p}_i^t)/{n^t}$. A smaller $\pi_i$ indicates that the $i$-th class is more likely to be the outlier classes. However, the weights are actually useless in the vanilla setting, so the general weight vector $\hbfw =(w_1,w_2,\ldots,w_c)^T$ is defined as
\begin{equation}\label{eq:ClassWeight}
\hat{w}_i = \left\{
\begin{array}{cc}
 \pi_i &,~~~~ \text{partial setting} \\
 1/c &,~~~~ \text{vanilla setting}
\end{array}
\right.,
~~i = 1,2,\ldots,c.
\end{equation}

The main advantages of our weighting strategy can be summarized as follows. 1) it assigns weights based on a constant mass since $\sum_i^c \hat{w}_i=1$, which gives a consistent interpretation of the weights under different settings; 2) it considers the weighted model from both instance-level (i.e., the weighted manifold alignment) and class-level (i.e., the weighted discriminant criterion). As the weight vector is computed from the mean value of predictions, it sometimes suffers from the class imbalance problem which is the disadvantage of this strategy.

In fact, the ideal $w_i$ should be $\frac{1}{c'}$ for the shared classes and 0 for the outlier classes. Denote the covariances computed from the estimated and ideal weights by ${\bf C}^{s'}_{\hat{\bf w}}$ and ${\bf C}^{s'}_{\bf w}$, respectively. The following theorem extends Theorem \ref{thm:Grass_Bound} to a weighting scheme.
\begin{theorem}\label{thm:Grass_Bound_partial}
Assuming the condition in Theorem \ref{thm:U_Bound} is specified by domains and $n^s=b_s\leq d$. Denote the error index under the partial setting by:
\begin{equation*}
e^{G}_{P}(d',\hbfw)= \alpha \|\bfw^2 - \hbfw^2\|_2 + 2\sqrt{2}E(\delta)e^{G}(d').
\end{equation*}
Then the following error bound:
\begin{equation*}
|d^{G}_{\mathcal{M}}({\bf C}^{s'}_{\bf w},{\bf C}^{t})-d^{G}_{\tilde{\mathcal{M}}}(\tilde{\bf C}^{s'}_{\hat{\bf w}},\tilde{\bf C}^{t})| \leq  e^G_{P}(d',\hbfw),
\end{equation*}
holds with probability at least $1-\delta$, where $\bfw^2$ and $\hbfw^2$ mean the element-wise square, and $\alpha(\hbfw)$ is a constant that depends on $\hbfw$.
\end{theorem}
This theorem shows that the error under the partial setting is bounded not only by the gap between the continuous eigenvalues in $e^{G}(d')$, but also by the precision of the weight vector $\hbfw$. The more accurate that $\hbfw$ is helps the model effectively mitigate the misalignment problem. The proof is provided in Section S.4 of the supplementary material.

\subsection{Optimization and Algorithm}\label{subsec:Optimization}
The networks are optimized by back-propagation in a mini-batch training manner. For convenience, we assume that the manifold network starts from the $l_m$-th layer and ends at the penultimate layer. Denote the network parameter of the $l$-th layer as $\bm{\Theta}_l$ ($l=1,2,\ldots,l_m,\ldots,L$). Note that $\bm{\Theta}$ not only includes the parameters of the manifold layers, but also contains the CNNs. The gradient of the objective function Eq.~\eqref{eq:Objective} referred by $\bm{\Theta}_l$ is divided into two parts according to the chain rule. The first part is the derivative of the objective function with respect to its input network features, and the second is the derivative of the network features with respect to the network parameter $\bm{\Theta}_l$. As the second part follows the standard network optimization paradigm, we only focus on the first part here.

As for the discriminant and alignment loss terms, the derivations begin with the objective function of the $l'$-th layer. Reformulating the inter-class loss $\mathcal{L}^{l'}_{inter}$ and intra-class loss $\mathcal{L}^{l'}_{intra}$ as
\begin{equation*}\label{eq:InterIntraLossRe}
\begin{split}
\mathcal{L}^{l'}_{inter}({\bf H}^s_{l'}) &= \frac{2}{c(c-1)}\sum_{i<j} \sum_{o=1}^{d_{l'}} \hat{h}^{s}_{l',oi} \hat{h}^{s}_{l',oj}, \\
\mathcal{L}^{l'}_{intra}({\bf H}^t_{l'}, {\bf P}^t) &= -\frac{1}{b_s k}\sum_{i=1}^{c} w_i \sum_{j=1}^{b_s} \scalebox{1.3}{$\chi$}_{ij} p^t_{ij} \sum_{o=1}^{d_{l'}} \bar{h}_{l',oi}^s h_{l',oj}^t.
\end{split}
\end{equation*}
Their derivatives are calculated as
\begin{equation*}\label{eq:InterIntraLayerGradient}
\begin{split}
\frac{\partial \mathcal{L}^{l'}_{inter}}{\partial \bm{\Theta}_l} =& \sum_{i<j} \sum_{o=1}^{d_{l'}} \left( \frac{\partial \mathcal{L}^{l'}_{inter}}{\partial h^{s}_{l',oi}} \frac{\partial h^{s}_{l',oi}}{\partial \bm{\Theta}_l} + \frac{\partial \mathcal{L}^{l'}_{inter}}{\partial h^{s}_{l',oj}} \frac{\partial h^{s}_{l',oj}}{\partial \bm{\Theta}_l} \right) \\
=& \frac{2}{c(c-1)} \sum_{i<j} \sum_{o=1}^{d_{l'}} \left( \frac{\hat{h}^{s}_{l',oj}}{n^s_{y_i}} \frac{\partial h^{s}_{l',oi}}{\partial \bm{\Theta}_l} + \frac{\hat{h}^{s}_{l',oi}}{n^s_{y_j}} \frac{\partial h^{s}_{l',oj}}{\partial \bm{\Theta}_l} \right), \\
\frac{\partial \mathcal{L}^{l'}_{intra}}{\partial \bm{\Theta}_l} =& \sum_{j=1}^{b_s} \sum_{o=1}^{d_{l'}} \frac{\partial \mathcal{L}^{l'}_{intra}}{\partial h_{l',oj}^t} \frac{\partial h_{l',oj}^t}{\partial \bm{\Theta}_l} +
\sum_{i=1}^{c}\sum_{j=1}^{b_s} \frac{\partial \mathcal{L}^{l'}_{intra}}{\partial p^{t}_{ij}} \frac{\partial p^{t}_{ij}}{\partial \bm{\Theta}_l} \\
=& -\frac{1}{b_s k} \left(\sum_{i=1}^{c}\sum_{j=1}^{b_s}\sum_{o=1}^{d_{l'}} w_i \scalebox{1.3}{$\chi$}_{ij} p^{t}_{ij} \bar{h}_{l',oi}^s \frac{\partial h_{l',oj}^t}{\partial \bm{\Theta}_l} \right. \\
&~~~~~~~~~~ \left. + \sum_{i=1}^{c}\sum_{j=1}^{b_s} w_i \scalebox{1.3}{$\chi$}_{ij} s^w_{l',ij} \frac{\partial p^{t}_{ij}}{\partial \bm{\Theta}_l} \right),
\end{split}
\end{equation*}
where $n^s_{y_i}$ is the number of the $y_i$-th class's source samples in the current batch, and
\begin{equation*}
\frac{\partial h^{s}_{l',oi}}{\partial \bm{\Theta}_l} = \frac{\partial h^{s}_{l',oj}}{\partial \bm{\Theta}_l} = \frac{\partial h_{l',oj}^t}{\partial \bm{\Theta}_l} = {\bf 0},~~~~\text{if}~~l'<l.
\end{equation*}
The derivative of discriminative structure loss $\mathcal{L}_{DS}$ is obtained by combing the layer-wise derivatives:
\begin{equation}\label{eq:DSGradient}
\frac{\partial \mathcal{L}_{DS}}{\partial \bm{\Theta}_l} = \sum_{l'=l_m}^{L-1} \left( \frac{\partial \mathcal{L}^{l'}_{inter}}{\partial \bm{\Theta}_l} + \frac{\partial \mathcal{L}^{l'}_{intra}}{\partial \bm{\Theta}_l} \right).
\end{equation}

In terms of manifold alignment loss, we take the case of the Grassmann manifold as an example. The manifold alignment loss $\mathcal{L}_{align}^{l'}$ requires the projection orthogonal bases of covariance matrices ${\bf C}^{s'}_{l'}$ and ${\bf C}^t_{l'}$, which are equivalent to the left-singular vectors of ${\bf H}^{s'}_{l'}$ and ${\bf H}^t_{l'}$, respectively. Specifically, the $d'_{l}$ projection vectors required in the Grassmann manifold are computed from the truncated SVD:
\begin{equation*}\label{eq:FeatureSVD}
{\bf H}^{s'}_{l'} = {\bf U}^{s'}_{l'} \bm{\Sigma}^{s'}_{l'} {\bf V}^{s'^T}_{l'},~~ {\bf H}^{t}_{l'} = {\bf U}^{t}_{l'} \bm{\Sigma}^{t}_{l'} {\bf V}^{t^T}_{l'},
\end{equation*}
where ${\bf U}^{s'}_{l'}, {\bf U}^{t}_{l'}\in \mathbb{R}^{d_{l'}\times d'_{l'}}$ and $\bm{\Sigma}^{s'}_{l'}, \bm{\Sigma}^{t}_{l'}\in \mathbb{R}^{d'_{l'} \times d'_{l'}}$. Following the matrix chain rule and the Jacobian of SVD \cite{papadopoulo2000estimating}, the derivative of $\mathcal{L}^{l'}_{align}$ is written as
\begin{equation*}\label{eq:AlignLayerGradient}
\begin{split}
\frac{\partial \mathcal{L}^{l'}_{align}}{\partial \bm{\Theta}_l} =& \sum_{i=1}^{b_s} \sum_{o=1}^{d_{l'}} \left( \frac{\partial \mathcal{L}^{l'}_{align}}{\partial h^{s'}_{l',oi}} \frac{{\partial h^{s'}_{l',oi}}}{\bm{\Theta}_l} + \frac{\partial \mathcal{L}^{l'}_{align}}{\partial h^{t}_{l',oi}} \frac{{\partial h^{t}_{l',oi}}}{\bm{\Theta}_l} \right) \\
=& \sum_{i=1}^{b_s} \sum_{o=1}^{d_{l'}} \left[ \text{tr} \left( {\bf A}_{l',oi}^{s'} \right)  \frac{{\partial h^{s'}_{l',oi}}}{\bm{\Theta}_l} + \text{tr} \left( {\bf A}_{l',oi}^{t} \right) \frac{{\partial h^{t}_{l',oi}}}{\bm{\Theta}_l} \right], \\
{\bf A}_{l',oi}^{s'} =&~ 4\left( \bm{\Omega}_{oi}^{{\bf U}^{s'}_{l'}} - {\bf U}^{{s'}^T}_{l'} {\bf U}^t_{l'} {\bf U}^{t^T}_{l'} {\bf U}^{{s'}}_{l'}  \bm{\Omega}_{oi}^{{\bf U}^{s'}_{l'}} \right),\\
{\bf A}_{l',oi}^{t} =&~ 4\left( \bm{\Omega}_{oi}^{{\bf U}^{t}_{l'}} -  {\bf U}^{{t}^T}_{l'} {\bf U}^{s'}_{l'} {\bf U}^{{s'}^T}_{l'} {\bf U}^{{t}}_{l'} \bm{\Omega}_{oi}^{{\bf U}^{t}_{l'}} \right),
\end{split}
\end{equation*}
where $\bm{\Omega}_{oi}^{{\bf U}^{s'}_{l'}}$ and $\bm{\Omega}_{oi}^{{\bf U}^{t}_{l'}}$ are antisymmetric matrices which can be computed by solving a set of linear systems \cite{papadopoulo2000estimating}. Detailed derivations are provided in Section S.5 of the supplementary material. Similarly, the derivative of alignment loss is deduced by combining the layer-wise derivatives:
\begin{equation}\label{eq:AlignGradient}
\frac{\partial \mathcal{L}_{AL}}{\partial \bm{\Theta}_l} = \sum_{l'=l_m}^{L-1} \frac{\partial \mathcal{L}^{l'}_{align}}{\partial \bm{\Theta}_l} .
\end{equation}

Recall that the cross-entropy loss $\mathcal{L}_{CE}$ is formulated as $\mathcal{L}_{CE} = \frac{1}{b_s}\sum_{i=1}^{b_s} -\log{p^s_{y_i i}}$. The derivative of the $\mathcal{L}_{CE}$ with respect to $\bm{\Theta}_l$ is
\begin{equation}\label{eq:CELossGradient}
\frac{\partial \mathcal{L}_{CE}}{\partial \bm{\Theta}_l} = \frac{1}{b_s} \sum_{i=1}^{b_s} \frac{\partial \mathcal{L}_{CE}}{\partial p^s_{y_i i}} \frac{\partial p^s_{y_i i}}{\partial \bm{\Theta}_l}
 = \sum_{i=1}^{b_s} -\frac{1}{b_s p^s_{y_i i}} \frac{\partial p^s_{y_i i}}{\partial \bm{\Theta}_l}.
\end{equation}

Finally, the gradient of $\bm{\Theta}_l$ is deduced from the Eq.~\eqref{eq:DSGradient}-\eqref{eq:CELossGradient}:
\begin{equation}\label{eq:TotalGradient}
\nabla \bm{\Theta}_l = \frac{\partial \mathcal{L}}{\partial \bm{\Theta}_l} = \frac{\partial \mathcal{L}_{CE}}{\partial \bm{\Theta}_l} + \lambda_1 \frac{\partial \mathcal{L}_{DS}}{\partial \bm{\Theta}_l} +
\lambda_2 \frac{\partial \mathcal{L}_{AL}}{\partial \bm{\Theta}_l}.
\end{equation}
The pseudo code for the optimization of the DMP framework is summarized in Algorithm \ref{alg:DMP}.

\begin{figure*}[t]
\begin{minipage}{0.245\linewidth}
    \centering{\includegraphics[width=0.99\linewidth]{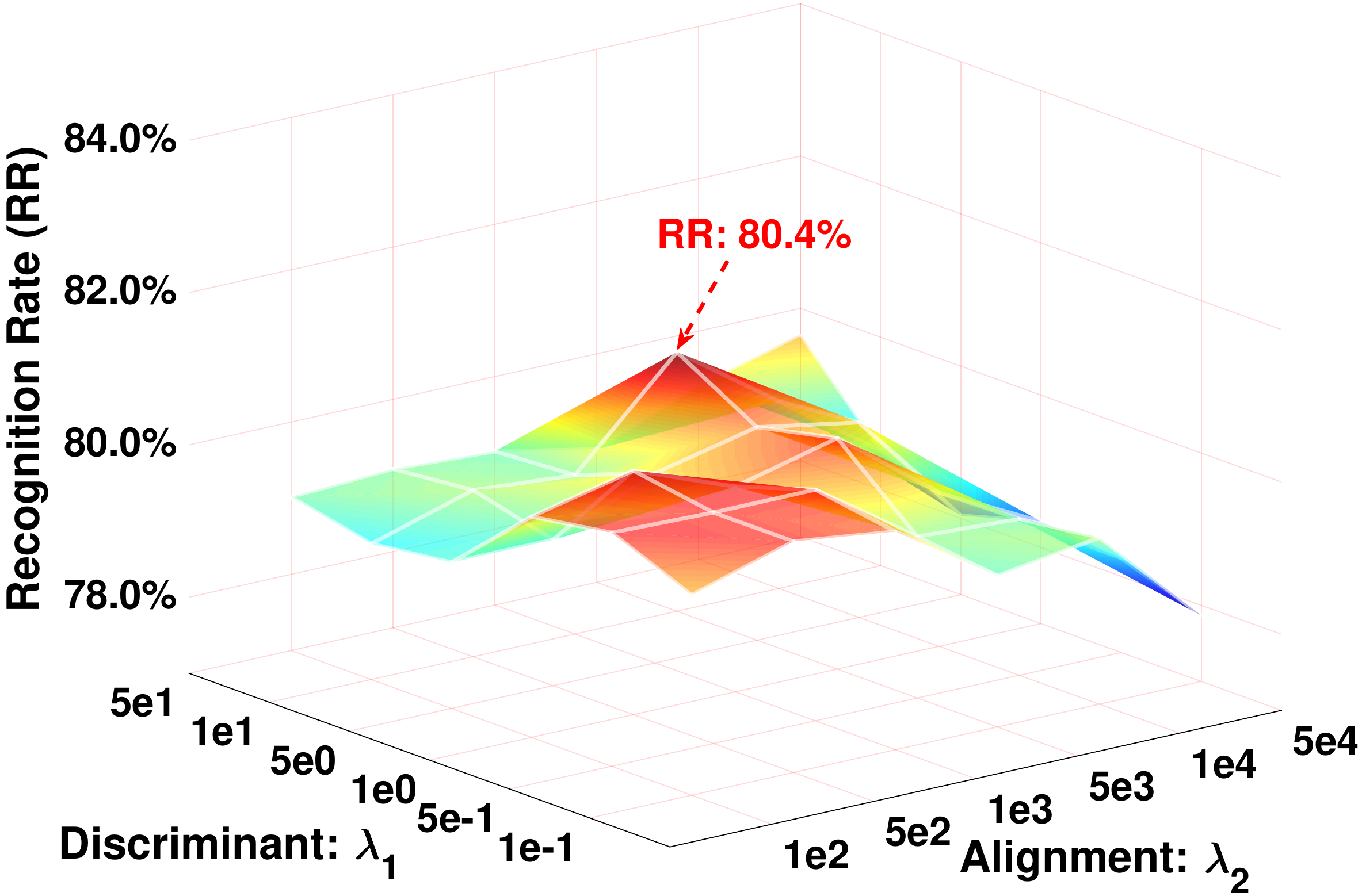}} \\
    (a) Task \textbf{I} $\rightarrow$ \textbf{P}
\end{minipage}
\hfill
\begin{minipage}{0.245\linewidth}
    \centering{\includegraphics[width=0.99\linewidth]{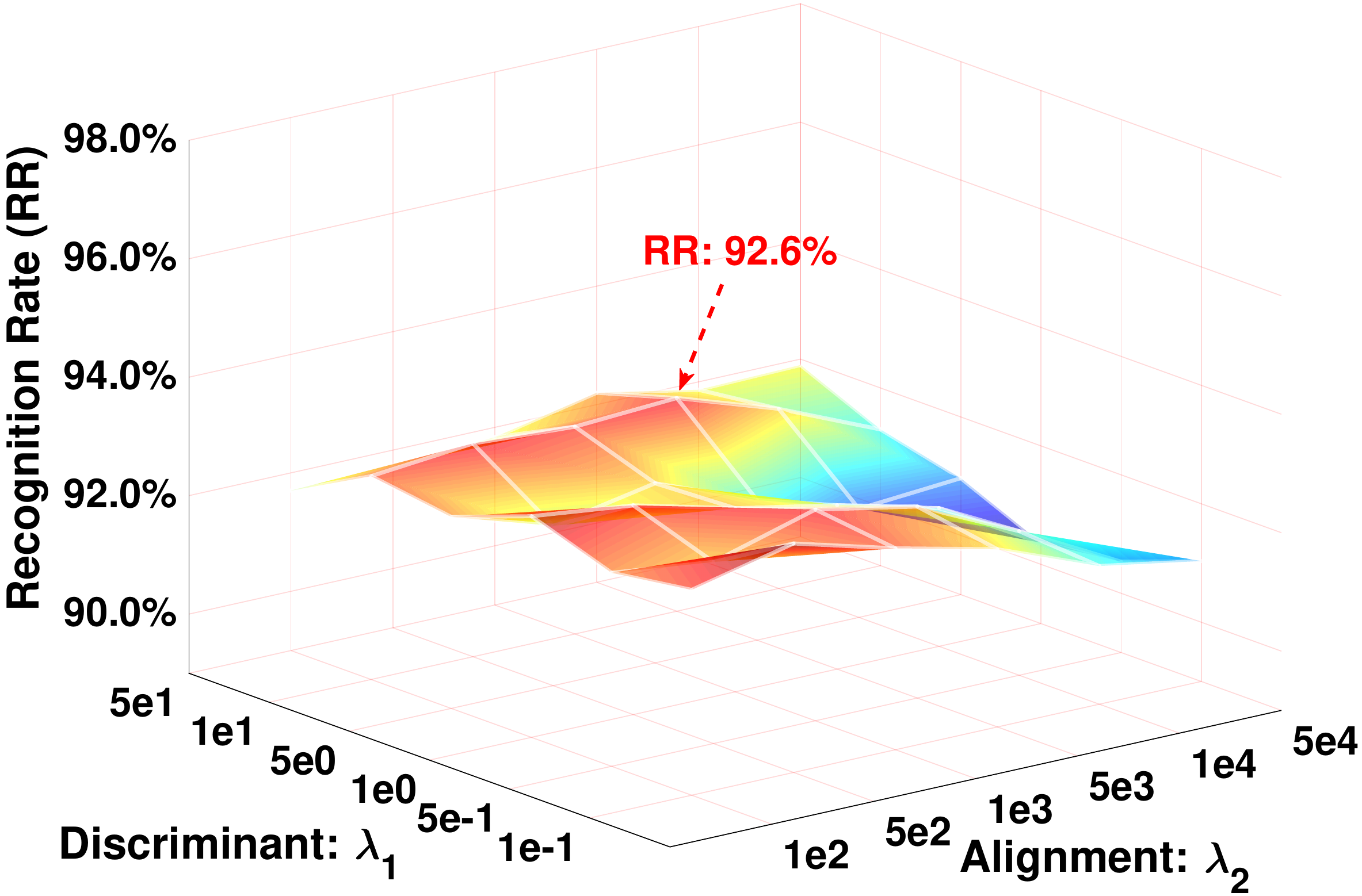}} \\
    (b) Task \textbf{P} $\rightarrow$ \textbf{I}
\end{minipage}
\hfill
\begin{minipage}{0.245\linewidth}
    \centering{\includegraphics[width=0.99\linewidth]{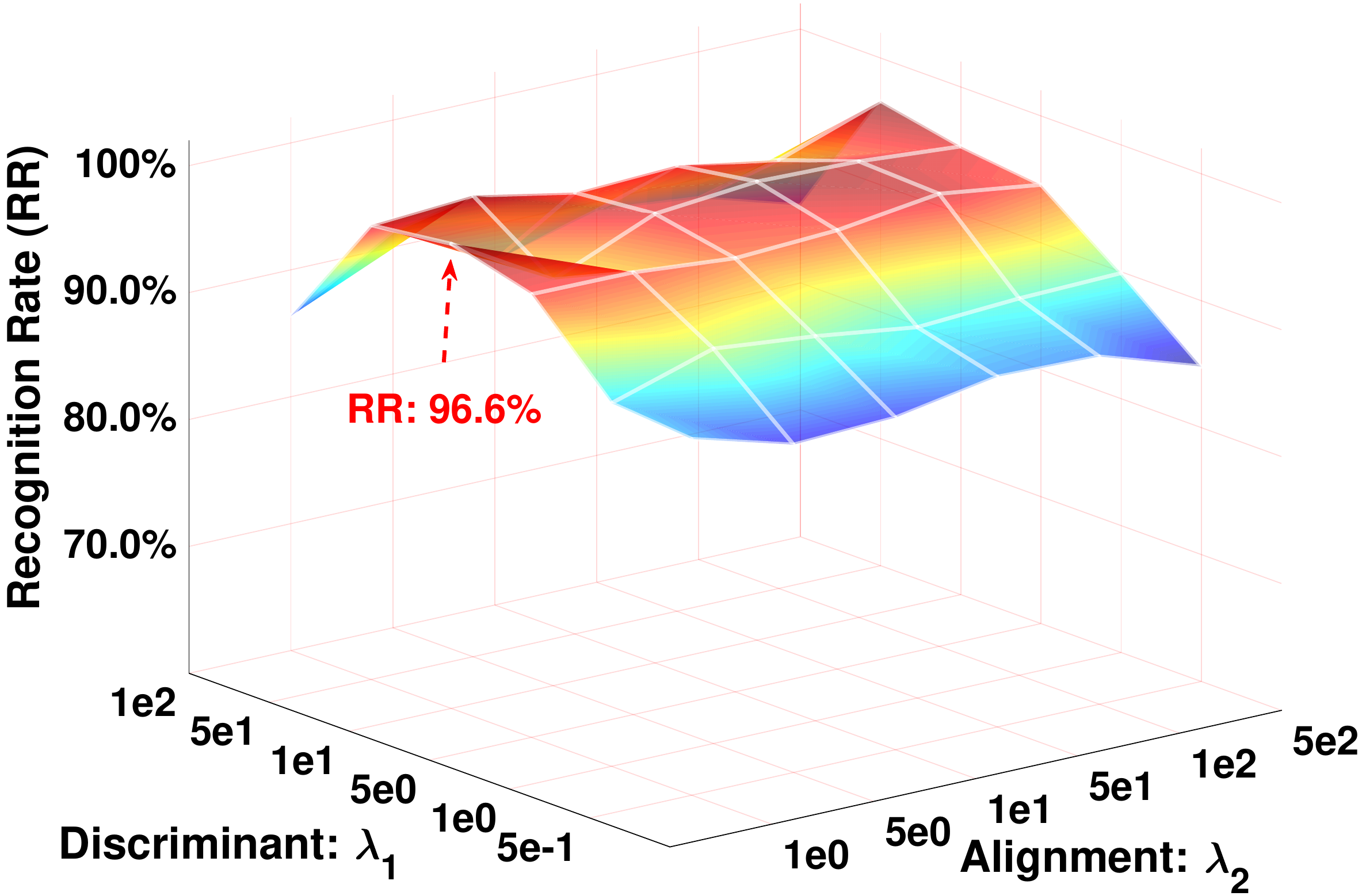}} \\
    (c) Task \textbf{A} $\rightarrow$ \textbf{W}
\end{minipage}
\hfill
\begin{minipage}{0.245\linewidth}
    \centering{\includegraphics[width=0.99\linewidth]{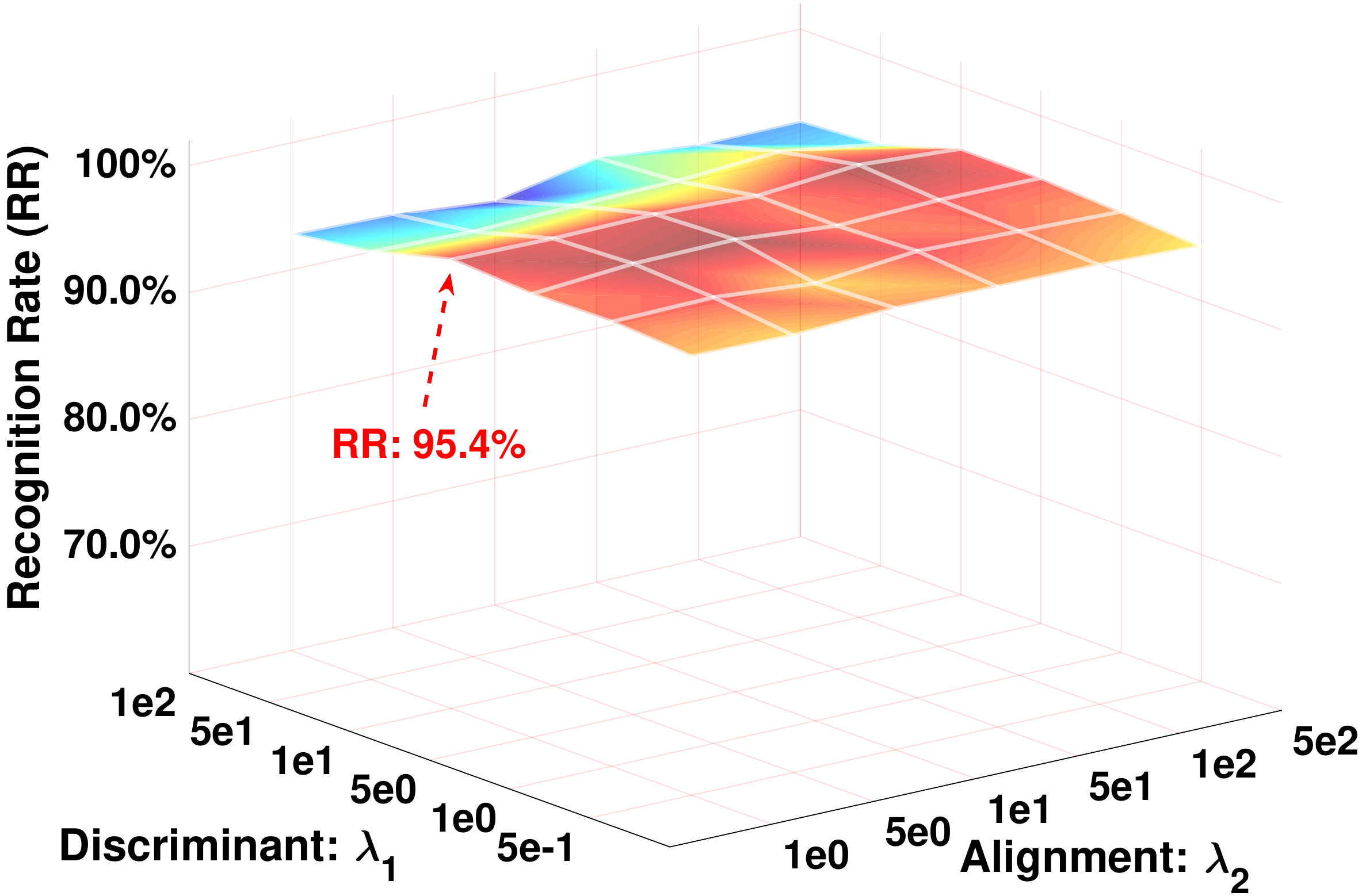}} \\
    (d) Task \textbf{W} $\rightarrow$ \textbf{A}
\end{minipage}
   \caption{3D heatmaps of the recognition rates with different penalty hyper-parameters under different settings. (a)-(b): \textit{vanilla} setting on ImageCLEF. (c)-(d): \textit{partial} setting on Office-31. Best viewed in color.}
\label{fig:Hyper}
\end{figure*}

\section{Experiments and Comparative Analysis}\label{sec:Experiment}
In this section, the proposed method is evaluated on four standard UDA datasets. The experimental setting is detailed in Section \ref{subsec:Data&Setting}. In Section \ref{subsec:Hyper-parameter}, we analyze the hyper-parameters of the DMP method. Based on the results of hyper-parameter anlysis, the comparison experiments for the vanilla and partial UDA problems are conducted in Sections \ref{subsec:VanillaExperiment} and \ref{subsec:PartialExperiment}, respectively, where the results of the ablation study are also provided. Further analysis on the proposed method is presented in Section \ref{subsec:MethodAnalysis}.
\subsection{Datasets and Experimental Setting}\label{subsec:Data&Setting}

\begin{figure*}[t]
\begin{minipage}{0.245\linewidth}
    \centering{\includegraphics[width=0.99\linewidth]{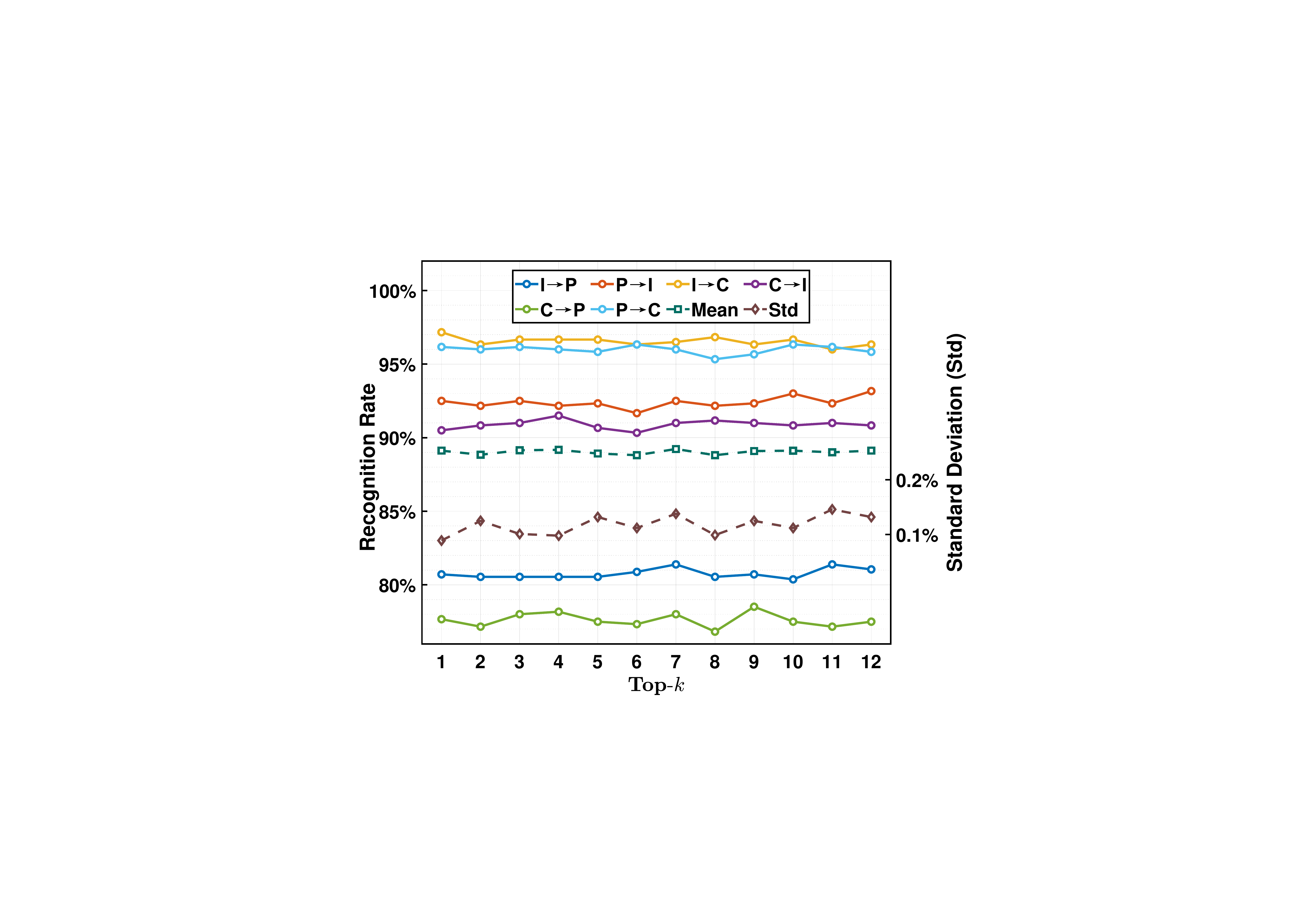}} \\
    (a) Sensitivity to parameter $k$
\end{minipage}
\hfill
\begin{minipage}{0.245\linewidth}
    \centering{\includegraphics[width=0.99\linewidth]{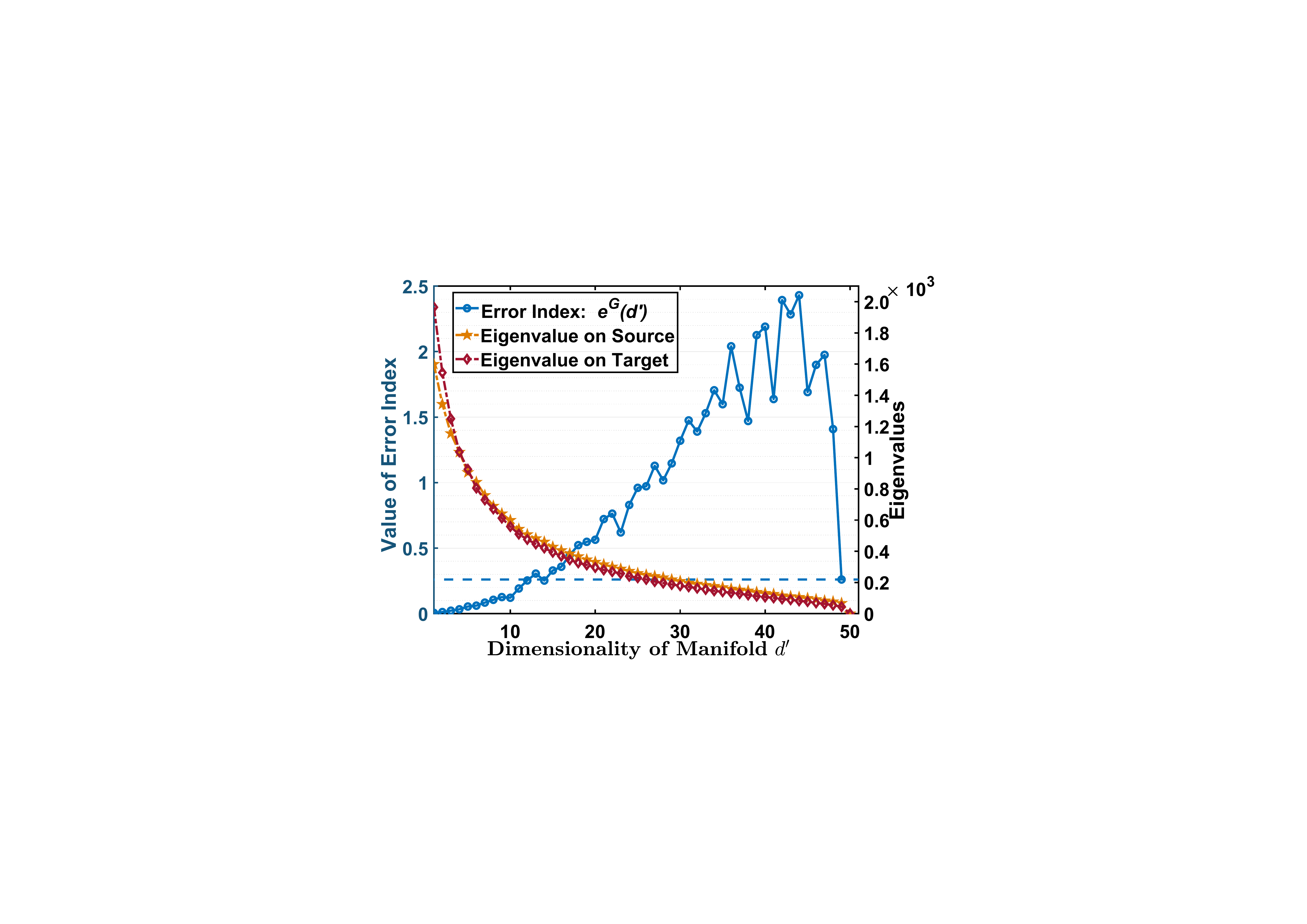}} \\
    (b) Visualization of $e^G(d')$
\end{minipage}
\hfill
\begin{minipage}{0.245\linewidth}
    \centering{\includegraphics[width=0.99\linewidth]{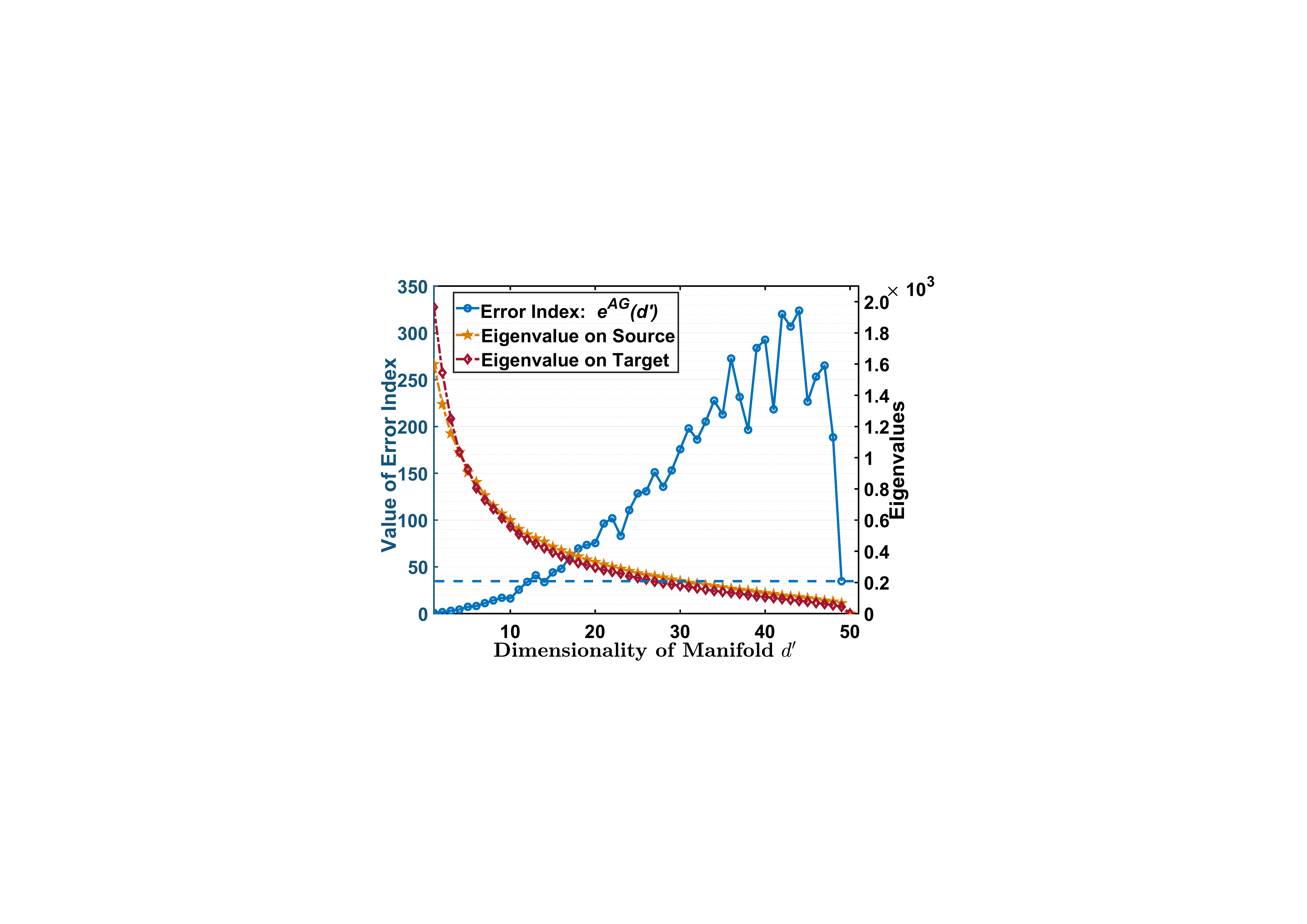}} \\
    (c) Visualization of $e^{AG}(d')$
\end{minipage}
\hfill
\begin{minipage}{0.245\linewidth}
    \centering{\includegraphics[width=0.99\linewidth]{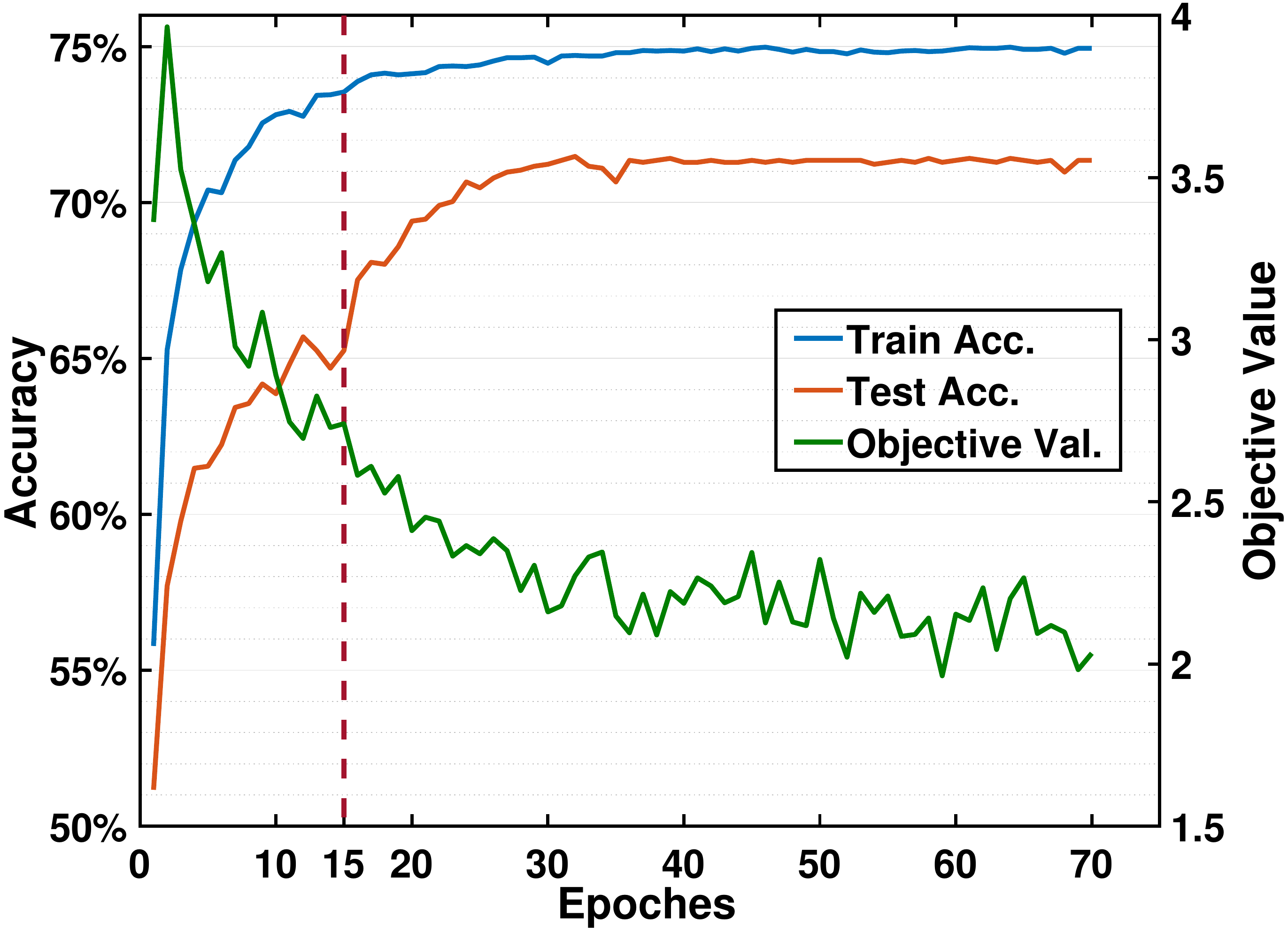}} \\
    (d) Convergence curves
\end{minipage}
   \caption{(a): Recognition rate curves of different truncated parameters $k$ on ImageCLEF dataset. The cyan and brown dash curves indicate the mean and standard deviation, respectively. (b)-(c): Error and eigenvalue curves w.r.t. the dimensionality $d'$. The $(b_s-1)$-th error index is highlighted by the horizontal dash line. (d): Recognition rate curves and the objective curve on the Office-31 dataset (\textbf{A}$\rightarrow$\textbf{W}). Best viewed in color.
   }
\label{fig:Err&Sen&Con}
\vspace{-10pt}
\end{figure*}

Four popular domain adaptation datasets were selected, and the standard protocols were adopted.

\textbf{Office 31} \cite{Office31} is an object recognition dataset. It contains 3 domains with a total of 4110 images, i.e., \textit{Amazon} (\textbf{A}), \textit{Webcam} (\textbf{W}) and \textit{Dslr} (\textbf{D}). All domains consist of 12 common classes, e,g., pen, monitor and speaker, and the sample sizes of the domains differ. Following the protocols in \cite{cao2018PADA,zhang2018importance}, the target domain consists of 10 classes extracted from the Office 31 dataset under the partial setting.

\textbf{Office-Home} \cite{OfficeHome} dataset contains 4 domains, i.e., \textit{Art} (\textbf{Ar}), \textit{Clipart} (\textbf{Cl}), \textit{Product} (\textbf{Pr}) and \textit{Real-World} (\textbf{Rw}), and 12 domain adaptation tasks. 15500 images are assigned into 65 categories for all domains. There are around 70 images for each category. For the partial setting, the first 25 classes (in alphabetical order) are taken as the target domains. Similarly, the sample sizes of different domains are unbalanced.

\textbf{ImageCLEF}\footnote{\url{https://www.imageclef.org/2014/adaptation}} dataset is obtained from the ImageCLEF Domain Adaptation challenge held in 2014. The domains \textit{Caltech} (\textbf{C}), \textit{ImageNet} (\textbf{I}), \textit{Pascal} (\textbf{P}), \textit{Bing} (\textbf{B}) were collected from four previous proposed datasets, i.e., Caltech-256 \cite{griffin2007caltech}, ImageNet ILSVRC2012 \cite{russakovsky2015imagenet}, PASCAL VOC2012 \cite{everingham2010pascal} and Bing \cite{bergamo2010exploiting}, respectively. Each domain consists of 600 images with 12 classes and all 4 domains are the same size. Following previous protocols \cite{long2019learning,ganin2016domain,Long2017Deep,long2018conditional}, we conduct adaptation tasks between \textit{Caltech}, \textit{ImageNet} and \textit{Pascal}. For the partial setting, the target domain consists of the first 6 classes (in alphabetical order) of the vanilla data.

\textbf{VisDA-2017} \cite{VisDA-2017} is a large-scale visual domain adaptation challenge dataset. It aims to transfer the knowledge learned from sufficient synthetic data to real-life visual scenes. The source domain \textbf{S} (\textbf{synthetic}) collected 152397 images generated from 3D models. The target domain \textbf{R} (\textbf{real-image}) extracted 55388 cropped images from the Microsoft COCO databse \cite{lin2014microsoft}. All collected images were assigned into 12 common classes. Following the challenge track, we carried out a \textbf{S} $\rightarrow$ \textbf{R} task for the vanilla setting, and a \textbf{S} $\rightarrow$ \textbf{R6} (the first 6 classes in alphabetical order) task for the partial setting.

In the following experiments, ResNet models \cite{he2016deep} pre-trained on ImageNet were used for feature extraction. Specifically, ResNet-50 and Adam Optimizer ($lr=0.0002$, $\beta_1=0.9$, $\beta_2 = 0.999$) with batch size of 50 were used on Office 31, Office-Home and ImageCLEF datasets. ResNet-101 and the modified mini-batch SGD (momentum = 0.9, weight decay = 5$e$-4) with batch sizes of 32 were employed on VisDA-2017 challenge. The initial learning rate was set as 0.003 and then adjusted by the annealing learning schedule described in \cite{ganin2016domain}. A two layer Riemmanian manifold learning scheme was carried out in all experiments, with the first layer ($d_1=1024$) activated by Leaky ReLU ($\alpha$ = 0.2) and the second ($d_2=512$) by Tanh. The learning rates for the feature extraction layers and Riemannian manifold layers learned from scratch were set as $0.1lr$ and $lr$. We implemented DMP on a GPU PyTorch platform with an NVIDIA GTX TITAN Xp.

\subsection{Hyper-parameter Analysis}\label{subsec:Hyper-parameter}

The penalty parameters $\lambda_1$, $\lambda_2$ and truncated parameter $k$ were investigated first. We searched the penalty parameters with the Top-1 preserving scheme. Then the truncated parameter $k$ was analyzed based on the optimal parameters $\lambda_1$, $\lambda_2$ in the last stage.

For the vanilla setting, the experiments were conducted on ImageCLEF, and $\lambda_1$ and $\lambda_2$ were tested for each value in groups \{1$e$-1,5$e$-1,1$e$0,5$e$0,1$e$1,5$e$1\} and \{1$e$2,5$e$2,1$e$3,5$e$3,1$e$4,5$e$4\}, respectively. For the partial setting, the experiments were carried out on partial Office-31, and the optimal values of $\lambda_1$ and $\lambda_2$ were searched from \{5$e$-1,1$e$0,5$e$0,1$e$1,5$e$1,1$e$2\} and \{1$e$0,5$e$0,1$e$1,5$e$1,1$e$2,5$e$2\}, respectively. All settings were repeated 50 times and heatmaps of mean results are presented in Figure \ref{fig:Hyper}. The highest accuracies, i.e., the \textit{peaks} in the figures, are marked with red arrows. Regions nearby the \textit{peaks} are flat, which means the proposed method are stable in those regions. In the vanilla tasks, setting $(\lambda_1,\lambda_2) = (1e1,5e3)$ achieved the 1st and 2nd best recognition rates in tasks \textbf{I} $\rightarrow$ \textbf{P} and \textbf{P} $\rightarrow$ \textbf{I}, respectively. As the domain alignment should be conservative in the partial setting, the optimal alignment parameter $\lambda_2$ is smaller than that in the vanilla setting. Figure \ref{fig:Hyper}(c)-(d) shows that the optimal parameter setting, i.e., $(\lambda_1,\lambda_2)$ = (1$e$1,1$e$0), achieved high accuracies in partial tasks \textbf{A} $\rightarrow$ \textbf{W} and \textbf{W} $\rightarrow$ \textbf{A}.

We fixed parameters $(\lambda_1,\lambda_2)$=(1$e$1,5$e$3), and then investigated the sensitivity of the proposed method with reference to the truncated parameter $k$. The recognition curves of all six transfer tasks on ImageCLEF are shown in Figure \ref{fig:Err&Sen&Con}(a). Our method is robust to the selection of $k$, as all curves are flat and stable. The mean accuracy, shown as the cyan dashed line, suggests that the accuracies of different truncated settings are nearly the same. However, it can be observed from the brown dashed line that truncation makes the method more robust and the standard deviation of the Top-1 scheme is the smallest.

To explore the minimal errors of Grassmann and affine Grassmann distances, a numerical simulation was conducted on the ImageCLEF dataset. As the eigenvalues always decrease rapidly at the beginning and then become flat, the error bounds of dimensionality $d'$ located in the flattened eigenvalue region are too high to assess. As shown in Figure \ref{fig:Err&Sen&Con}(b)-(c), the trend of eigenvalues is consistent with the description. Though the dramatic decrease in the initial stage results in a lower error, the information in that area is unconvincing and insufficient to support the measurement of manifolds. Since there is a natural gap between the ($b_s-1$)-th and the $b_s$-th dominant eigenvalues, ($b_s-1$)-th error index is smaller than most of the other errors. We highlight the ($b_s-1$)-th error index by the blue dashed line, and observe only errors of $d'=\{1,2,\ldots,12,14\}$ that are lower than the ($b_s-1$)-th error. The errors of the Grassmann distance and affine Grassmann distance have the same trend, this is because the covariances of different domains are aligned via manifold metrics.
Empirically, the dimensions of the Grassmann and affine Grassmann manifolds were set as ($b_s-1$) hereafter.

\begin{table}[t]
\setlength{\abovecaptionskip}{0.cm}
\setlength{\belowcaptionskip}{-0.01cm}
\caption{Ablation study of the manifold selection on ImageCLEF.}
\label{tab:Manifold_Ablation}
\renewcommand{\tabcolsep}{0.3pc} 
\renewcommand{\arraystretch}{1.0} 
\centering
\begin{tabular}{cc||ccccccc}
\toprule[1pt]
\multicolumn{2}{c||}{Manifolds} & I$\rightarrow$P & P$\rightarrow$I & I$\rightarrow$C & C$\rightarrow$I & C$\rightarrow$P & P$\rightarrow$C & Mean \\
\hline
\hline
\multirow{3}{*}{Vanilla} & $G$      & $80.7$ & $92.5$ & $97.2$ & $90.5$ & $77.7$ & $96.2$ & 89.1 \\
                         & $AG$     & $81.0$ & $92.0$ & $96.3$ & $91.0$ & $76.5$ & $96.5$ & 88.9 \\
                         & $LE$     & $81.4$ & $92.5$ & $96.3$ & $91.3$ & $78.0$ & $95.3$ & 89.1 \\
\hline
\multirow{3}{*}{Partial} & $G$      & $82.4$ & $94.5$ & $96.7$ & $94.3$ & $78.7$ & $96.4$ & 90.5 \\
                         & $AG$     & $79.8$ & $90.9$ & $95.7$ & $89.7$ & $75.9$ & $95.4$ & 87.9 \\
                         & $LE$     & $81.0$ & $92.0$ & $96.3$ & $91.0$ & $76.5$ & $96.5$ & 88.9 \\
\bottomrule[1pt]
\end{tabular}
\vspace{-10pt}
\end{table}

\begin{table*}[t]
\setlength{\abovecaptionskip}{0.cm}
\setlength{\belowcaptionskip}{-0.01cm}
\caption{Class-wise recognition rates (\%) on VisDA-2017 (ResNet-101) and recognition rates (\%) on Office-Home, ImageCLEF and Office-31 (ResNet-50) under the Vanilla Setting. The superscripts denote standard deviations hereafter.}
\label{tab:4Dataset_vanilla}
\renewcommand{\tabcolsep}{0.3pc} 
\renewcommand{\arraystretch}{1.0} 
\centering
\begin{tabular}{c||cccccccccccc|c}
\toprule[1pt]
\textbf{VisDA-2017} & Plane & bcycl & bus & car & horse & knife & mcyle & person & plant & sktbrd & train & truck & Mean \\
\hline
\hline
ResNet-101 \cite{he2016deep}                   & 55.1 &  53.3 &  61.9 &  59.1 &  80.6 &  17.9 &  79.7 &  31.2 &  81.0 &  26.5 &  73.5 &  8.5 &  52.4 \\
DAN \cite{long2019learning}                    & 87.1 &  63.0 &  76.5 &  42.0 &  90.3 &  42.9 &  85.9 &  53.1 &  49.7 &  36.3 &  85.8 &  20.7 &  61.1 \\				
DANN \cite{ganin2016domain}                    & 81.9 &  77.7 &  82.8 &  44.3 &  81.2 &  29.5 &  65.1 &  28.6 &  51.9 &  54.6 &  82.8 &  7.8 &  57.4 \\
CDAN \cite{long2018conditional}                & 85.2 &  66.9 &  83.0 &  50.8 &  84.2 &  74.9 &  88.1 &  74.5 &  83.4 &  76.0 &  81.9 &  38.0 &  73.7 \\
BSP+DANN \cite{chen2019transferability}        & 92.2 &  72.5 &  83.8 &  47.5 &  87.0 &  54.0 &  86.8 &  72.4 &  80.6 &  66.9 &  84.5 &  37.1 &  72.1 \\
BSP+CDAN \cite{chen2019transferability}        & 92.4 &  61.0 &  81.0 &  57.5 &  89.0 &  80.6 &  90.1 &  77.0 &  84.2 &  \textbf{77.9} & 82.1 &  38.4 &  75.9 \\
HAFN \cite{xu2019larger}                       & 92.7 & 55.4 & 82.4 & 70.9 & 93.2 & 71.2 & 90.8 & 78.2 & 89.1 & 50.2 & \textbf{88.9} & 24.5 & 73.9 \\
SAFN \cite{xu2019larger}                       & 93.6 & 61.3 & 84.1 & 70.6 & \textbf{94.1} & 79.0 & 91.8 & 79.6 & 89.9 & 55.6 & 89.0 & 24.4 & 76.1 \\
\hline
DMP (No AL)                                  & 92.8 & 15.3 & \textbf{86.7} & \textbf{86.3} & 93.8 & 70.7 & \textbf{95.2} & 68.9 & \textbf{95.8} & 40.4 & 85.1 & 5.6 & 69.7 \\
DMP (No DS)                                  & 90.2 & 66.5 & 70.2 & 65.8 & 79.8 & 81.8 & 84.7 & 70.1 & 82.0 & 46.5 & 88.1 & 27.7 & 71.1 \\
DMP                                          & 92.1 & 75.0 & 78.9 & 75.5 & 91.2 & \textbf{81.9} & 89.0 & 77.2 & 93.3 & 77.4 & 84.8 & 35.1 & \textbf{79.3} \\
\bottomrule[1pt]
\toprule[1pt]
\textbf{Office-Home} & Ar$\rightarrow$Cl & Ar$\rightarrow$Pr & Ar$\rightarrow$Rw & Cl$\rightarrow$Ar & Cl$\rightarrow$Pr & Cl$\rightarrow$Rw &
Pr$\rightarrow$Ar & Pr$\rightarrow$Cl & Pr$\rightarrow$Rw & Rw$\rightarrow$Ar & Rw$\rightarrow$Cl & Rw$\rightarrow$Pr & Mean \\
\hline
\hline
ResNet-50 \cite{he2016deep}              & 34.9 & 50.0 & 58.0 & 37.4 & 41.9 & 46.2 & 38.5 & 31.2 & 60.4 & 53.9 & 41.2 & 59.9 & 46.1 \\
DAN \cite{long2019learning}              & 43.6 & 57.0 & 67.9 & 45.8 & 56.5 & 60.4 & 44.0 & 43.6 & 67.7 & 63.1 & 51.5 & 74.3 & 56.3 \\				
DANN \cite{ganin2016domain}              & 45.6 & 59.3 & 70.1 & 47.0 & 58.5 & 60.9 & 46.1 & 43.7 & 68.5 & 63.2 & 51.8 & 76.8 & 57.6 \\
JAN \cite{Long2017Deep}                  & 45.9 & 61.2 & 68.9 & 50.4 & 59.7 & 61.0 & 45.8 & 43.4 & 70.3 & 63.9 & 52.4 & 76.8 & 58.3 \\
CDAN \cite{long2018conditional}          & 49.0 & 69.3 & 74.5 & 54.4 & 66.0 & 68.4 & 55.6 & 48.3 & 75.9 & 68.4 & 55.4 & 80.5 & 63.8 \\
CDAN+E \cite{long2018conditional} & 50.7 & 70.6 & 76.0 & 57.6 & 70.0 & 70.0 & 57.4 & 50.9 & 77.3 & 70.9 & 56.7 & 81.6 & 65.8 \\
BSP+DANN \cite{chen2019transferability}  & 51.4 & 68.3 & 75.9 & 56.0 & 67.8 & 68.8 & 57.0 & 49.6 & 75.8 & 70.4 & 57.1 & 80.6 & 64.9 \\
BSP+CDAN \cite{chen2019transferability}  & 52.0 & 68.6 & 76.1 & 58.0 & 70.3 & 70.2 & 58.6 & 50.2 & 77.6 & \textbf{72.2} & \textbf{59.3} & \textbf{81.9} & 66.3 \\
HAFN \cite{xu2019larger}                 & 50.2 & 70.1 & 76.6 & 61.1 & 68.0 & 70.7 & 59.5 & 48.4 & 77.3 & 69.4 & 53.0 & 80.2 & 65.4 \\
SAFN \cite{xu2019larger}                 & 52.0 & 71.7 & 76.3 & 64.2 & 69.9 & 71.9 & 63.7 & 51.4 & 77.1 & 70.9 & 57.1 & 81.5 & 67.3 \\
\hline
DMP (No AL)                            & 51.9 & 72.8 & 77.1 & 63.0 & \textbf{72.0} & 71.3 & 60.5 & 49.5 & 78.4 & 71.5 & 54.4 & 82.8 & 67.1 \\
DMP (No DS)                            & 51.2 & 72.4 & \textbf{77.7} & 63.0 & 71.4 & 71.4 & 58.6 & 44.6 & \textbf{79.1} & 71.1 & 53.4 & 81.5 & 66.3 \\
DMP                                    & \textbf{52.3} & \textbf{73.0} & 77.3 & \textbf{64.3} & \textbf{72.0} & \textbf{71.8} & \textbf{63.6} & \textbf{52.7} & 78.5 & 72.0 & 57.7 & 81.6 & \textbf{68.1} \\
\bottomrule[1pt]
\end{tabular}
\\[2pt]
\begin{tabular}{c||ccccccc||ccccccc}
\toprule[1pt]
\multirow{2}{*}{Methods} & \multicolumn{7}{c||}{ImageCLEF} & \multicolumn{7}{c}{Office-31} \\
 & I$\rightarrow$P & P$\rightarrow$I & I$\rightarrow$C & C$\rightarrow$I & C$\rightarrow$P & P$\rightarrow$C & Mean & A$\rightarrow$W & D$\rightarrow$W & W$\rightarrow$D & A$\rightarrow$D & D$\rightarrow$A & W$\rightarrow$A & Mean \\
\hline
\hline
ResNet-50 \cite{he2016deep}             & $74.8^{0.3}$ & $83.9^{0.1}$ & $91.5^{0.3}$ & $78.0^{0.2}$ & $65.5^{0.3}$ & $91.2^{0.3}$ & 80.7 & $68.4^{0.2}$ & $96.7^{0.1}$ & $99.3^{0.1}$ & $ 68.9^{0.2}$ & $62.5^{0.3}$ & $60.7^{ 0.3}$ & 76.1 \\
DAN \cite{long2019learning}             & $74.5^{0.4}$ & $82.2^{0.2}$ & $92.8^{0.2}$ & $86.3^{0.4}$ & $69.2^{0.4}$ & $89.8^{0.4}$ & 82.5 & $80.5^{0.4}$ & $97.1^{0.2}$ & $99.6^{0.1}$ & $ 78.6^{0.2}$ & $63.6^{0.3}$ & $62.8^{ 0.2}$ & 80.4 \\				
DANN \cite{ganin2016domain}             & $75.0^{0.3}$ & $86.0^{0.3}$ & $96.2^{0.4}$ & $87.0^{0.5}$ & $74.3^{0.5}$ & $91.5^{0.6}$ & 85.0 & $82.0^{0.4}$ & $96.9^{0.2}$ & $99.1^{0.1}$ & $ 79.7^{0.4}$ & $68.2^{0.4}$ & $67.4^{ 0.5}$ & 82.2 \\
JAN \cite{Long2017Deep}                 & $76.8^{0.4}$ & $88.0^{0.2}$ & $94.7^{0.2}$ & $89.5^{0.3}$ & $74.2^{0.3}$ & $91.7^{0.3}$ & 85.8 & $85.4^{0.3}$ & $97.4^{0.2}$ & $99.8^{0.2}$ & $ 84.7^{0.3}$ & $68.6^{0.3}$ & $70.0^{ 0.4}$ & 84.3 \\
CDAN \cite{long2018conditional}         & $76.7^{0.3}$ & $90.6^{0.3}$ & $97.0^{0.4}$ & $90.5^{0.4}$ & $74.5^{0.3}$ & $93.5^{0.4}$ & 87.1 & $93.1^{0.2}$ & $98.2^{0.2}$ & $\textbf{100.0}^{0.0}$ & $89.8^{0.3}$ & $70.1^{0.4}$ & $68.0^{0.4}$ & 86.6 \\
CDAN+E \cite{long2018conditional}       & $77.7^{0.3}$ & $90.7^{0.2}$ & $\textbf{97.7}^{0.3}$ & $\textbf{91.3}^{0.3}$ & $74.2^{0.2}$ & $94.3^{0.3}$ & 87.7 & $\textbf{94.1}^{0.1}$ & $98.6^{0.1}$ & $\textbf{100.0}^{0.0}$ & $\textbf{92.9}^{0.2}$ & $71.0^{0.3}$ & $69.3^{0.3}$ & \textbf{87.7} \\
HAFN \cite{xu2019larger}                & $76.9^{0.4}$ & $89.0^{0.4}$ & $94.4^{0.1}$ & $89.6^{0.6}$ & $74.9^{0.2}$ & $92.9^{0.1}$ & 86.3 & $83.4^{0.7}$ & $98.3^{0.1}$ & $99.7^{0.1}$ & $ 84.4^{0.7}$ & $69.4^{0.5}$ & $68.5^{ 0.3}$ & 83.9 \\
SAFN \cite{xu2019larger}                & $78.0^{0.4}$ & $91.7^{0.5}$ & $96.2^{0.1}$ & $91.1^{0.3}$ & $77.0^{0.5}$ & $94.7^{0.3}$ & 88.1 & $88.8^{0.4}$ & $98.4^{0.0}$ & $99.8^{0.0}$ & $ 87.7^{1.3}$ & $69.8^{0.4}$ & $69.7^{ 0.2}$ & 85.7 \\
\hline
DMP (No AL)                             & $78.0^{0.1}$ & $91.1^{0.1}$ & $95.6^{0.2}$ & $88.7^{0.3}$ & $74.8^{0.1}$ & $94.8^{0.2}$ & 87.3 & $87.5^{0.4}$ & $98.9^{0.1}$ & $\textbf{100.0}^{0.0}$ & $86.8^{0.3}$ & $67.6^{0.1}$ & $64.1^{0.3}$ & 84.1 \\
DMP (No DS)                             & $78.9^{0.1}$ & $90.5^{0.2}$ & $94.0^{0.1}$ & $87.8^{0.1}$ & $76.7^{0.2}$ & $93.0^{0.1}$ & 86.8 & $82.7^{0.2}$ & $97.7^{0.1}$ & $\textbf{100.0}^{0.0}$ & $82.0^{0.2}$ & $66.2^{0.2}$ & $65.5^{0.2}$ & 82.3 \\
DMP                                     & $\textbf{80.7}^{0.1}$ & $\textbf{92.5}^{0.1}$ & $97.2^{0.1}$ & $90.5^{0.1}$ & $\textbf{77.7}^{0.2}$ & $\textbf{96.2}^{0.2}$ & \textbf{89.1} & $93.0^{0.3}$ & $ \textbf{99.0}^{0.1}$ & $\textbf{100.0}^{0.0}$ & $91.0^{0.4}$ & $\textbf{71.4}^{0.2}$ & $\textbf{70.2}^{0.2}$ & 87.4 \\
\bottomrule[1pt]
\end{tabular}
\vspace{-10pt}
\end{table*}

The ablation study of manifold selection was conducted on the ImageCLEF dataset, and the results are shown in Table \ref{tab:Manifold_Ablation}. The Grassmann, affine Grassmann and Log-Euclidean metrics are abbreviated as $G$, $AG$ and $LE$, respectively. The accuracies of the different manifolds are almost the same under the vanilla setting, but the Grassmann manifold is better than the others under the partial setting. This is because the affine Grassmann distance and Log-Euclidean metric refer to the mean vectors and eigenvalues and are more sensitive to the weight vector ${\bf w}$. Empirically, the Grassmann manifold was selected for the domain alignment hereafter.

Figure \ref{fig:Err&Sen&Con}(d) shows the convergence curves on Office-31 \textbf{A}$\rightarrow$\textbf{W} adaptation task. At the beginning, the objective values decrease quickly and the recognition rates tend to be stable in epochs 10-15. The intra-class structure constraint was imposed after 15 epochs, which led to a continuous improvement in the recognition rate and alleviated the over-fitting problem on the source domain.

\subsection{Comparative Experiments under Vanilla Setting}\label{subsec:VanillaExperiment}
In this section, we compare the proposed method with other state-of-the-art vanilla UDA approaches: DAN \cite{long2019learning}, Domain Adversarial Neural Network (DANN) \cite{ganin2016domain}, JAN \cite{Long2017Deep}, SWD \cite{lee2019sliced}, CDAN \cite{long2018conditional}, BSP based on DANN (BSP+DANN) or CDAN (BSP+CDAN) \cite{chen2019transferability}, Hard AFN (HAFN) and Stepwise AFN (SAFN) \cite{xu2019larger}.

\begin{table*}[t]
\setlength{\abovecaptionskip}{0.cm}
\setlength{\belowcaptionskip}{-0.01cm}
\caption{Recognition rates (\%) on ImageCLEF, Office-31, Office-Home and VisDA-2017 under the Partial Setting (ResNet-50).}
\label{tab:4Datasets_partial}
\renewcommand{\tabcolsep}{0.3pc} 
\renewcommand{\arraystretch}{1.0} 

\begin{tabular}{c||ccccccc||ccccccc}
\toprule[1pt]
\multirow{2}{*}{Methods} & \multicolumn{7}{c||}{Office-31} & \multicolumn{7}{c}{ImageCLEF} \\
 & A$\rightarrow$W & D$\rightarrow$W & W$\rightarrow$D & A$\rightarrow$D & D$\rightarrow$A & W$\rightarrow$A & Mean & I$\rightarrow$P & P$\rightarrow$I & I$\rightarrow$C & C$\rightarrow$I & C$\rightarrow$P & P$\rightarrow$C & Mean \\
\hline
\hline
ResNet-50 \cite{he2016deep} & $75.6^{1.1}$ & $96.3^{0.9}$ & $98.1^{0.7}$ & $83.4^{1.1}$ & $83.9^{1.0}$ & $85.0^{0.9}$ & 87.1 & $78.3^{0.2}$ & $86.9^{0.2}$ & $91.0^{0.2}$ & $84.3^{0.4}$ & $72.5^{0.4}$ & $91.5^{0.3}$ & 84.1 \\
DANN \cite{ganin2016domain} & $73.6^{0.2}$ & $96.3^{0.3}$ & $98.7^{0.2}$ & $81.5^{0.2}$ & $82.8^{0.2}$ & $86.1^{0.2}$ & 86.5 & $78.1^{0.2}$ & $86.3^{0.2}$ & $91.3^{0.4}$ & $84.0^{0.3}$ & $72.1^{0.3}$ & $90.3^{0.2}$ & 83.7 \\
PADA \cite{cao2018PADA}     & $86.5^{0.3}$ & $99.3^{0.5}$ & $\textbf{100.0}^{0.0}$ & $82.2^{0.4}$ & $92.7^{0.3}$ & $\textbf{95.4}^{0.3}$ & 92.7 & $81.7^{0.2}$ & $92.1^{0.2}$ & $94.6^{0.2}$ & $89.8^{0.2}$ & $77.7^{0.3}$ & $94.1^{0.1}$ & 88.3 \\
SAN \cite{cao2018SAN}       & $93.9^{0.5}$ & $99.3^{0.5}$ & $99.4^{0.1}$ & $94.3^{0.3}$ & $94.2^{0.4}$ & $88.7^{0.4}$ & 95.0 & $81.6^{0.2}$ & $91.1^{0.2}$ & $95.9^{0.3}$ & $90.4^{0.6}$ & $78.5^{0.2}$ & $\textbf{97.1}^{0.3}$ & 89.1 \\
HAFN \cite{xu2019larger}    & $87.5^{0.3}$ & $96.7^{0.4}$ & $99.2^{0.3}$ & $87.3^{0.5}$ & $89.2^{0.2}$ & $90.7^{0.2}$ & 91.7 & $79.1^{0.2}$ & $87.7^{0.1}$ & $93.7^{0.1}$ & $90.3^{0.2}$ & $77.8^{0.1}$ & $94.7^{0.3}$ & 87.2 \\
SAFN \cite{xu2019larger}    & $87.5^{0.7}$ & $96.6^{0.2}$ & $99.4^{0.7}$ & $89.8^{1.5}$ & $92.6^{0.2}$ & $92.7^{0.1}$ & 93.1 & $79.5^{0.2}$ & $90.7^{0.2}$ & $93.0^{0.1}$ & $90.3^{0.1}$ & $77.8^{0.2}$ & $94.0^{0.2}$ & 87.5 \\
\hline
DMP                         & $94.5^{0.5}$ & $99.9^{0.1}$ & $\textbf{100.0}^{0.0}$ & $95.0^{1.0}$ & $94.7^{0.3}$ & $\textbf{95.4}^{0.3}$ & 96.6 & $81.5^{0.2}$ & $94.3^{0.1}$ & $96.2^{0.1}$ & $93.0^{0.3}$ & $78.2^{0.2}$ & $96.5^{0.1}$ & 90.0 \\
DMP+ent                     & $\textbf{96.6}^{0.9}$ & $\textbf{100.0}^{0.0}$ & $\textbf{100.0}^{0.0}$ & $\textbf{96.4}^{0.9}$ & $\textbf{95.1}^{0.2}$ & $\textbf{95.4}^{0.1}$ & \textbf{97.2} & $\textbf{82.4}^{0.2}$ & $\textbf{94.5}^{0.3}$ & $\textbf{96.7}^{0.2}$ & $\textbf{94.3}^{0.1}$ & $ \textbf{78.7}^{0.8}$ & $96.4^{0.2}$ & \textbf{90.5} \\
\bottomrule[1pt]
\end{tabular}
\\[2pt]
\renewcommand{\tabcolsep}{0.239pc} 
\renewcommand{\arraystretch}{1.0} 
\begin{tabular}{c||ccccccccccccc||c}
\toprule[1pt]
\multirow{2}{*}{Methods} & \multicolumn{13}{c||}{Office-Home} & VisDA-2017 \\
  & Ar$\rightarrow$Cl & Ar$\rightarrow$Pr & Ar$\rightarrow$Rw & Cl$\rightarrow$Ar & Cl$\rightarrow$Pr & Cl$\rightarrow$Rw &
Pr$\rightarrow$Ar & Pr$\rightarrow$Cl & Pr$\rightarrow$Rw & Rw$\rightarrow$Ar & Rw$\rightarrow$Cl & Rw$\rightarrow$Pr & Mean & S$\rightarrow$R6 \\
\hline
\hline
ResNet-50 \cite{he2016deep} & 46.3 & 67.5 & 75.9 & 59.1 & 59.9 & 62.7 & 58.2 & 41.8 & 74.9 & 67.4 & 48.2 & 74.2 & 61.4 & 45.3 \\
DANN \cite{ganin2016domain} & 43.8 & 67.9 & 77.5 & 63.7 & 59.0 & 67.6 & 56.8 & 37.1 & 76.4 & 69.2 & 44.3 & 77.5 & 61.7 & 51.0 \\
IWAN \cite{zhang2018importance} & 53.9 & 54.5 & 78.1 & 61.3 & 48.0 & 63.3 & 54.2 & 52.0 & 81.3 & 76.5 & 56.8 & 82.9 & 63.6 & - \\
PADA \cite{cao2018PADA}     & 52.0 & 67.0 & 78.7 & 52.2 & 53.8 & 59.0 & 52.6 & 43.2 & 78.8 & 73.7 & 56.6 & 77.1 & 62.1 & 53.5 \\
SAN \cite{cao2018SAN}       & 44.4 & 68.7 & 74.6 & 67.5 & 65.0 & 77.8 & 59.8 & 44.7 & 80.1 & 72.2 & 50.2 & 78.7 & 65.3 & - \\
ETN \cite{cao2019learning}  & \textbf{59.2} & 77.0 & 79.5 & 62.9 & 65.7 & 75.0 & 68.3 & 55.4 & 84.4 & 75.7 & 57.7 & \textbf{84.5} & 70.5 & - \\
HAFN \cite{xu2019larger}    & 53.4 & 72.7 & 80.8 & 64.2 & 65.3 & 71.1 & 66.1 & 51.6 & 78.3 & 72.5 & 55.3 & 79.0 & 67.5 & 65.1 \\
SAFN \cite{xu2019larger}    & 58.9 & 76.3 & 81.4 & \textbf{70.4} & \textbf{73.0} & 77.8 & \textbf{72.4} & 55.3 & 80.4 & 75.8 & 60.4 & 79.9 & 71.8 & 67.7 \\
\hline
DMP                       & 54.0 & 71.9 &	81.3  & 63.2 & 61.6 & 70.0 & 62.3 & 49.5 & 77.2 & 73.4 & 54.1 & 79.4 & 66.5 & 67.6 \\
DMP+ent                   & 59.0 & \textbf{81.2} & \textbf{86.3} & 68.1 & 72.8 & \textbf{78.8} & 71.2 & \textbf{57.6} & \textbf{84.9} & \textbf{77.3} & \textbf{61.5} & 82.9 & \textbf{73.5} & \textbf{72.7} \\
\bottomrule[1pt]
\end{tabular}
\vspace{-10pt}
\end{table*}

The parameters $\lambda_1$ and $\lambda_2$ were set as $1e1$ and $5e3$, respectively. The Top-$1$ scheme was adopted for the target intra-class loss in Eq. \eqref{eq:IntraLoss&Truncated}. For the ablation study, the models without discriminative structure loss and manifold metric alignment loss are abbreviated as DMP (No DS) and DMP (No AL), respectively.

The experimental results on the Visda-2017 dataset are shown in the top of Table \ref{tab:4Dataset_vanilla}. The recognition rates for each class are reported. It was observed that DMP outperforms the other methods by a large margin in mean accuracy, which is directly derived from the average of class-wise accuracies. It indicates the overall performance of adaptation methods and demonstrates their ability to handle the class imbalance problem. To achieve a higher classification result, methods need to deal with their own ``hard'' classes (e.g., \textit{person} for DANN \cite{ganin2016domain}) and common barriers (e.g., \textit{track}). For example, DMP (No AL) achieves top accuracies in most class-wise results, but the mean accuracy is only 69.7\%. The class-wise results of DMP are more balanced than other methods because the inter-class structure learning in Eq. \eqref{eq:InterLoss} treats all classes equally. As shown in the middle of Table \ref{tab:4Dataset_vanilla}, the proposed method improves the mean accuracy to 68.1\% and achieves the highest accuracy in most of the adaptation tasks on the Office-Home dataset. The second best method is SAFN \cite{chen2019transferability} which improves it mean accuracy to 67.3\% and obtains highest results in tasks \textbf{Rw}$\rightarrow$\textbf{Ar}, \textbf{Rw}$\rightarrow$\textbf{Cl} and \textbf{Rw}$\rightarrow$\textbf{Pr}.

The ablation results also validate the effectiveness of the Riemannian manifold learning framework when both loss terms are used. As the discriminative structure loss provides a separable structure and manifold metric alignment loss bridges the distribution discrepancy between the source and target domains based on the Grassmann distance, both loss terms are important. In Table \ref{tab:4Dataset_vanilla}, the accuracy of DMP is at least 1\% higher than the other variants. The overall results demonstrate the importance and effectiveness of discriminant information, as BSP \cite{chen2019transferability} and DMP are significantly more accurate than the other methods.

The results on the ImageCLEF dataset are shown in the bottom of Table \ref{tab:4Dataset_vanilla}. As the discrepancy between the source and target domains on the ImageCLEF dataset is relatively smaller than that in the other datasets, the baseline model ResNet-50 \cite{he2016deep} achieves 80.7\% accuracy on average. In this case, a more discriminative model is essential for improvement of recognition. DMP encodes the discriminant criterion and alignment constraint simultaneously, thus it outperforms other methods by at least 1.4\%. CDAN+E \cite{long2018conditional} exploits the entropy information and improves accuracy to 87.7\%. The bottom of Table \ref{tab:4Dataset_vanilla} suggests that the accuracies of DMP surpass most of the competitors except for CDAN+E \cite{long2018conditional} on the Office-31 dataset, as CDAN+E is stronger than CDAN by encoding the entropy from the target predictions to the classifier. Note DMP is only 0.3\% lower than CDAN+E on average and achieves highest accuracies on \textbf{D}$\rightarrow$\textbf{A} and \textbf{W}$\rightarrow$\textbf{A}.

\subsection{Comparative Experiment Under Partial Setting}\label{subsec:PartialExperiment}
Comparative experiments were conducted on four datasets under the partial setting. Several advanced partial UDA methods were selected for comparison: DANN \cite{ganin2016domain}, IWAN \cite{zhang2018importance}, PADA \cite{cao2018PADA}, SAN \cite{cao2018SAN}, ETN \cite{cao2019learning}, HAFN and SAFN \cite{xu2019larger}.

The parameters $\lambda_1$ and $\lambda_2$ were set as $1e1$ and $1e0$, respectively. The Top-1 preserving DMP was adopted. Since part of the source information is useless and even negative for adaptation, target entropy loss, which has been widely applied in UDA problems \cite{long2019learning,xu2019larger}, was employed here. It helps the target domain preserve its intrinsic structure by pushing the decision hyper-plane to the low density region. We abbreviate the target entropy version of DMP as DMP+ent.

The results on the Office 31 and ImageCLEF datasets under the partial setting are presented in the top of Table \ref{tab:4Datasets_partial}. The mean accuracy of DMP+ent is higher than others. The improvements are significant in tasks \textbf{A}$\rightarrow$\textbf{W}, \textbf{A}$\rightarrow$\textbf{D}, \textbf{P}$\rightarrow$\textbf{C} and \textbf{C}$\rightarrow$\textbf{I}, and DMP+ent is at least 1.4\% higher than the other methods. The original DMP method is only 0.5\% to 0.6\% lower than the target entropy variant DMP+ent, which demonstrates DMP is effective in mining and preserving the target discriminative structure.

The results on the Office-Home and VisDA-2017 datasets are shown in the bottom of Table \ref{tab:4Datasets_partial}. The results of DMP+ent are significantly higher than those of DMP. The main reasons for the significant improvement are the complex data structure and large domain discrepancy in the Office-Home and VisDA-2017 datasets. The performance of the original DMP is still better than most of the other methods except for AFN and ETN. The SAFN method surpasses the ETN method with 71.8\% in accuracy, and the accuracy of DMP+ent is at least 1.7\% higher than other methods.

\begin{figure}[t]
\centering
\includegraphics[width=0.97\linewidth]{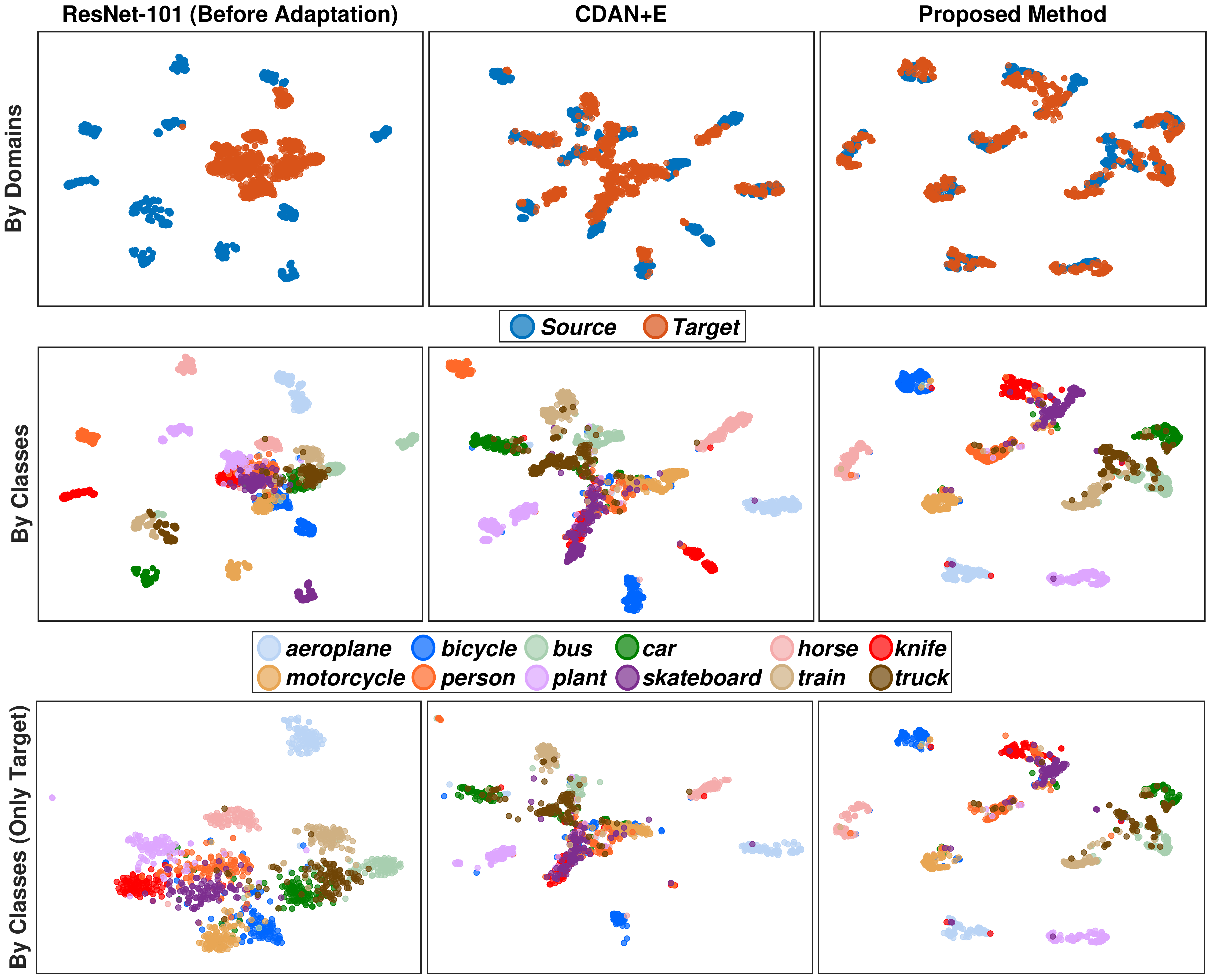}
   \caption{Visualization of learned features using 2-D t-SNE \cite{maaten2008visualizing} on the VisDA-2017 dataset. The first and second rows show the feature representations colored by domains and classes, respectively. The third row shows the target samples colored by classes. Best viewed in color.}
\label{fig:tSNE}
\vspace{-10pt}
\end{figure}

\begin{figure*}[t]
\begin{minipage}{0.22\linewidth}
    \centering{\includegraphics[width=0.99\linewidth]{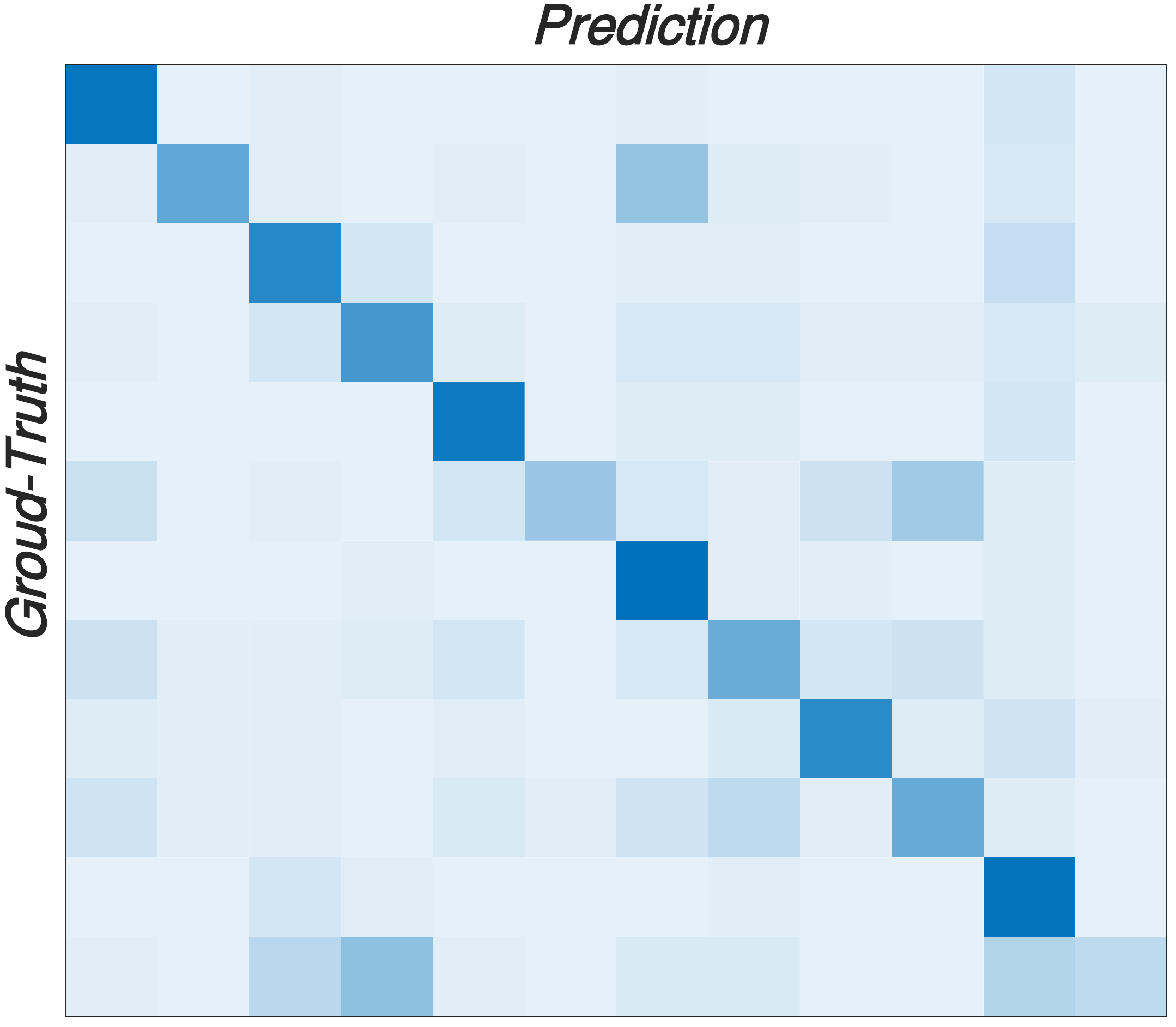}} \\
     (a) ResNet under Vanilla Setting
\end{minipage}
\hfill
\begin{minipage}{0.22\linewidth}
    \centering{\includegraphics[width=0.99\linewidth]{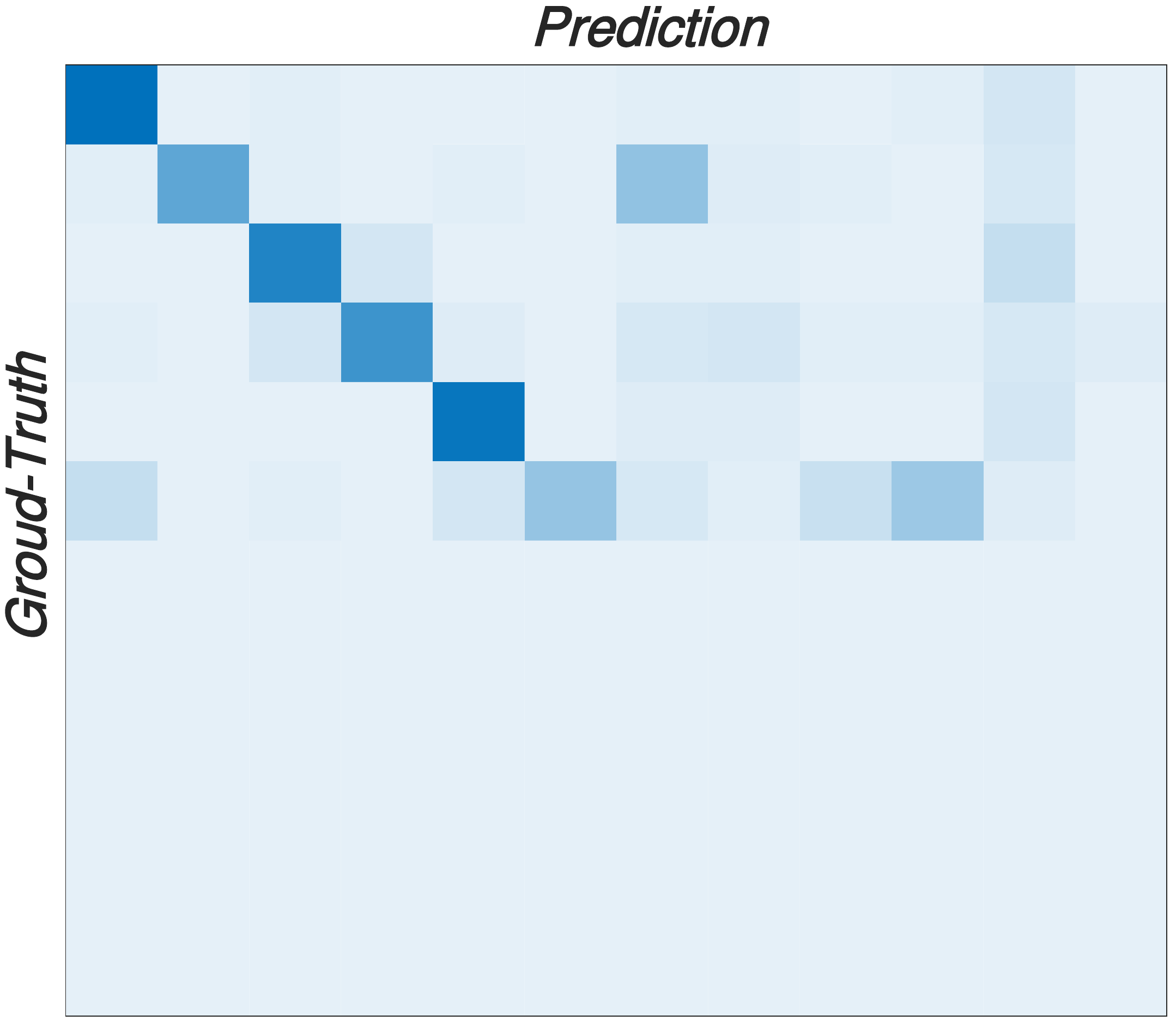}} \\
     (b) ResNet under Partial Setting
\end{minipage}
\hfill
\begin{minipage}{0.22\linewidth}
    \centering{\includegraphics[width=0.99\linewidth]{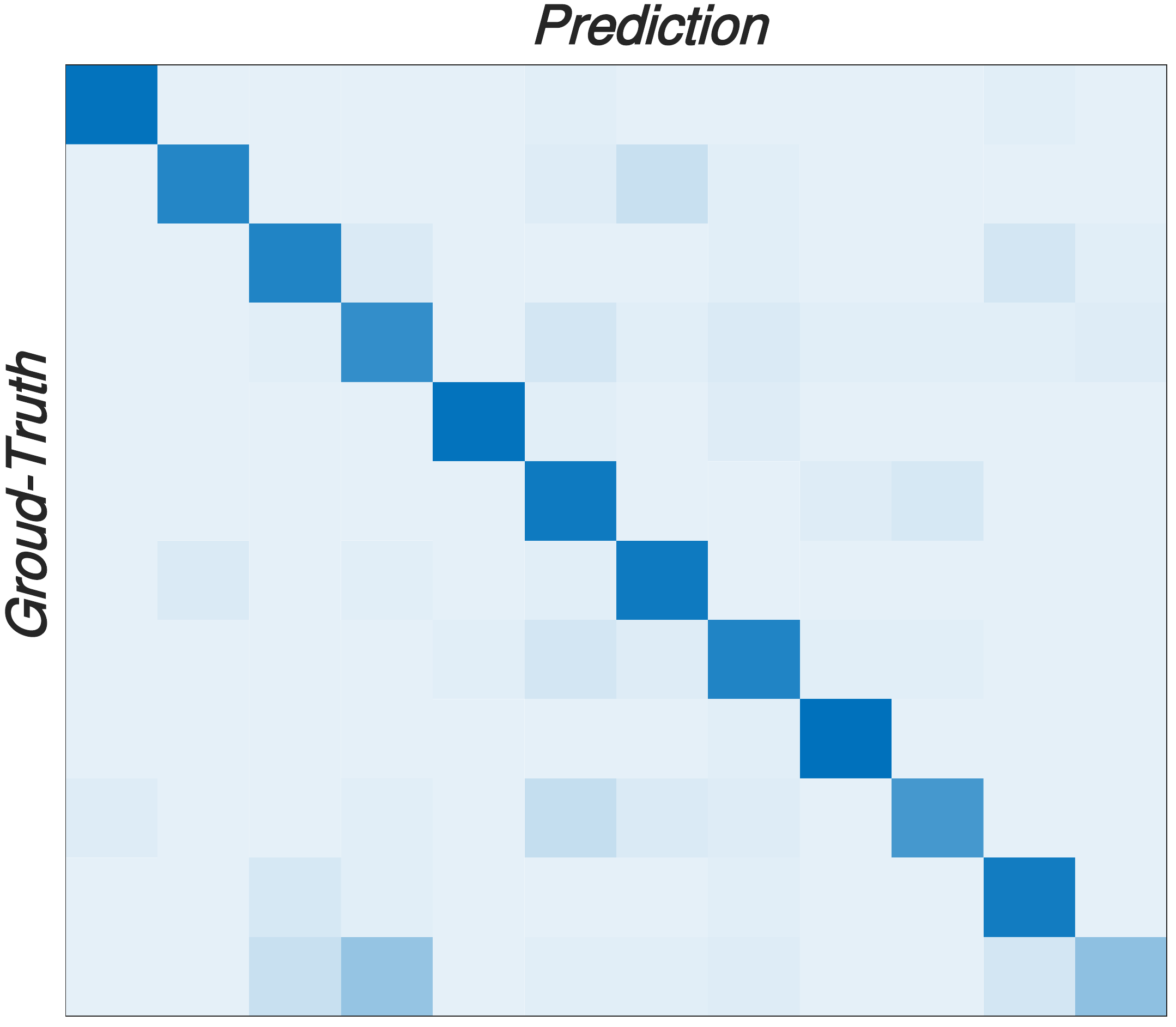}} \\
     (c) DMP under Vanilla Setting
\end{minipage}
\hfill
\begin{minipage}{0.22\linewidth}
    \centering{\includegraphics[width=0.99\linewidth]{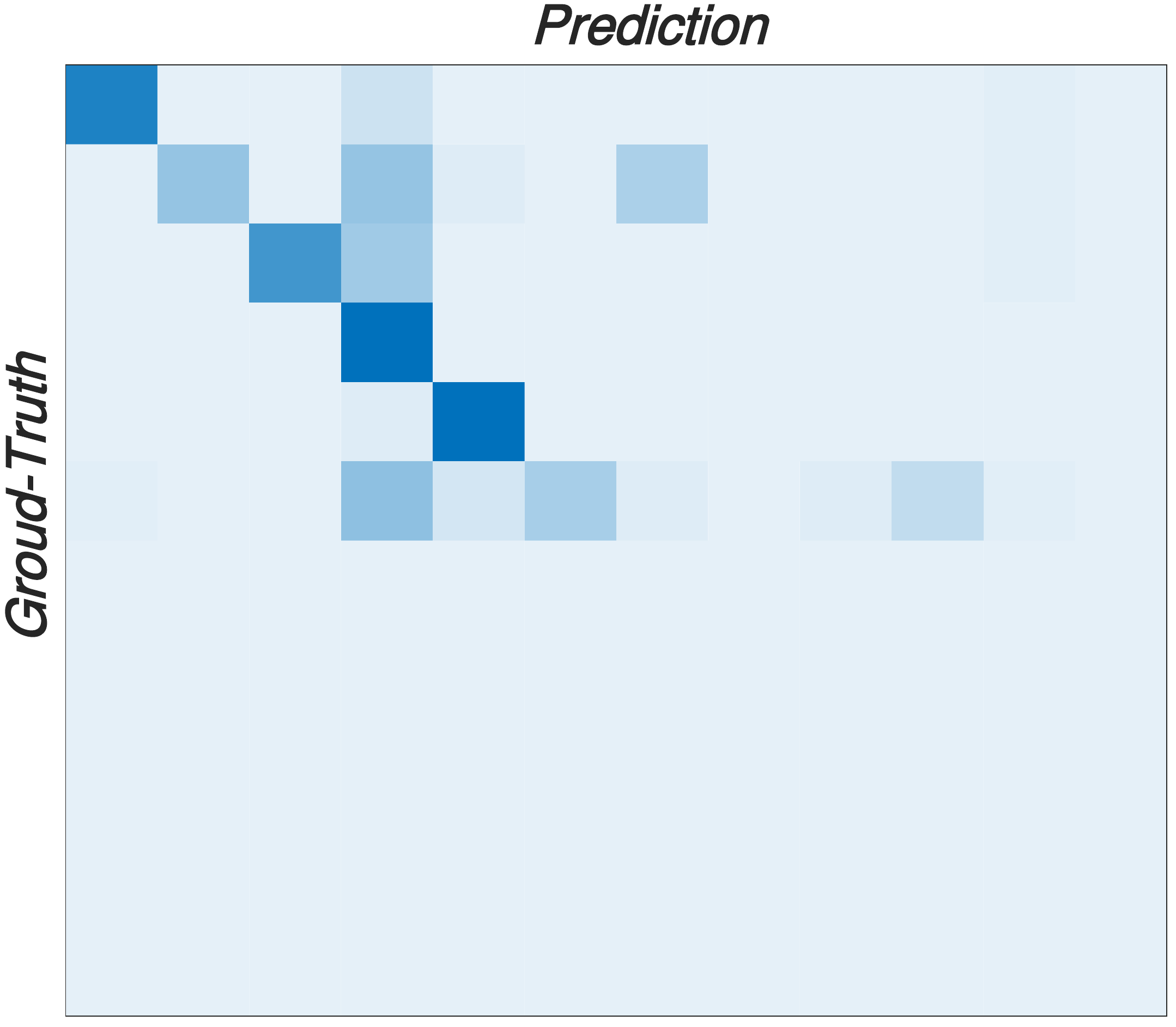}} \\
     (d) DMP under Partial Setting
\end{minipage}
   \caption{Confusion matrices of the target domain on the VisDA-2017 dataset; darker colors represent larger values. (a)-(b): ResNet backbone network, which is regarded as the before adaptation case; (c)-(d): DMP network, which shows the performance after adaptation. Best viewed in color.}
\label{fig:Confusion_Matrix}
\end{figure*}

\subsection{Method Analysis}\label{subsec:MethodAnalysis}

\begin{figure*}[t]
\begin{minipage}{0.245\linewidth}
    \centering{\includegraphics[width=0.99\linewidth]{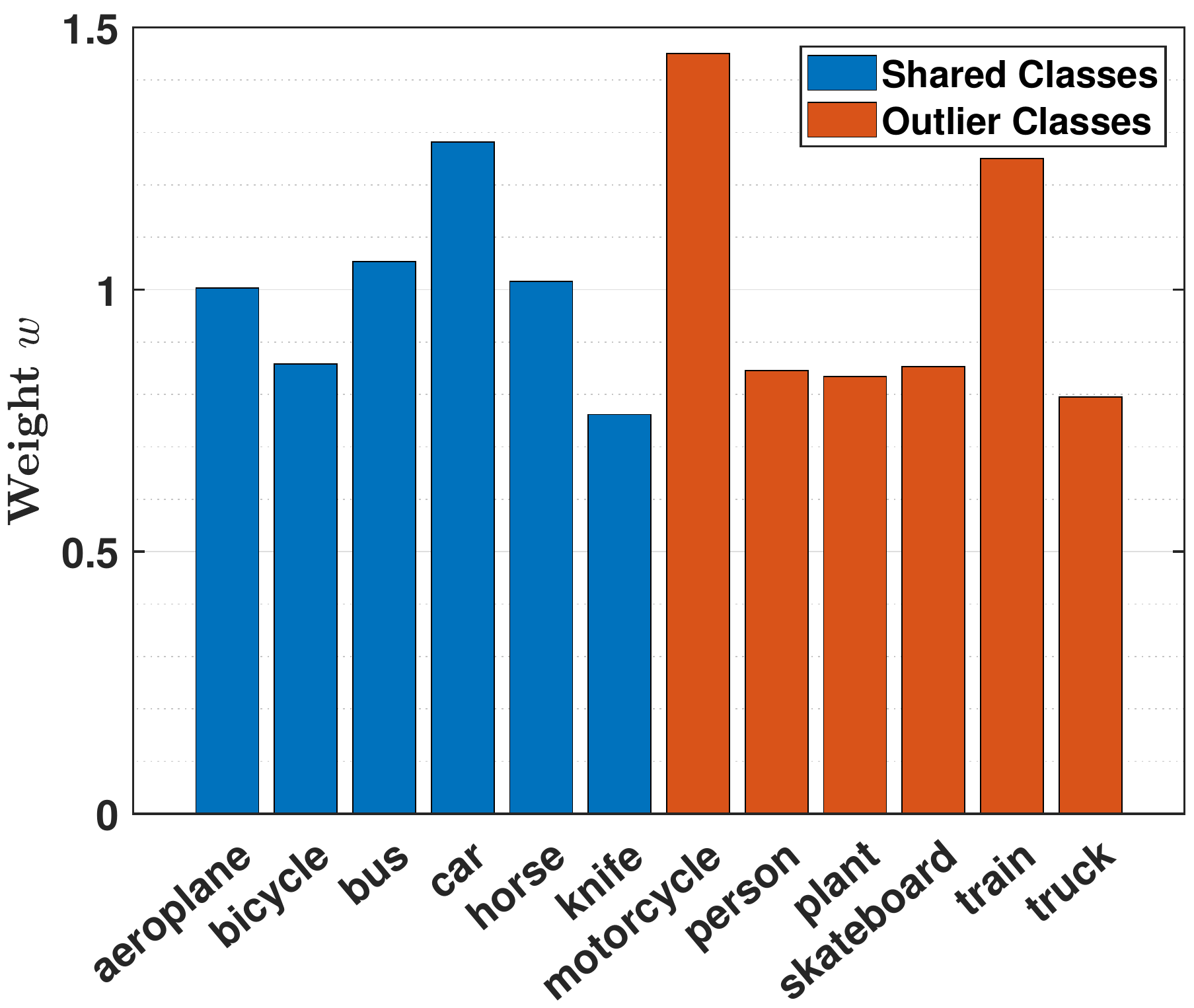}} \\
    (a) Weights of ResNet-50
\end{minipage}
\hfill
\begin{minipage}{0.245\linewidth}
    \centering{\includegraphics[width=0.99\linewidth]{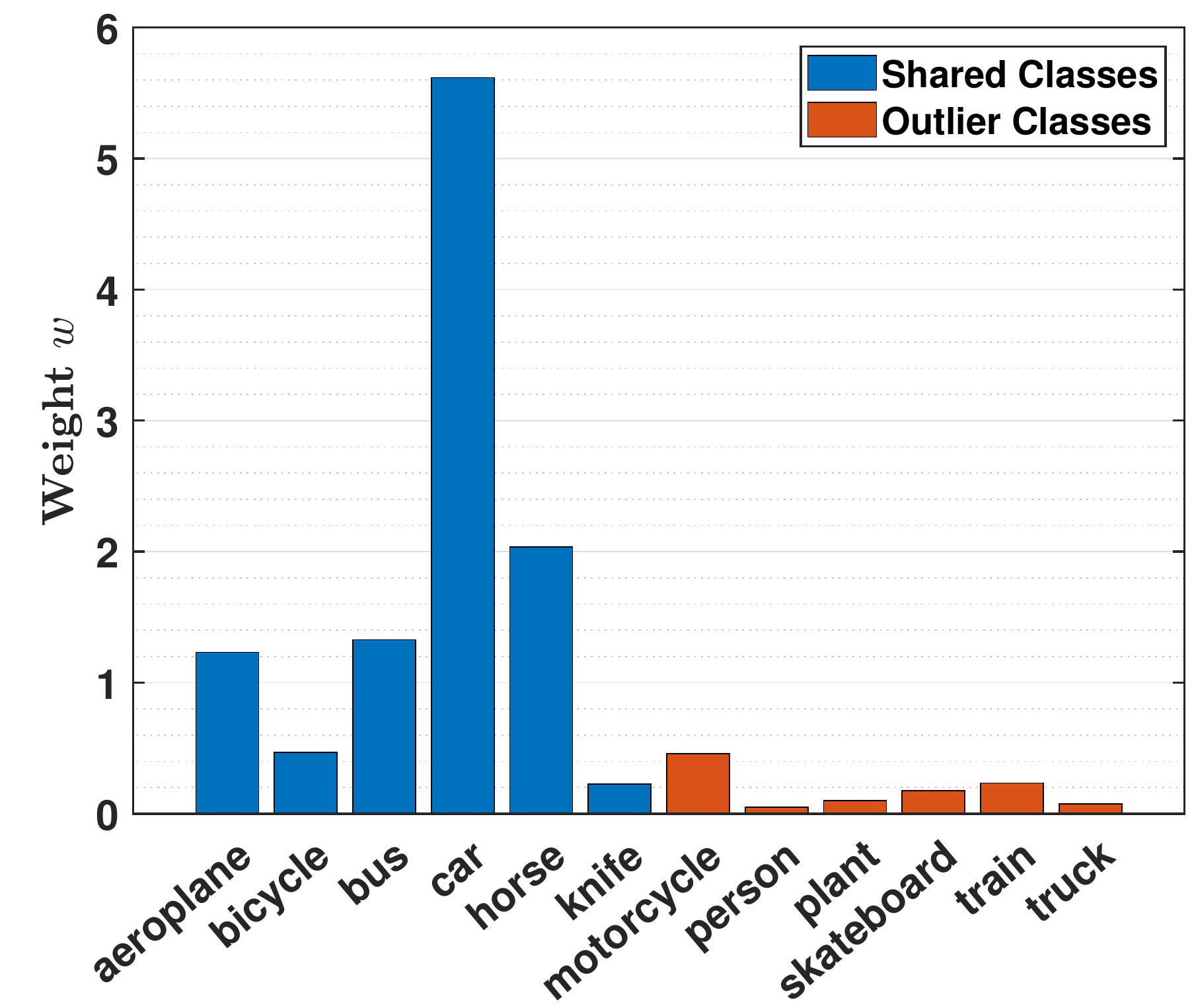}} \\
    (b) Weights of DMP
\end{minipage}
\hfill
\begin{minipage}{0.245\linewidth}
    \centering{\includegraphics[width=0.99\linewidth]{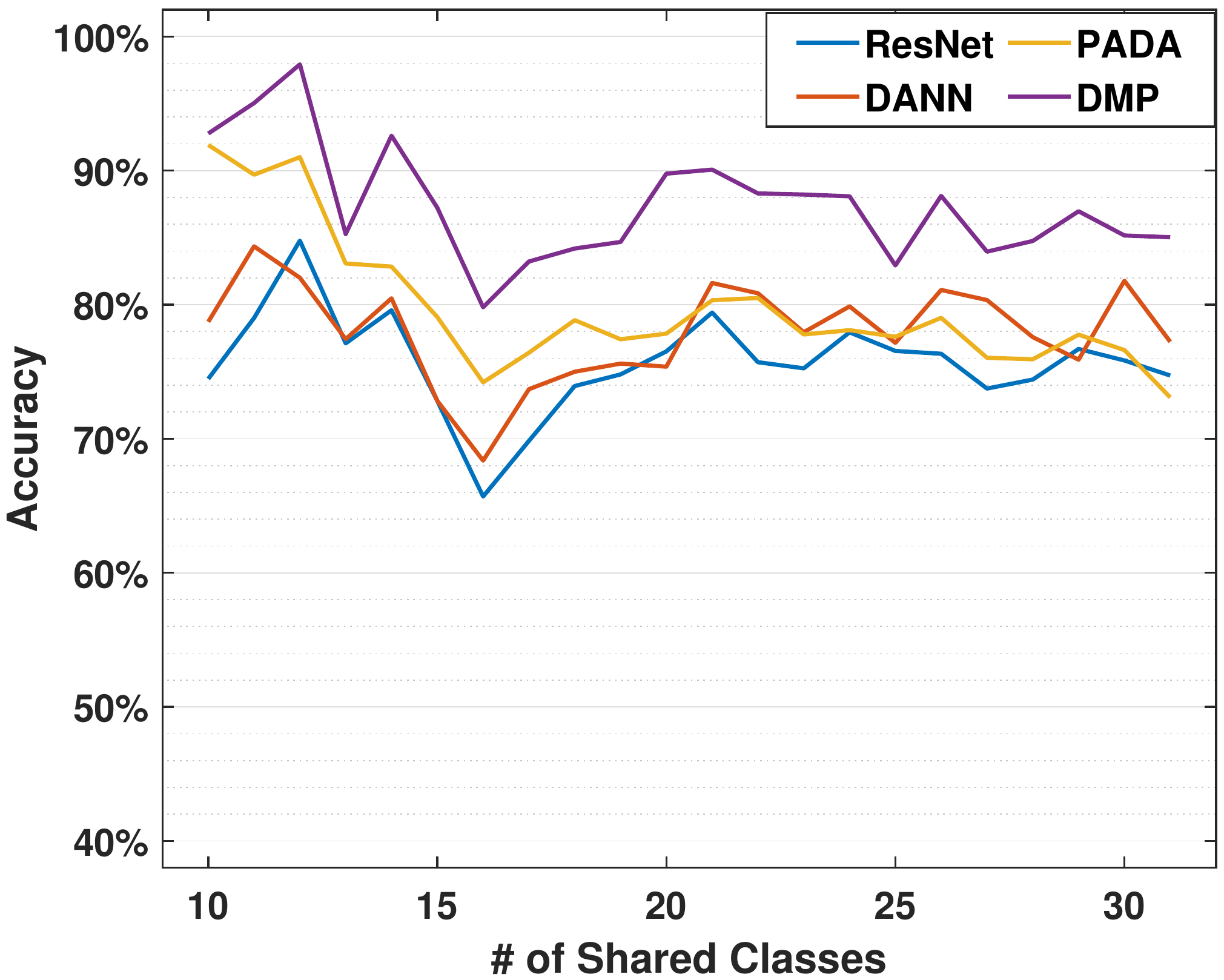}} \\
    (c) A$\rightarrow$W
\end{minipage}
\hfill
\begin{minipage}{0.245\linewidth}
    \centering{\includegraphics[width=0.99\linewidth]{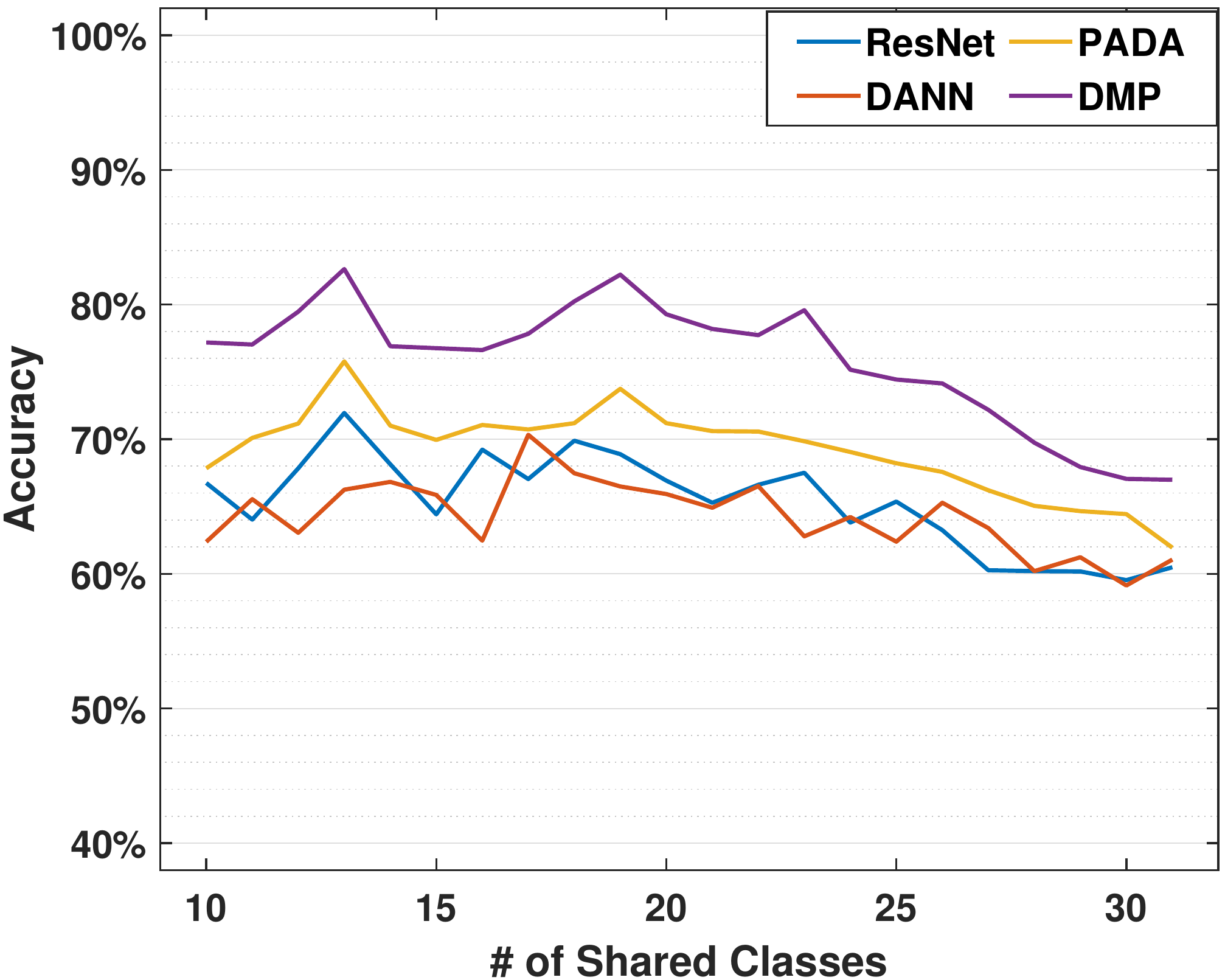}} \\
    (d) W$\rightarrow$A
\end{minipage}
   \caption{(a)-(b): Class weights computed by different methods on the VisDA-2017 dataset under the partial setting. (c)-(d): Accuracy curves by different numbers of shared classes on the Office-31 dataset under the partial setting. Best viewed in color.}
\label{fig:Class_Weight&Overlap}
\vspace{-10pt}
\end{figure*}

\noindent\textbf{Features Visualization.}
To visualize the quality of the adaptation performance, we randomly selected 1200 images from the source and target domains of VisDA-2017 \cite{VisDA-2017}. These 1200 images were collected from 12 classes with 100 images per class. ResNet-101 \cite{he2016deep} and CDAN+E \cite{long2018conditional} were used to compare with DMP.

Figure \ref{fig:tSNE} shows the 2-D representation spaces obtained from the t-SNE \cite{maaten2008visualizing} algorithm. As shown in the first column, there is no alignment constraint between the source and target distributions in the ResNet-101 model. Though the categories on the source domain are separable, the target samples are really indistinguishable. CDAN+E shortens the distance between the source and target domains by using adversarial alignment. Some of the classes have been dragged away from the center, e.g., \textit{plant, car, horse, aeroplane} and \textit{bicycle}. However, the center is still an unresolved region, where \textit{skateboard} and \textit{knife} totally overlap. In the third column, our method further optimizes the structure of the embedding space. The categories are aligned better than ResNet-101 and CDAN+E, leading to a more compact target space. Since DMP achieves the discriminant criterion transductively, the intra-class samples are more compact than in the other methods, and all classes are more separable in both the source and target domains.

\noindent\textbf{Class-wise Confusion and Weight Visualization.}
In this section, we provide the confusion matrices of the ResNet baseline (before adaptation) and DMP on VisDA-2017 under vanilla and partial settings in Figure \ref{fig:Confusion_Matrix}. This shows the quantity of misclassified samples, which cannot be seen in the accuracy table and t-SNE visualization. ResNet-101 and ResNet-50 were employed as backbone networks for the vanilla and partial adaptations, respectively. In Figure \ref{fig:Confusion_Matrix}(a), the off-diagonal elements of ResNet (before adaptation) are larger than DMP (after adaptation) in Figure \ref{fig:Confusion_Matrix}(c), which demonstrates that DMP alleviates the confusion between these classes. For the partial adaptation in Figure \ref{fig:Confusion_Matrix}(b) and \ref{fig:Confusion_Matrix}(d), DMP improves the accuracies of \textit{car} (4th diagonal element) and \textit{horse} (5th diagonal element) to about 95\%. From the perspective of mitigating negative transfer, we can take the misclassified mass of the outlier classes, i.e. the last 6 classes in the partial setting, as a performance evaluation. The density of the outlier region in Figure \ref{fig:Confusion_Matrix}(d) is much lower than that in Figure \ref{fig:Confusion_Matrix}(b), which demonstrates that DMP is effective in identifying the outlier classes and alleviating negative transfer.

To visualize the weighting strategy proposed in Section \ref{subsec:WeightedDMP}, we also plot histograms of the class weights computed from ResNet and DMP in Figure \ref{fig:Class_Weight&Overlap}(a)-(b) under the partial setting. The ResNet model assigns excessive weights to the outlier classes (in orange). DMP assists weight learning by aligning the domains partially and embedding the shared classes discriminatively.

\begin{figure}[t]
\begin{minipage}{0.326\linewidth}
    \centering{\includegraphics[width=0.99\linewidth]{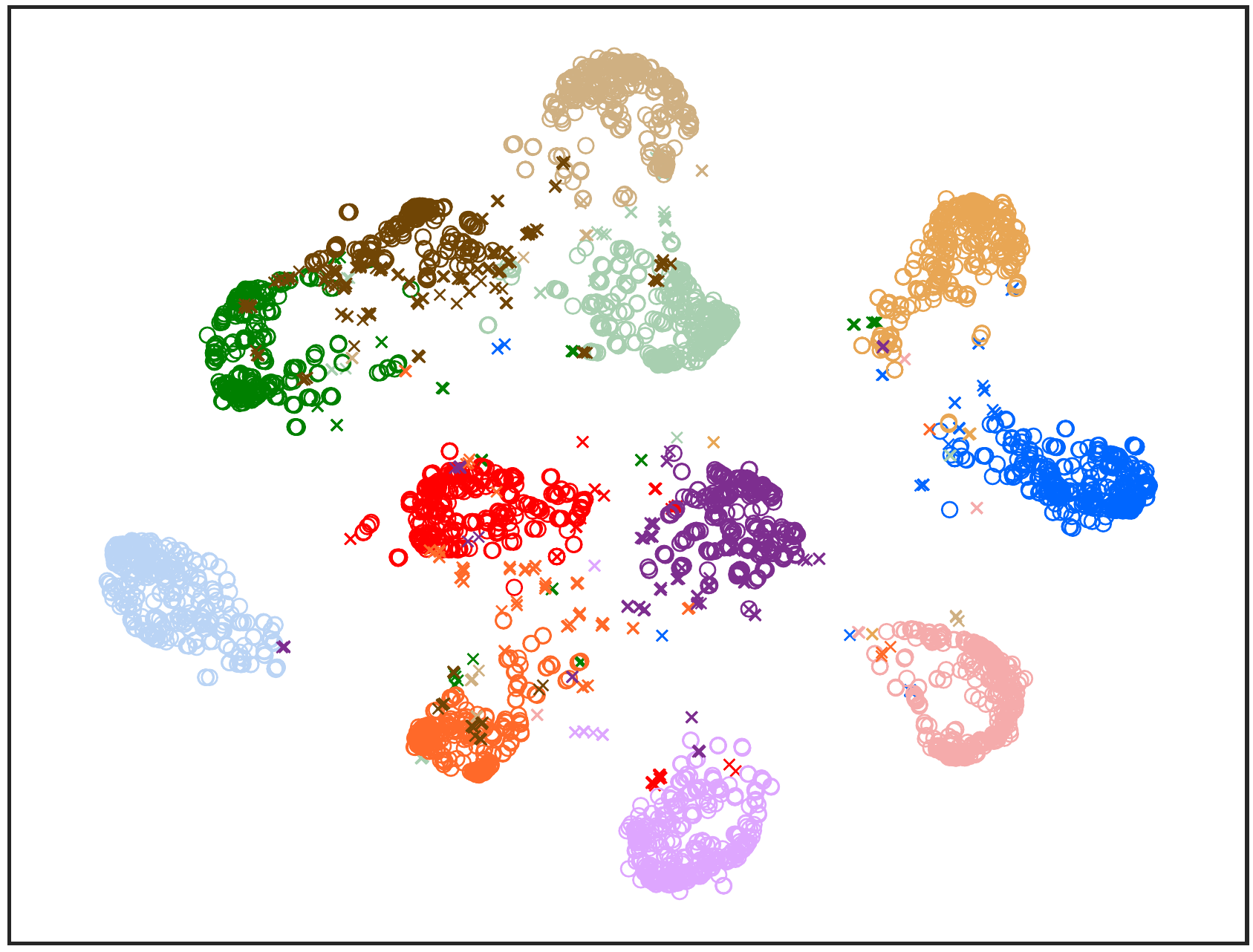}} \\
    (a) $\alpha=0.1$
\end{minipage}
\hfill
\begin{minipage}{0.326\linewidth}
    \centering{\includegraphics[width=0.99\linewidth]{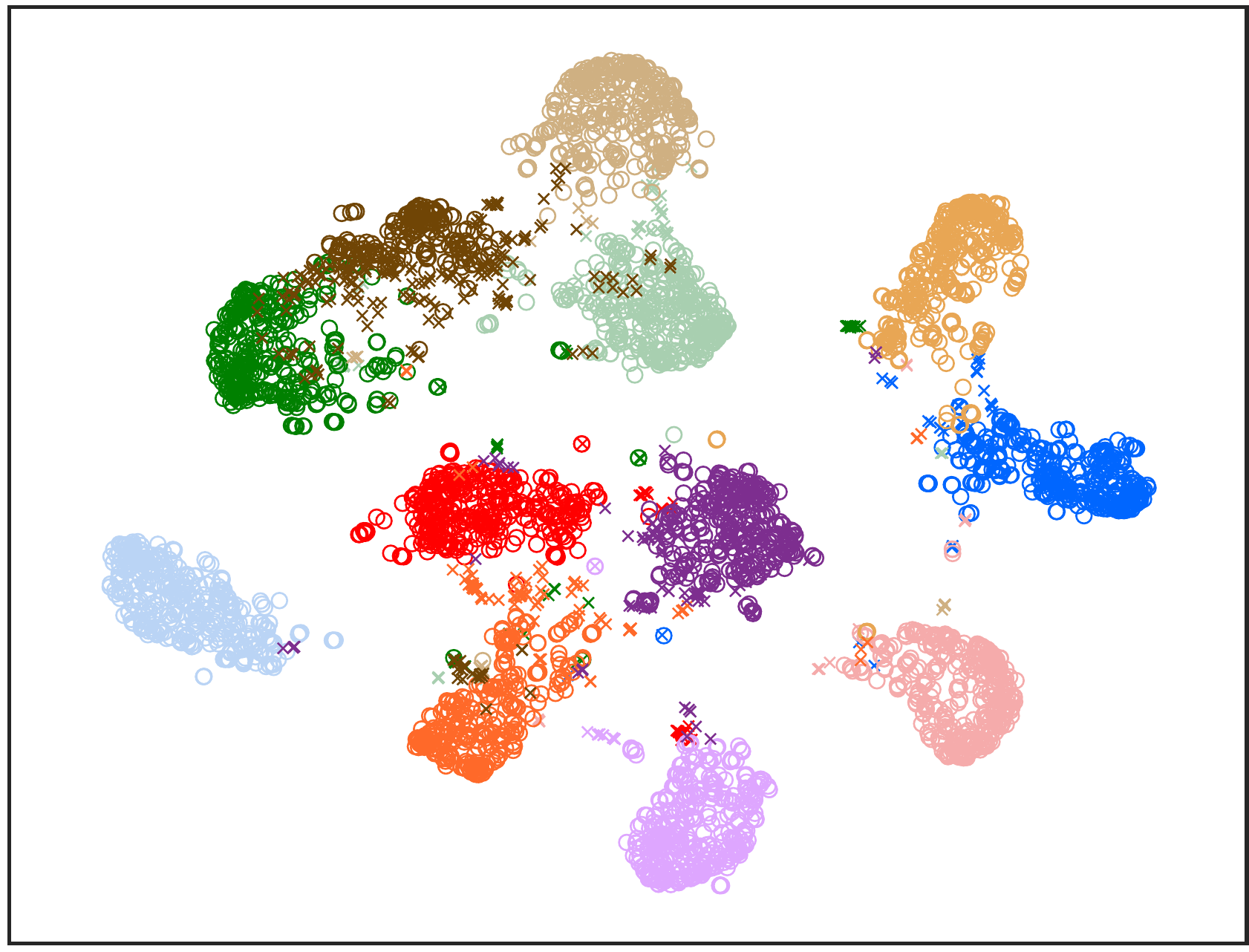}} \\
    (b) $\alpha=0.3$
\end{minipage}
\hfill
\begin{minipage}{0.326\linewidth}
    \centering{\includegraphics[width=0.99\linewidth]{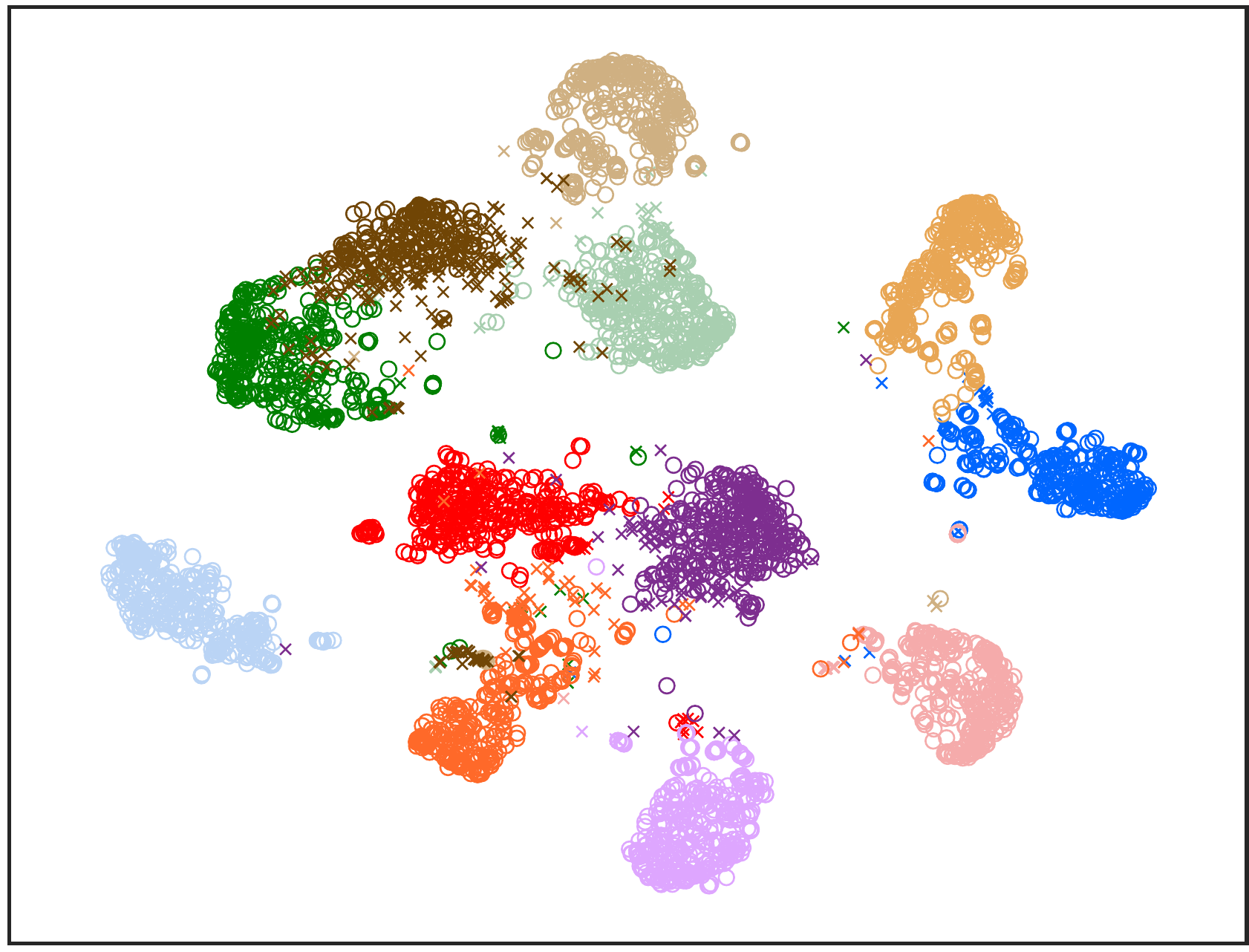}} \\
    (c) $\alpha=0.5$
\end{minipage}
   \caption{Visualization of interpolations using 2-D t-SNE \cite{maaten2008visualizing} on VisDA-2017 with different values of the interpolation parameter $\alpha$. The features are colored by classes. `$\circ$' and `$\times$' represent the correctly and incorrectly classified samples, respectively. Best viewed in color.}
\label{fig:DiscriInterpola}
\end{figure}

\noindent\textbf{Experiment of Shared Classes.}
To evaluate the robustness of methods under different shared class schemes, we varied the target class number from 10 to 31 on the Office 31 dataset while fixing the source class number as 31. The shared classes are extracted in alphabetical order, so the 10 shared classes here are different from the random selection in previous protocols \cite{cao2018PADA,zhang2018importance}. We compared the performance of DMP with ResNet baseline, DANN\cite{ganin2016domain} and PADA\cite{cao2018PADA}. Figure \ref{fig:Class_Weight&Overlap}(c) shows the accuracies in task A$\rightarrow$W. We observed that the accuracy curves of all methods tended to decrease with the increasing number of of shared classes. DMP outperforms the other methods for all shared class cases. When the number of shared classes is small, the accuracies of DANN are lower than other methods as it aligns the target domain with the entire source domain and this results in negative transfer. In Figure \ref{fig:Class_Weight&Overlap}(d), all curves seem more stable and DMP achieves the highest results in all cases. 

\begin{table}[t]
\vspace{-1pt}
\setlength{\abovecaptionskip}{0.cm}
\setlength{\belowcaptionskip}{-0.01cm}
\caption{Class-wise recognition rates (\%) of interpolations on VisDA-2017.
}
\label{tab:DiscriInterpola}
\renewcommand{\tabcolsep}{0.1pc} 
\renewcommand{\arraystretch}{1.0} 
\centering
\resizebox{1\columnwidth}{!}
{
\begin{tabular}{c||cccccccccccc|c}
\toprule[1pt]
$\alpha$ & Plane & bcycl & bus & car & horse & knife & mcyle & person & plant & sktbrd & train & truck & Mean \\
\hline
\hline
0   & 98.0 & 85.0 & 86.0 & 83.0 & 94.0 & 84.0 & 95.0 & 76.0 & 95.0 & 69.0 & 91.0 & 48.0 & 83.7 \\
0.1 & 98.0 & 87.3 & 87.7 & 83.8 & 94.0 & 88.3 & 96.0 & 75.9 & 96.0 & 72.0 & 91.6 & 50.4 & 85.1 \\
0.3 & 99.9 & 91.1 & 90.2 & 90.3 & 94.8 & 93.0 & 98.7 & 80.1 & 96.9 & 81.3 & 93.4 & 53.4 & 88.6 \\
0.5 & 100.0 & 92.1 & 90.9 & 92.4 & 95.3 & 95.8 & 99.8 & 81.8 & 97.7 & 84.0 & 98.2 & 56.1 & 90.4 \\
\bottomrule[1pt]
\end{tabular}
}
\vspace{-12pt}
\end{table}

\noindent\textbf{Disccriminability of Interpolation.}
To further evaluate the discriminability of DMP, we computed unseen interpolations from 1200 randomly selected target images (100 images per class) as $\alpha*\mathbf{h}_i^t+(1-\alpha)*\mathbf{h}_j^t ~ (i\neq j)$, where $\mathbf{h}_i$ and $\mathbf{h}_j$ are from the same class. The interpolation parameter $\alpha$ was selected from \{$0.1,0.3,0.5$\}. The quantitative results are shown in Table \ref{tab:DiscriInterpola}. We observed that the accuracies of interpolations are higher than the data before interpolation (i.e., $\alpha=0$), and the accuracies of larger $\alpha$ are higher, which validates the discriminability of the interpolated embedding features. There are two possible reasons for this phenomenon. First, the fully-connected layer based classification rule guarantees that the predictions of interpolated embedding features are just the interpolations of predictions. Second, the interpolations with larger $\alpha$ are closer to the class centers, thus they are likely to have more discriminative ability. The qualitative analysis through t-SNE visualization of interpolations is presented in Figure \ref{fig:DiscriInterpola}. Note there are 50k+ interpolations, thus many points overlap. We observed that with the increase of $\alpha$, the interpolations gradually move to the class centers. For many misclassified samples, which are scattered at the wrong side of decision boundary, their interpolations are pulled back to their ground-truth clusters and correctly classified. We also observed that the interpolations under all $\alpha$ settings are still inter-class separable and intra-class compact. This is mainly because the embedding features are discriminative.

\begin{table}[t]
\setlength{\abovecaptionskip}{0.cm}
\setlength{\belowcaptionskip}{-0.01cm}
\caption{\textbf{TOP}: Recognition rates (\%) under different sample size settings on ImageCLEF. \textbf{Bottom}: Time comparison on four experimental datasets. Batch and Epoch are abbreviated as \textit{B} and \textit{E} in units.}
\label{tab:SampleSize&TimeComparison}
\renewcommand{\tabcolsep}{0.3pc}
\centering
\begin{tabular}{c||ccc|ccc}
\toprule[1pt]
\multirow{2}{*}{\diagbox{$n^s_c$}{$n^t_c$}} & \multicolumn{3}{c|}{I$\rightarrow$P} & \multicolumn{3}{c}{P$\rightarrow$I} \\
 & 10 & 30 & 50 & 10 & 30 & 50 \\
\hline
\hline
10 & 80.0 & 78.4 & 77.6 & 89.5 & 90.3 & 90.1 \\
30 & 83.5 & 80.7 & 79.4 & 91.4 & 91.8 & 91.2 \\
50 & 84.3 & 81.4 & 80.7 & 91.4 & 92.1 & 92.5 \\
\bottomrule[1pt]
\end{tabular}
\\[5pt]
\begin{tabular}{c||ccc|c}
\toprule[1pt]
Dataset  & Office-31 & Office-Home & ImageCLEF & VisDA-2017 \\
Task & A$\rightarrow$W & Ar$\rightarrow$Cl & I$\rightarrow$P & S$\rightarrow$R \\
\hline
ResNet \cite{he2016deep}       & 6.7\textit{s/E}  & 60.6\textit{s/E} & 5.1\textit{s/E} & 0.453\textit{s/B} \\
CDAN+E \cite{long2018conditional} & 10.7\textit{s/E} & 72.0\textit{s/E} & 7.7\textit{s/E} & 1.238\textit{s/B} \\
DMP                               & 10.7\textit{s/E} & 70.8\textit{s/E} & 6.8\textit{s/E} & 0.767\textit{s/B} \\
\bottomrule[1pt]
\end{tabular}
\vspace{-8pt}
\end{table}

\noindent\textbf{Experiment on Sample Size and Time Comparison.}
As the performance of discriminative structure learning is in reference to the sample size, we chose \{$10, 30, 50$\} samples per class for each domain in ImageCLEF to form several subsets. The proposed model was trained and tested on these subsets under the standard UDA protocol. The results are shown in top of Table \ref{tab:SampleSize&TimeComparison}, where $n^{s/t}_c$ represents the amount of samples per class on the source/target domain. We observed that a larger source sample size $n_c^s$ provides a higher accuracy since more ground-truth labels are used to learn discriminative structure. For some cases in task \textbf{P}$\rightarrow$\textbf{I}, a larger $n_c^t$ also increases the precision. This demonstrates the target predictive information is also helpful in mining the discriminative information. The accuracies at $(n^{s}_c, n^{t}_c)=(30,30)$ are slightly lower than that at $(50,50)$, it indicates that DMP is still effective when the sample size is small. This is because DMP enhances the discriminability of the target features by simultaneously using the source labels and target predictive information.

We also evaluated the efficiency of DMP by comparing the training time. Results in the bottom of Table \ref{tab:SampleSize&TimeComparison} show that DMP is faster than CDAN and slightly slower than ResNet. It validates that the time efficiency is mainly decided by the architecture of the baseline network and the post alignment network (e.g., adversarial alignment). As DMP only adds three low-dimensional fully connected layers for discriminative alignment, it does not introduce many parameters relative to the backbone network and adversarial alignment while improving the recognition rates significantly.

\section{Conclusion}\label{sec:Conclusion}
In this paper, we proposed a discriminative manifold propagation method for both vanilla and partial UDA problems. Both transferability and discriminability are simultaneously reached by the manifold alignment and discriminative embedding. To optimize the structure of the target domain, the source labels and target predictive information were encoded probabilistically and transductively into the discriminant criterion. A global discriminative structure was approximated via the pre-built prototypes. The theoretical error bounds, which are guaranteed to find the optimal dimensions for the Grassmann and affine Grassmann manifolds during the alignment, were derived. Numerical simulation and extensive comparisons demonstrated the effectiveness of the proposed method.

\bibliographystyle{IEEEtran}
\bibliography{DMP}

\vspace{-15pt}
\begin{IEEEbiography}[{\resizebox{1.0in}{1.3in}{\includegraphics*{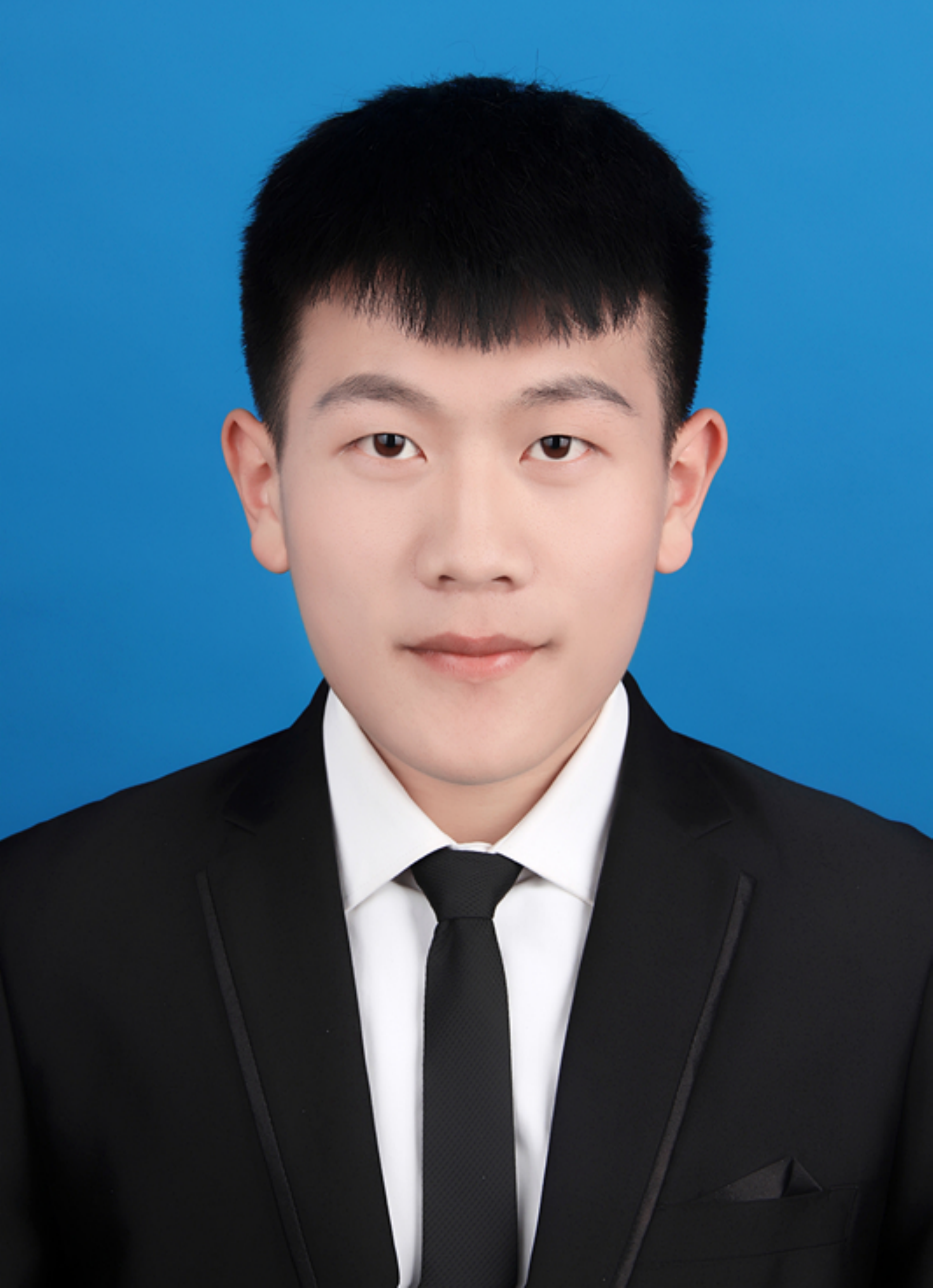}}}]{You-Wei Luo}
received the B.S. degree in statistics from China University of Mining and Technology, Xuzhou, China, in 2018.  He is currently pursuing the Ph.D. degree with  the School of Mathematics, Sun Yat-sen University, Guangzhou, China. His research interests include image processing, manifold learning and transfer learning.
\end{IEEEbiography}

\vspace{-25pt}
\begin{IEEEbiography}[{\resizebox{1.0in}{1.3in}{\includegraphics*{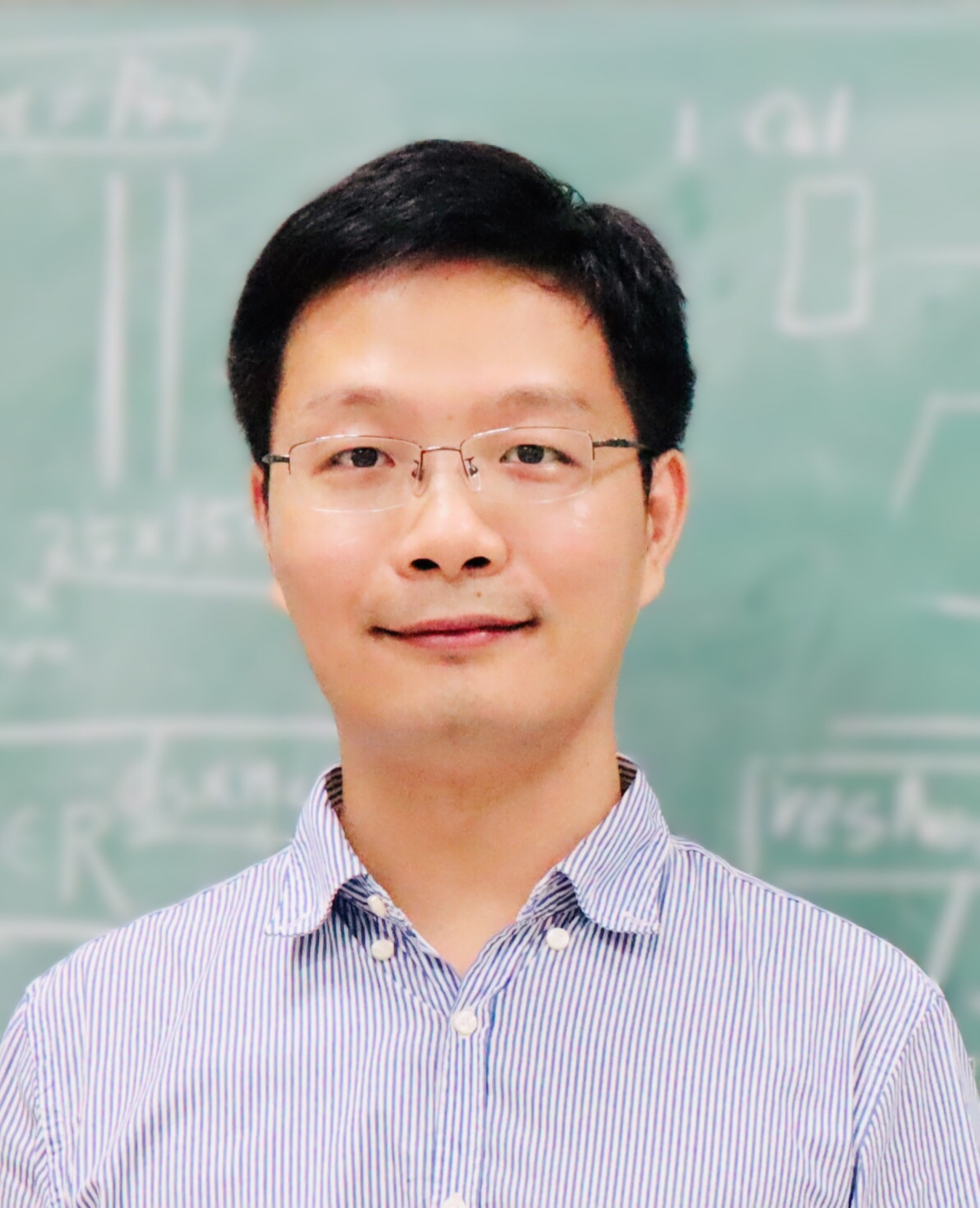}}}]{Chuan-Xian Ren}
received the PhD degree from Sun Yat-Sen University, Guangzhou, China, in 2010. He is currently Associate professor of the School of Mathematics, Sun Yat-Sen University. His research interests include image processing, pattern recognition and machine learning.
\end{IEEEbiography}

\vspace{-25pt}
\begin{IEEEbiography}[{\resizebox{1.0in}{1.3in}{\includegraphics*{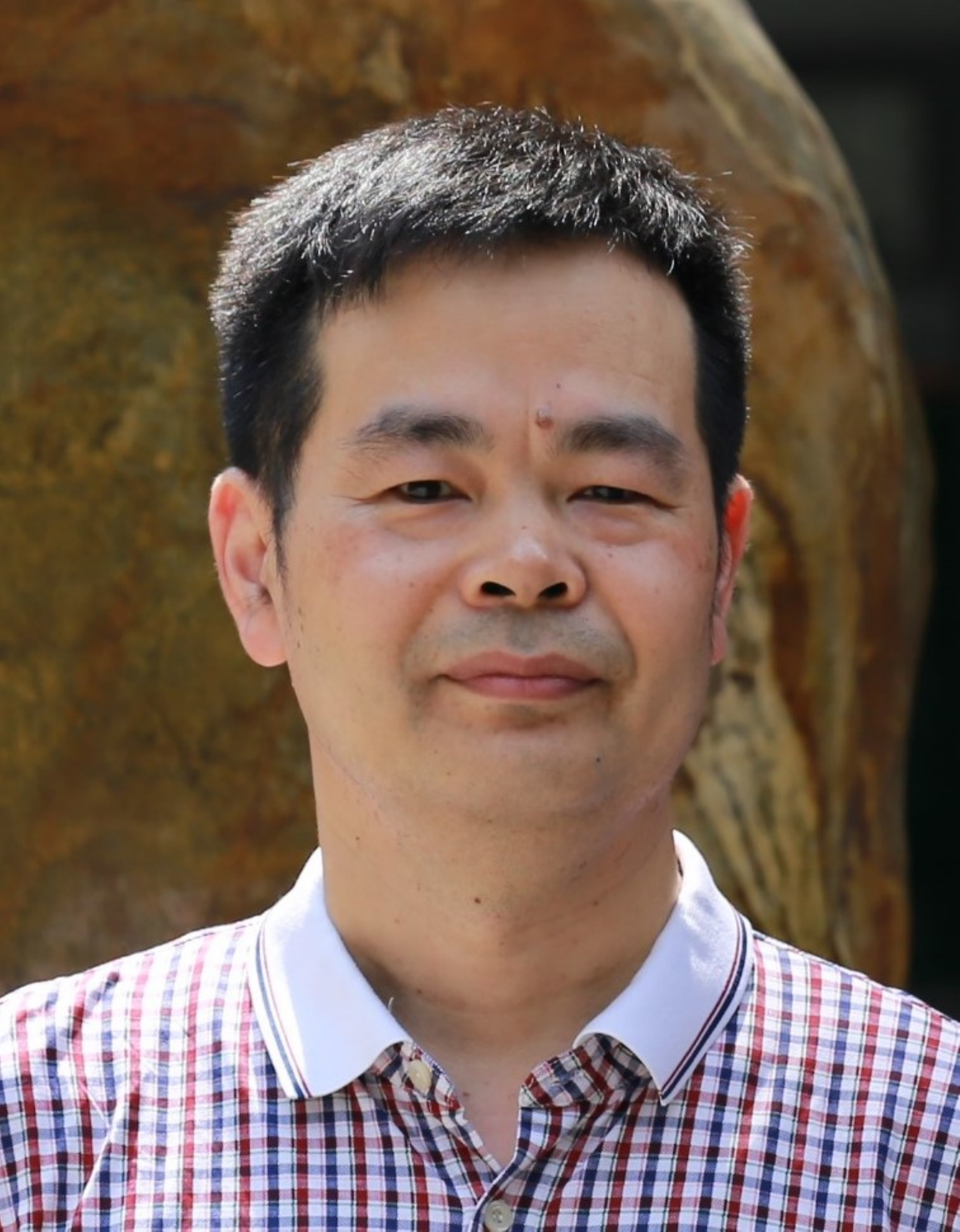}}}]{Dao-Qing Dai}
(M'07-SM'19) received the Ph.D. degree in mathematics from Wuhan University, Wuhan, China, in 1990. From 1998 to 1999, he was an Alexander von Humboldt Research Fellow with Free University, Berlin, Germany. He is currently a Professor with the School of Mathematics, Sun Yat-Sen University, China. He has authored or co-authored over 100 refereed technical papers. His current research interests include image processing, wavelet analysis, and pattern recognition.
\end{IEEEbiography}

\vspace{-25pt}
\begin{IEEEbiography}[{\resizebox{1.0in}{1.3in}{\includegraphics*{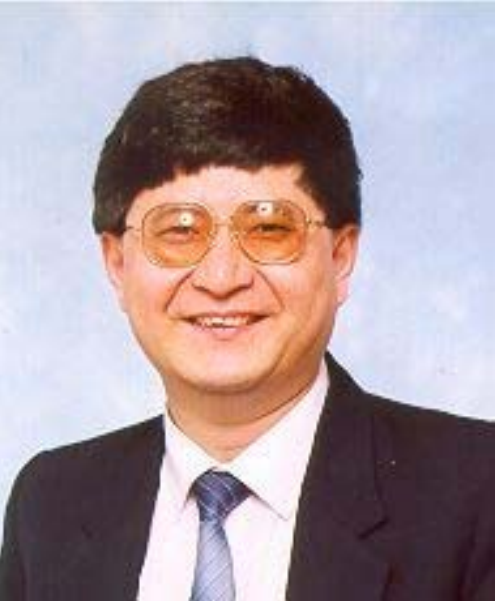}}}]{Hong Yan}
(S'88-M'89-SM'93-F'06) received his PhD degree from Yale University. He was Professor of Imaging Science at the University of Sydney and currently is Chair Professor of Computer Engineering and Wong Chung Hong Professor of Data Engineering at City University of Hong Kong. Professor Yan's research interests include image processing, pattern recognition, and bioinformatics. He has over 600 journal and conference publications in these areas. Professor Yan is an IEEE Fellow and IAPR Fellow. He received the 2016 Norbert Wiener Award from the IEEE Systems, Man and Cybernetics Society for contributions to image and biomolecular pattern recognition techniques. He is a member of the European Academy of Sciences and Arts.
\end{IEEEbiography}

\end{document}